\documentclass{article}

\usepackage{microtype}
\usepackage{graphicx}
\usepackage{svg}
\usepackage{subcaption}
\usepackage{caption}
\usepackage{tikz}
\usepackage{booktabs} 
\usepackage{multirow}
\usepackage{pgffor}
\usepackage{graphbox}

\usepackage{amsmath,amsfonts,bm}

\usepackage{amsthm}
\newtheorem{Problem}{Problem}

\def\eqref#1{equation~\ref{#1}}

\def\1{\bm{1}}

\def\vz{{\bm{z}}}

\def\mA{{\bm{A}}}

\def\mG{{\bm{G}}}

\def\mX{{\bm{X}}}

\def\mZ{{\bm{Z}}}

\DeclareMathAlphabet{\mathsfit}{\encodingdefault}{\sfdefault}{m}{sl}
\SetMathAlphabet{\mathsfit}{bold}{\encodingdefault}{\sfdefault}{bx}{n}

\def\gE{{\mathcal{E}}}

\def\gG{{\mathcal{G}}}

\def\gL{{\mathcal{L}}}

\def\gV{{\mathcal{V}}}

\def\sR{{\mathbb{R}}}

\def\emA{{A}}

\DeclareMathOperator*{\argmin}{arg\,min}

\newcommand{\stitle}[1]{\vspace{1ex}\noindent{\bf #1}}

\usepackage{hyperref}
\usepackage{color}
\usepackage{array}
\newcolumntype{C}[1]{>{\centering\arraybackslash}p{#1}}

\usepackage[accepted]{icml2024}

\usepackage{amssymb}
\usepackage{amsthm}
\usepackage{enumitem}
\usepackage{dsfont}

\newcommand{\bamo}{BA-2motifs}
\newcommand{\bathree}{BA-3motifs}
\newcommand{\benz}{Benzene}
\newcommand{\mutag}{MUTAG}
\newcommand{\fluo}{Fluoride-Carbonyl}
\newcommand{\alk}{Alkane-Carbonyl}
\newcommand{\ours}{ProxyExplainer}

\theoremstyle{plain}

\usepackage[textsize=tiny]{todonotes}

\icmltitlerunning{Generating In-Distribution Proxy Graphs for Explaining Graph Neural Networks }

\begin{document}

\twocolumn[
\icmltitle{Generating In-Distribution Proxy Graphs for Explaining Graph Neural Networks }
\icmlsetsymbol{equal}{*}

\begin{icmlauthorlist}
\icmlauthor{Zhuomin Chen}{fiu}
\icmlauthor{Jiaxing Zhang}{njit}
\icmlauthor{Jingchao Ni}{uh}
\icmlauthor{Xiaoting Li}{visa}
\icmlauthor{Yuchen Bian}{amzs}
\icmlauthor{Md Mezbahul Islam}{fiu}
\icmlauthor{Ananda Mohan Mondal}{fiu}
\icmlauthor{Hua Wei}{asu}
\icmlauthor{Dongsheng Luo}{fiu}

\end{icmlauthorlist}

\icmlaffiliation{fiu}{Knight Foundation School of Computing and Information Sciences, Florida International University, Miami, USA}
\icmlaffiliation{njit}{New Jersey Institute of Technology, Newark, USA}
\icmlaffiliation{uh}{Department of Computer Science, University of Houston, Houston, USA}
\icmlaffiliation{visa}{Visa Research, USA}
\icmlaffiliation{amzs}{Amazon Search A9, USA}
\icmlaffiliation{asu}{School of Computing and Augmented Intelligence, Arizona State University, Tempe, USA}

\icmlcorrespondingauthor{Zhuomin Chen}{zchen051@fiu.edu}
\icmlcorrespondingauthor{Dongsheng Luo}{dluo@fiu.edu}

\icmlkeywords{XAI, Graph Neural Networks}

\vskip 0.3in
]

\printAffiliationsAndNotice{} 

\begin{abstract}
Graph Neural Networks (GNNs) have become a building block in graph data processing, with wide applications in critical domains. The growing needs to deploy GNNs in high-stakes applications necessitate explainability for users in the decision-making processes. A popular paradigm for the explainability of GNNs is to identify explainable subgraphs by comparing their labels with the ones of original graphs. This task is challenging due to the substantial distributional shift from the original graphs in the training set to the set of explainable subgraphs, which prevents accurate prediction of labels with the subgraphs. To address it, in this paper, we propose a novel method that generates proxy graphs for explainable subgraphs that are in the distribution of training data. We introduce a parametric method that employs graph generators to produce proxy graphs. A new training objective based on information theory is designed to ensure that proxy graphs not only adhere to the distribution of training data but also preserve explanatory factors. Such generated proxy graphs can be reliably used to approximate the predictions of the labels of explainable subgraphs. Empirical evaluations across various datasets demonstrate our method achieves more accurate explanations for GNNs.

\end{abstract}

\section{Introduction}
\label{sec:intro}
Graph Neural Networks (GNNs) have emerged as a pivotal technology for handling graph-structured data, demonstrating remarkable performance in various applications including node classification and link prediction~\citep{kipf2017semisupervised,hamilton2017inductive,velivckovic2017graph,scarselli2008graph}. Their growing use in critical sectors such as healthcare and fraud detection has escalated the need for explainability in their decision-making processes~\cite{wu2022survey,li2022survey,zhang2024trustworthy}. To meet this demand, a variety of explanation methods have been recently developed to interpret the behavior of GNN models. These methods primarily concentrate on identifying a subgraph that significantly impacts the model's prediction for a particular instance~\cite{ying2019gnnexplainer,luo2020parameterized}.

A prominent approach to explain GNNs involves the Graph Information Bottleneck (GIB) principle~\cite{wu2020graph}. This principle focuses on extracting a compact yet informative subgraph from the input graph, ensuring that this subgraph retains sufficient information for the model to maintain its original prediction. A key aspect of the GIB approach is evaluating the predictive capability of such a subgraph. Typically, this is accomplished by feeding the subgraph into the GNN model and comparing its prediction against that of the complete input graph. 

\begin{figure}
    \centering
    \begin{subfigure}[b]{0.22\textwidth}
        \centering
        \includegraphics[width=1.2\textwidth]{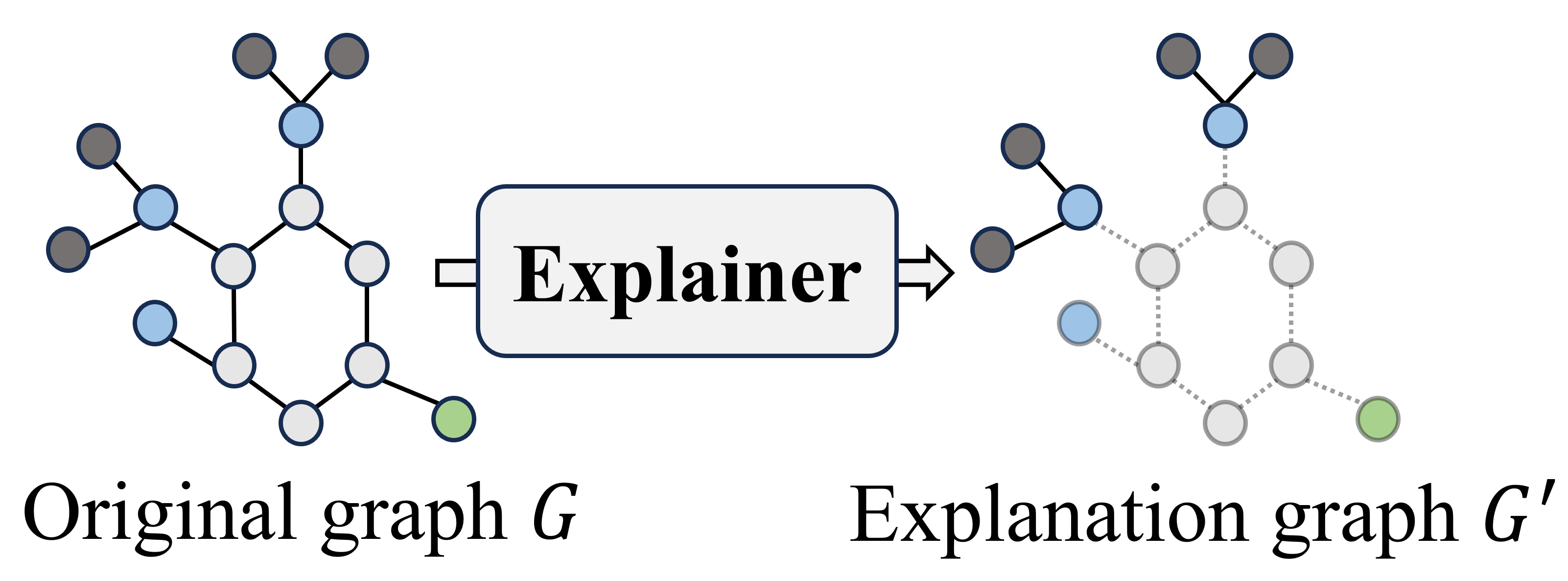}
        \caption{Explanation process}
        \label{fig:introduction_sub1}
    \end{subfigure}
    \hfill 
    \hspace{8mm}
    \hfill 
    \begin{subfigure}[b]{0.2\textwidth}
        \centering
        \includegraphics[width=0.9\textwidth]{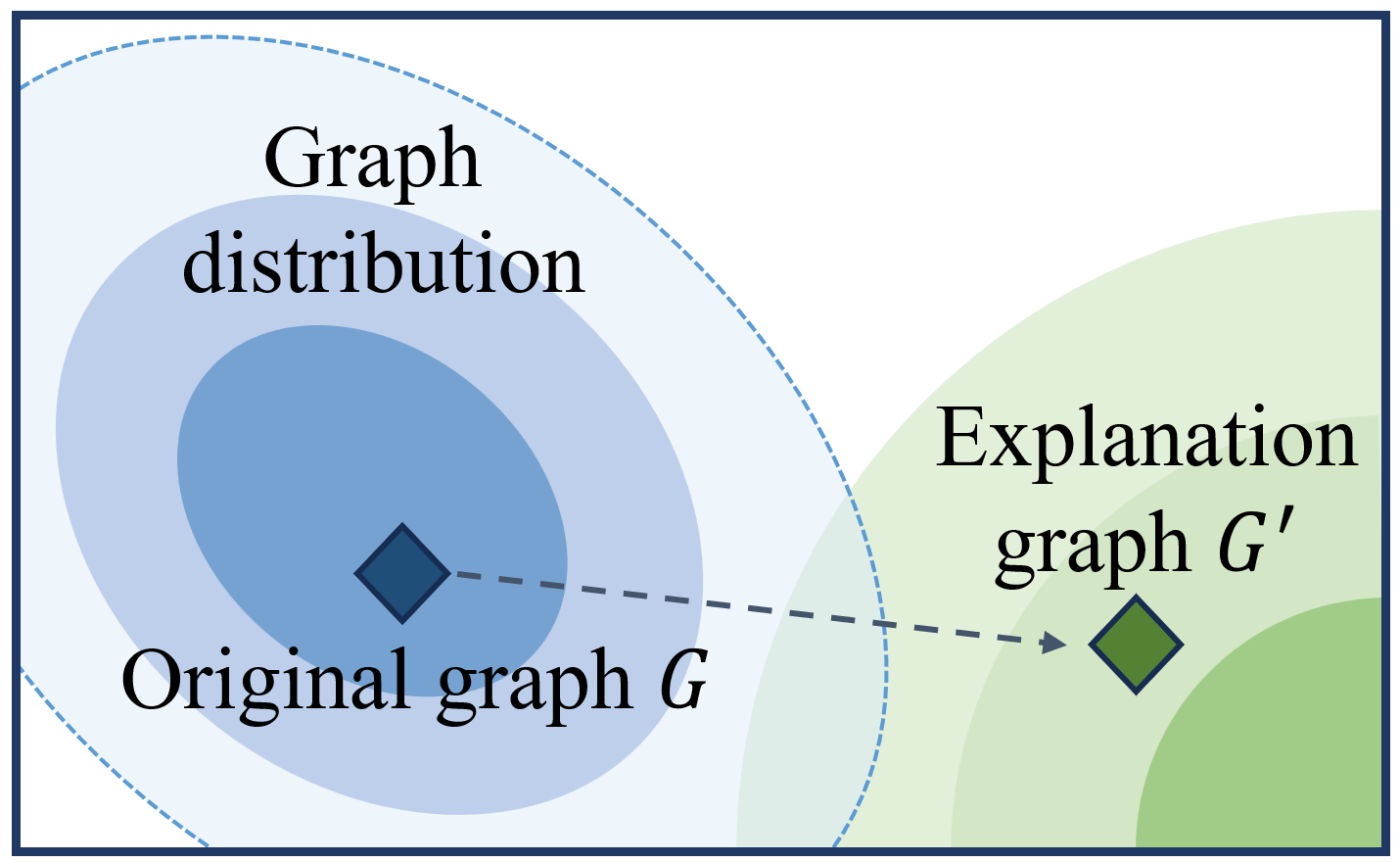}
        \caption{OOD problem}
        \label{fig:introduction_sub2}
    \end{subfigure}
    \caption{Examples of the explanation process and out-of-distribution problem. (a) is the explanation process for a graph learning model. The original graph $G$ undergoes an explanation process, resulting in an explanation graph $G'$ that highlights the most significant features and relationships; (b) shows the explanation graph $G'$ is out of distribution where the GNN is trained.}
    \label{fig:introduction}
\end{figure}

Although it is intuitively correct, the underlying assumption of the aforementioned approach -- GNN model can make accurate predictions on explanation subgraphs -- may not always hold. 
As shown in Figure~\ref{fig:introduction}, explanation subgraphs can significantly deviate from the distribution of original graphs,
leading to an Out-Of-Distribution (OOD) issue~\citep{zhang2023mixupexplainer,fang2023cooperative, fang2023evaluating, zheng2024towards}. For instance, in the MUTAG dataset~\citep{debnath1991structure}, each graph represents a molecule, with nodes symbolizing atoms and edges indicating chemical bonds. The molecular graphs in this dataset usually contain hundreds of edges. In contrast, the $NO_2$ functional group, identified as a key subgraph influencing positive mutagenicity in a molecule, comprises merely 2 edges. This stark contrast in structural properties leads to a significant difference in the distributions of explanation subgraphs and original graphs. Since the model was trained with original graphs,  the reliability of predictions on subgraphs is undermined due to the distribution shifting problem.

Several pioneering studies have attempted to address this distributional challenge~\cite{fang2023cooperative,zhang2023mixupexplainer,zhang2023regexplainer}. For example, CGE regards the GNN model as a transparent, fully accessible system. It considers the GNN model as a teacher network and employs an additional ``student'' network to predict the labels of explanation subgraphs~\cite{fang2023cooperative}. As another example, MixupExplainer~\cite{zhang2023mixupexplainer} generates a mixed graph for the explanation by blending it with a non-explanatory subgraph from a different input graph. This method posits that the mixup graph aligns with the distribution of the original input graphs. However, this claim is predicated on a rather simplistic assumption that the explanation and non-explanatory subgraphs are independently drawn.  However, in real-world applications, these methods often face practical limitations. The dependence on a ``white box'' model in CGE and the oversimplified assumptions in MixupExplainer are not universally applicable. Instead, GNN models are usually given as ``black boxes'', and the graphs in real-life applications do not conform to strict independence constraints, highlighting the need for more versatile and realistic approaches to the OOD problem.

In response to these challenges, in this work, we introduce an innovative concept of proxy graphs to the realm of explainable GNNs. These proxy graphs are designed to be both explanation-preserving and distributionally aligned with the training data. By this means, the predictive capabilities of explanation subgraphs can be reliably inferred from the corresponding proxy graphs. We begin with a thorough investigation into the feasible conditions necessary for generating such in-distributed proxy graphs. 
Leveraging our findings, we further propose a novel architecture that incorporates variational graph auto-encoders to produce proxy graphs. Specifically, we utilize a graph auto-encoder to reconstruct the explainable subgraph and another variational auto-encoder to generate a non-explanatory subgraph. A proxy graph is then obtained by combining two output subgraphs of auto-encoders. We delineate our main contributions as follows:
\begin{itemize}[leftmargin=*]
    \item We systematically analyze and address the challenge of the OOD issue in explainable GNNs, which is pivotal for enhancing the reliability and interpretability of GNNs in real-world applications.
    \item We introduce an innovative parametric method that incorporates graph auto-encoders to produce in-distributed proxy graphs that are both situated in the original data distribution and preserve essential explanation information. This facilitates more precise and interpretable explanations in GNN applications.
    \item Through comprehensive experiments on various real-world datasets, we substantiate the effectiveness of our proposed approach, showcasing its practical utility and superiority in producing explanations. 
\end{itemize}

\section{Notations and Preliminary}

\subsection{Notations and Problem Formulation} 
We denote a graph $G$ from an graph set $\gG$ by a triplet $(\gV, \gE; \mX)$, where $\gV = \{v_1, v_2, ..., v_n\}$ is the node set and $ \gE \subseteq \gV \times \gV$ is the edge set.  $\mX \in \sR^{n\times d}$ is the node feature matrix, where $d$ is the feature dimension and the $i$-th row is the feature vector associated with node $v_i$.  The adjacency matrix of $G$ is denoted by $\mA \in \{0,1\}^{n\times n}$, which is determined by the edge set $\mathcal{E}$ such that $\emA_{ij} = 1$ if $(v_i,v_j)\in \mathcal{E}$, $\emA_{ij} = 0$, otherwise. In this paper, we focus on
the graph classification task, and node classification can be converted to computation graph classification problem~\cite{ying2019gnnexplainer,luo2020parameterized}. Specifically, for 
graph classification task, each graph $G$ is associated with a label $Y \in \mathcal{Y}$. The to-be-explained GNN model $f(\cdot)$ has been well-trained to classify $G$ into its class,   {\em i.e.}, {$f:\gG\mapsto \{1,2,\cdots,|\mathcal{Y}|\}$}. 

Following the existing works \cite{ying2019gnnexplainer,luo2020parameterized, yuan2022explainability,huang2024factorized}, the explanation methods under consideration in this paper are model/task agnostic and treat GNN models as black boxes --- {\em i.e.}, the so-called \textit{post-hoc, instance-level} explanation methods. Formally, our research problem is described as follows:
\begin{Problem} [Post-hoc Instance-level GNN Explanation]
\label{prob:exp}
Given a to-be-explained GNN model $f(\cdot)$ and a set of graphs $\gG$,  the goal of post-hoc instance-level explanation is to learn a parametric function so that for an arbitrary graph $G\in \gG$,  it finds a compact subgraph $G^* \subseteq G$ that can ``explain'' the prediction $f(G)$. The parametric mapping $\Psi_\psi:\gG\mapsto \gG^*$ is called an explanation function, where $\gG^*$ is the alphabet of $G^*$, and $\psi$ is the parameter of the explanation function.
\end{Problem}

\subsection{Graph Information Bottleneck as the Objective Function}
The Information Bottleneck (IB) principle, foundational in learning dense representations, suggests that optimal representations should balance minimal and sufficient information for predictions~\cite{tishby2000information}. This concept has been adapted for GNNs through the Graph Information Bottleneck (GIB) approach, consolidating various post-hoc GNN explanation methods like GNNExplainer~\cite{ying2019gnnexplainer} and PGExplainer~\cite{luo2020parameterized}. The GIB framework aims to find a subgraph $G^*$ to explain a GNN model's prediction on a graph $G$ as~\cite{zhang2023mixupexplainer}:

\begin{equation}
\label{eq:GIB}
     G^* = \argmin_{G'} I(G, G') - \alpha I(Y,G'),
\end{equation}
where $G'$ is a candidate explanatory subgraph, $Y$ is the label, and  $\alpha$ is a balance parameter. The first term is the mutual information between the original graph $G$ and the explanatory subgraph $G'$ and the second term is negative mutual information of $Y$  and the explanatory subgraph $G'$. At a high level, the first term encourages detecting a small and dense subgraph for explanation, and the second term requires that the explanatory subgraph is label preserving. 

Due to the intractability of mutual information between $Y$ and $G'$ in~\eqref{eq:GIB}, some existing works~\cite{wu2020graph,miao2022interpretable} derived a parameterized variational lower bound of $I(Y, G')$:
\begin{equation}\label{eq:IYG}
I(Y , G') \geq  \mathbb{E}_{G',Y}[\log P(Y|G')] + H(Y),
\end{equation}
where the first term measures how well the subgraphs predict the labels. A higher value indicates that the subgraphs are, on average, more predictive of the correct labels.  The second term $H(Y)$ quantifies the amount of inherent unpredictability or variability in the labels. By introducing~\eqref{eq:IYG} to~\eqref{eq:GIB}, a tractable upper bound is used as the objective: 

\begin{equation}\label{eq:GIBupper}
G^* = \argmin_{G'}{I(G, G') - \alpha \mathbb{E}_{G',Y}[\log P(Y|G')]},
\end{equation}
where $H(Y)$ is omitted due to its independence to $G'$.  

\noindent \textbf{The OOD Problem in GIB.} In the existing research, the estimation of $P(Y|G')$ is typically achieved by applying the to-be-explained model $f(\cdot)$ to the input graph $G'$~\cite{ying2019gnnexplainer,miao2022interpretable}. A critical assumption in 
these approaches involves the GNN model's ability to accurately predict candidate explanation subgraphs $G'$. This assumption, however, 
overlooks the OOD problem, where the distribution of explanation subgraphs significantly deviates from that of the original training graphs~\cite{zhang2023mixupexplainer,fang2023evaluating, fang2023cooperative, fang2022regularization, zheng2024towards,amara2023ginx}.
Formally, let $P_\gG$ 
be the distribution of the training graphs, and $P_{\gG'}$ 
be the distribution of the explanation subgraphs. The core issue in the GIB objective function is caused by $P_\gG \neq P_{\gG'}$. This distributional disparity undermines the predictive reliability of the model $f(\cdot)$, trained on $P_\gG$ when applied to subgraphs from $P_{\gG'}$. As a result, the predictive power of explanations provided by $f(G')$ is an unreliable approximation of $P(Y|G')$ in ~\eqref{eq:GIBupper}.

\section{Graph Information Bottleneck with Proxy Graphs}
\label{sec:method}
In this section, we propose the concept of \textit{proxy graphs} to mitigate the aforementioned OOD issue. A proxy graph not only retains the label information present in $G'$ but also conforms to the distribution of the original graph dataset. Specifically, we assume that proxy graphs $\tilde{G}$ are drawn from 
the distribution $P_\gG$ and reformulate the estimation of $P(Y|G')$ by marginalizing over the distribution of proxy graphs as follows:

\begin{equation}
\label{eq:surGIB}
    P(Y|G') = \mathbb{E}_{\tilde{G} \sim P_\gG} [P(Y|\tilde{G}) \cdot P(\tilde{G}|G')].
\end{equation}

In~\eqref{eq:surGIB}, we address the OOD challenge by predicting $Y$ using a proxy graph $\tilde{G}$ instead of directly using $G'$. This approach is particularly effective when the conditional probability $P(Y|\cdot)$ is approximated by the model $f(\cdot)$. To facilitate the maximization of the likelihood as outlined in~\eqref{eq:surGIB}, we further approximate $P(\tilde{G}|G')$ with a parameterized function, denoted as $Q_{\boldsymbol{\phi}}(\tilde{G}|G')$. The formal representation is thus given by
\begin{equation}\label{eq:surGIBpar}
P(Y|G') = \mathbb{E}_{\tilde{G} \sim P_\gG} [P(Y|\tilde{G}) \cdot Q_{\boldsymbol{\phi}}(\tilde{G}|G')],
\end{equation}
where $\boldsymbol{\phi}$ denotes model parameters.  

Given the combinatorial complexity inherent in graph structures, it is computationally infeasible to enumerate all potential proxy graphs $\tilde{G}$ from the unknown distribution $P_\gG$, which is necessary for calculating~\eqref{eq:surGIBpar}. To overcome this challenge, we propose to approximate $P_\gG$ with the parameterized function $Q_{\boldsymbol{\phi}}(\tilde{G}|G')$, and estimate \eqref{eq:surGIBpar} with a Monte Carlo estimator that samples proxy graphs from $Q_{\boldsymbol{\phi}}(\tilde{G}|G')$. That is
\begin{equation}\label{eq:estimator}
\begin{aligned}
P(Y|G') = \mathbb{E}_{\tilde{G} \sim Q_{\boldsymbol{\phi}}(\tilde{G}|G')} [P(Y|\tilde{G})],
\end{aligned}
\end{equation}
with constraints
\begin{equation}\label{eq:constraint}
\begin{aligned}
Q_{\boldsymbol{\phi}}(\tilde{G}|G') \approx P_\gG, \quad H(Y|\tilde{G}) \approx H(Y|G'),
\end{aligned}
\end{equation}
where the first constraint ensures that $\tilde{G}$ is sampled from a distribution that approximates $P_\gG$, effectively addressing the OOD challenge. The second constraint guarantees that the label information preserved in $\tilde{G}$ is similar to that in $G'$. Therefore, combining \eqref{eq:GIBupper}, \eqref{eq:estimator} and \eqref{eq:constraint}, our proxy graph-induced objective function becomes:

\begin{equation}\label{eq:train}
\begin{aligned}
&\argmin_{G'}{I(G, G') - \alpha \mathbb{E}_{G',Y}[\log \mathbb{E}_{\tilde{G} \sim Q_{\boldsymbol{\phi}}(\tilde{G}|G')} [P(Y|\tilde{G})]]}\\
&\text{s.t.}~~~Q_{\boldsymbol{\phi}}(\tilde{G}|G') \approx P_\gG,~~~H(Y|\tilde{G})\approx H(Y|G').
\end{aligned}
\end{equation}

\textbf{Bi-level optimization.}
In~\eqref{eq:train}, we formulate a joint optimization loss function that aims to identify the optimal explanation alongside its corresponding proxy graphs. Building upon this, we refine the framework by replacing the first constraint with a distributional distance measure, specifically the Kullback-Leibler (KL) divergence, between $Q_{\boldsymbol{\phi}}(\tilde{G}|G')$ and $P_\gG$. This leads to the development of a bi-level optimization model. Formally, the model is expressed as follows:

\begin{equation}\label{eq:bi-level-opt}
\begin{aligned}
& \argmin_{G'}{I(G, G') - \alpha \mathbb{E}_{G',Y}[\log \mathbb{E}_{\tilde{G} \sim Q_{\boldsymbol{\phi}^*}(\tilde{G}|G')} [P(Y|\tilde{G})]]}\\
& \quad \text{where } \boldsymbol{\phi}^* = \argmin_{\boldsymbol{\phi}}{ \text{KL}(Q_{\boldsymbol{\phi}}(\tilde{G}|G'), P_\gG)}, \\
& \quad \quad \quad \text{s. t. } H(Y|\tilde{G}) \approx H(Y|G').
\end{aligned}
\end{equation}

\subsection{Derivation of Outer Optimization}
\label{sec:outter}
For the outer optimization objective, we elaborate on 
instantiating the first term $I(G, G')$. Akin to the original graph, we denote the explanation subgraph $G'$ by $(\gV,\gE',\mX)$, whose adjacency matrix is $\mA'$.
We follow~\cite{miao2022interpretable} to include a variational approximation distribution $R(G')$ for the distribution $P(G')$. Then, we obtain its upper bound as follows:
\begin{equation}
 I(G, G') \leq \mathbb{E}_G[\text{KL}(P (G'|G)||R(G'))].   
\end{equation}

We follow the Erdős–Rényi model~\cite{erdHos1960evolution} and assume that each edge in $G'$ has a probability of being present or absent, independently of the other edges. Specifically, 
we assume that the existence of an edge $(u,v)$ in $G'$ is determined by a Bernoulli variable $A_{u,v}' \sim \text{Bern}(\pi'_{uv})$. 
Thus, $P (G'|G)$ can be 
decomposed as $P (G'|G) = \prod_{(u,v)\in\gE} P(A_{uv}'|\pi'_{uv})$. The bound is always true for any prior distribution $R(G')$. 
We follow an existing work and assume that in the prior distribution, the existence of an edge $(u,v)$ in $G'$ is determined by another Bernoulli variable $A^{''}_{u,v} \sim \text{Bern}(r)$, where $r\in [0,1]$ is a hyper-parameter~\cite{miao2022interpretable}, independently to the graph $G$. We have $R(G') = P(|\gE|)\prod_{(u,v)\in \gE} P(A^{''}_{uv})$.
Thus, the KL divergence between $P (G'|G)$ and the above marginal distribution $R(G')$ becomes:
\begin{equation}
\label{eq:igg}
\begin{aligned}
    & \text{KL}(P (G'|G)||R(G'))  \\
           = &   \sum_{(u,v)\in \gE} \pi'_{uv}\log \frac{\pi'_{uv}}{r} + (1-\pi'_{uv})\log \frac{1-\pi'_{uv}}{1-r} + \text{Const}. \\ 
\end{aligned}
\end{equation}

By replacing $I(G, G')$ with the above tractable upper bound, we obtain the loss function, denoted by $\mathcal{L}_\text{exp}$, 
for training the explainer $\Psi(\cdot)$.

\subsection{Derivation of Inner Optimization}
\label{sec:inner}
For the inner optimization objective, we implement the first constraint by minimizing the distribution distance between $Q_{\boldsymbol{\phi}}(\tilde{G}|G')$ and $P_\gG$. Under the Erdős–Rényi assumption, the distribution loss is equivalent to the
cross-entropy loss between $\tilde{G}$ given $G'$ and $G$ over the full adjacency matrix~\cite{chen2023dexplainer}. Considering that $G$ is usually sparse 
rather than a fully connected graph, in practice, we adopt a weighted version to emphasize more connected node pairs~\cite{wang2016structural}. Formally, we have the following distribution loss.
\begin{equation}
    \gL_\text{dist} = \frac{\beta}{|\gE|} \sum_{(u,v) \in \gE} \log(\tilde{p}_{uv}) + \frac{1}{|\bar{\gE}|}\sum_{(u,v) \in \bar{\gE}} \log(1-\tilde{p}_{uv}),
\end{equation}
where $\bar{\gE}$ is the set of node pairs that are unconnected in $G$,  $\tilde{p}_{uv}$ is the probability of node pair $(u,v)$ in $\tilde{G}$, and $\beta$ is a hyper-parameter to for the trade-off between connected and unconnected node pairs.

The second constraint requires the mutual information of $Y$ and $\tilde{G}$ is the same as that of $Y$ and  $G'$. Due to the OOD problem, it is non-trivial to directly compute $P(Y|G')$ or $H(Y|G')$.  Instead, we implement this constraint with a novel graph generator that $\tilde{G}$ is obtained by combining $G'$ and a non-explanatory subgraph~\cite{zhang2023mixupexplainer}. 
In practice, we implement the non-explanatory part by perturbing the 
residual subgraph $G-G'$.  
The intuition is that if an explanation comprises label information, it is unlikely to change the prediction by manipulating the remaining non-explanatory part, which is widely adopted in the literature~\cite{fang2023evaluating,zheng2024towards}.

\begin{figure}
    \centering
    \includegraphics[width=0.5\textwidth]{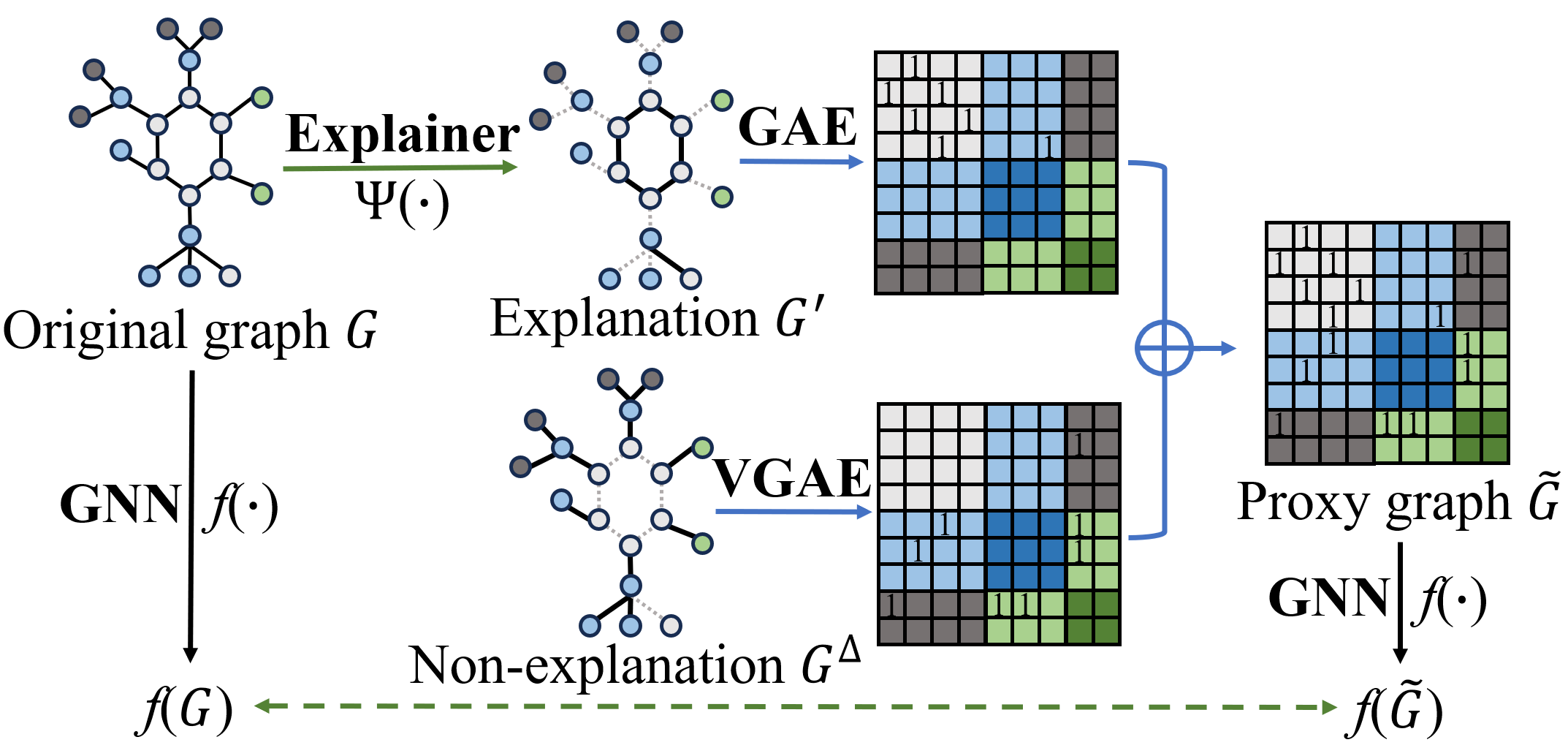}  
    \caption{ {\ours} consists of two components: an Explainer and a Proxy Graph Generator. The Explainer takes a graph $G$ as input and produces an explainable subgraph $G'$,  The proxy generator creates in-distribution proxy graphs that preserve the label information in $G'$. The proxy generator consists of a graph auto-encoder (GAE) and a variational graph auto-encoder (VGAE).}
    \label{fig:framework}
\end{figure}

\section{The {\ours}}
Based on our novel GIB with proxy graphs, in this section, we introduce a straightforward yet theoretically robust instantiation, named {\ours}. As shown in Figure \ref{fig:framework}, the architecture of our model comprises two key modules: the explainer and the proxy graph generator.  The explainer takes the original graph $G$ as its input and outputs a subgraph $G'$ as an explanation, which is optimized through the outer objective in~\eqref{eq:bi-level-opt}. The proxy graph generator generates an in-distributed proxy graph,  which is optimized with the inner objective. 

\subsection{The Explainer}
 For the sake of optimization, we follow the existing works to relax the element in $\mA$ from binary values to continuous values in the range $[0,1]$~\cite{luo2020parameterized,luo2024towards}. We adopt a generative explainer due to its effectiveness and efficiency~\cite{chen2023generative}. Specifically, we decompose the to-be-explained model $f(\cdot)$ into two functions that $f_\text{enc}(\cdot)$ learns node representations and $f_\text{cls}(\cdot)$ predicts graph labels based on node embeddings. Formally, we have $\mZ = f_\text{enc}(\mX,\mA)$ and $\hat{Y}=f_\text{cls}(\mZ)$, where $\mZ$ is the node representation matrix and $\hat{Y}$ is prediction. Routinely, $f_\text{cls}(\cdot)$ consists of a pooling layer followed by a classification layer and $f_\text{enc}(\cdot)$ consists of the other layers. Following the independence assumption in Section~\ref{sec:outter}, we approximate Bernoulli distributions with binary concrete distributions~\cite{maddison2017the,jang2017categorical}. Specifically, the probability of sampling an edge $(u,v)$ is computed by an MLP parameterized by $\psi$, denoted by $g_{\psi}(\cdot)$. Formally, we have
\begin{equation}
    \begin{aligned}
     &    \omega_{uv} = g_\psi([\vz_u;\vz_v]), \\
     & \epsilon \sim \text{Uniform}(0,1),\\
    &    \emA_{uv}' = \sigma((\log\epsilon - \log(1-\epsilon) + \omega_{uv})/\tau),
    \end{aligned}
\end{equation}
where $[\vz_u;\vz_v]$ is the concatenation of node representations $\vz_u$ and $\vz_v$, $\epsilon$ is a independent variable, $\sigma(\cdot)$ is the Sigmoid function, and $\tau$ is a temperature hyper-parameter for approximation. According to~\cite{luo2020parameterized}, the parameter $\pi'_{uv}$ of Bernoulli distribution in \eqref{eq:igg} can be obtained by $\pi'_{uv} =\frac{\exp(\omega_{uv})}{1+\exp(\omega_{uv})}$.

\subsection{The Proxy Graph Generator}
As shown in our analysis in Section~\ref{sec:inner},  we demonstrate the synthesis of a proxy graph through the amalgamation of the explanation subgraph $G'$ and the perturbation of its non-explanatory subgraph, represented as $G^{\Delta}=(\gV,\gE^\Delta,\mX)$. We define the edge set of $G^{\Delta}, \gE^\Delta$, as the differential set $\gE - \gE'$. Correspondingly, the adjacency matrix $\mA^\Delta$ is derived through $\mA -\mA'$. Building upon this foundation, and as illustrated in Figure~\ref{fig:framework}, our proposed framework introduces a dual-structured mechanism comprising two distinct graph auto-encoders(GAE)~\cite{bank2023autoencoders}. To be more specific, we first include an encoder to learn a latent matrix $\mZ'$ according to $\mA'$ and $\mX'$, and a decoder network recovers $\mA'$ based on $\mZ'$. Formally, we have:
\begin{equation}
    \label{eq:autoencoder1}
    \mZ' = \text{ENC}_1(\mA',\mX), \quad \tilde{\mA}' = \text{DEC}(\mZ'),
\end{equation}
where $\text{ENC}_1$ and $\text{DEC}$ are the encoder network and decoder network. Our framework is flexible to the choices of these two networks. $\tilde{\mA}' \in \sR^{n\times n}$ is the reconstructed adjacency matrix. 

To introduce perturbations into the non-explanatory segment, our approach employs a Variational Graph Auto-Encoder (VGAE), a generative model adept at creating varied yet structurally coherent graph data. This capability is pivotal in generating nuanced variations of the non-explanatory subgraph.
The VGAE operates by first encoding the non-explanatory subgraph $G^\Delta$ into a latent probabilistic space, characterized by Gaussian distributions. This process is articulated as:
\begin{equation}
    \label{eq:encoder2}
    \begin{aligned}
        \bm{\mu}^\Delta = \text{ENC}_1(\mA^\Delta,\mX), \quad \bm{\sigma}^\Delta = \text{ENC}_2(\mA^\Delta,\mX),
\end{aligned}   
\end{equation}
where $\text{ENC}_1$  and $\text{ENC}_2$ are encoder networks that learn the mean $\bm{\mu}^\Delta$ and variance $\bm{\sigma}^\Delta$ of the Gaussian distributions, respectively.
Following this, the latent representations $\mZ^\Delta$ are sampled from these distributions, ensuring that each generated instance is a unique variation of the original. The decoder network then reconstructs the perturbed non-explanatory subgraph from these sampled latent representations:
\begin{equation}
    \label{eq:decoder2}
    \begin{aligned}
        \mZ^\Delta\sim \mathcal{N}(\bm{\mu}^\Delta,\text{diag}(\bm{\sigma}^\Delta)^2), \quad \tilde{\mA}^\Delta = \text{DEC}(\mZ^\Delta),
\end{aligned}   
\end{equation}
where $\tilde{\mA}^\Delta \in \sR^{n\times n}$ represents the adjacency matrix of the perturbed non-explanatory subgraph. This novel use of VGAE facilitates the generation of diverse yet representative perturbations, crucial for enhancing the interpretability of explainers in our proxy graph framework. The adjacency matrix of a proxy graph is then obtained by 
\begin{equation}
    \tilde{\mA} = \tilde{\mA}' + \tilde{\mA}^\Delta.
\end{equation}

\noindent \textbf{Loss function.}
To train the proxy graph generator, we introduce a standard Gaussian distribution as the prior for the latent space representations in the VGAE, specifically $\mZ \sim \mathcal{N}(0,\mathcal{I})$, where $\mathcal{I}$ represents the identity matrix.  Then, the loss function is as follows. 

\begin{equation}
\label{eq:proxyloss}
    \mathcal{L}_\text{proxy} = \mathcal{L}_\text{dist} + \lambda \mathcal{L}_\text{KL}
    ,
\end{equation}
where $\mathcal{L}_\text{dist}$ is equivalent to cross-entropy between $\tilde{G}$ and $G$~\cite{chen2023dexplainer}. $\mathcal{L}_\text{KL}$ represents the KL divergence between the distribution of the latent representations $\mZ^{\Delta}$ and the assumed Gaussian prior. This term is crucial for regulating the variational aspect of the VGAE, ensuring that the generated perturbations are meaningful and controlled. 
$\lambda$ is a hyper-parameter.

\stitle{Alternate Training.} 
To train the explainer and the proxy graph generator networks, we follow existing works~\citep{zheng2024towards} to use an alternate training schedule that trains the proxy graph generator network $M$ times and then trains the explainer network 
once. $M$ is a hyper-parameter determined by grid search.  The detailed algorithm description of our model is shown in Appendix \ref{sec:algorithm}.

\section{Related Work}
\noindent \textbf{GNN Explanation.} The goal of explainability in GNNs is to ensure transparency in graph-based tasks. Recent works have been directed towards elucidating the rationale behind GNN predictions. These explanation methods can be broadly classified into two categories: instance-level and model-level approaches \cite{yuan2022explainability}. In this study, we focus on instance-level explanations, which aim to clarify the specific reasoning behind individual predictions made by GNNs. These methods are critical for understanding the decision-making process on a case-by-case basis to enhance the explainability of GNNs. For example, GNNExplainer \cite{ying2019gnnexplainer} excludes certain edges and node features to observe the changes in prediction. However, its single-instance focus limits its applicability to provide a global understanding of the to-be-explained model~\cite{chen2023generative}. PGExplainer \cite{luo2020parameterized,luo2024towards} introduces a parametric neural network to learn edge weights. Thus, once training is complete, it can explain new graphs without retraining. ReFine \cite{wang2021towards} integrates a pre-training phase that focuses on class comparisons and is fine-tuned to refine context-specific explanations. GStarX \cite{zhang2022gstarx} assigns an importance score to each node by caculating the Hamiache and Navarro values of the structure to obtain explanatory subgraphs. GFlowExplainer \cite{li2023dag} uses a generator to construct a TD-like flow matching condition to learn a policy for generating explanations by adding nodes sequentially.

\noindent \textbf{Distribution Shifting in Explanations.} The distribution shifting problem in post-hoc explanations has been increasingly recognized in explainable AI fields~\cite{chang2018explaining,qiu2022generating}. For example, FIDO~\cite{chang2018explaining} works on enhancing image classifier explanations, focusing on relevant contextual details that agree with the training data's distribution. A recent study tackles the distribution shifting problem in image explanations by introducing a module that assesses the similarity between altered data and the original dataset distribution~\cite{qiu2022generating}. In the graph domain, an ad-hoc strategy to mitigate distribution shifting is to initially reduce the size constraint coefficient during the explanation process~\cite{fang2022regularization}. MixupExplainer~\cite{zhang2023mixupexplainer} and RegExplainer~\cite{zhang2023regexplainer} propose non-parametric solutions by mixing up the explanation subgraph with a non-explainable part from another graph. However, these methods operates under the assumption that the explanatory and non-explanatory subgraphs in mixed graphs are independent, which may not hold in many real-life graphs.
\section{Experiments}
\label{sec:exp}

\begin{table*}[ht!]
    \centering
    \caption{Explanation accuracy in terms of AUC-ROC on edges.}
    \begin{tabular}{p{2.2cm} C{2cm} C{2cm} C{2cm}C{2cm}C{2cm}C{2cm}}  
    \toprule
    & {\mutag} & {\benz} & {\alk}  & {\fluo}  & {\bamo} & {\bathree}  \\ \midrule
    GradCAM   & 0.727$_{\pm 0.000}$ & 0.740$_{\pm 0.000}$ & 0.448$_{\pm 0.000}$  & 0.694$_{\pm 0.000}$  & 0.714$_{\pm 0.000}$  & 0.709$_{\pm 0.000}$   \\  
    GNNExplainer & 0.682$_{\pm 0.009}$ & 0.485$_{\pm 0.001}$ & 0.551$_{\pm 0.003}$ & 0.574$_{\pm 0.002}$  & 0.644$_{\pm 0.007}$ & 0.511$_{\pm 0.002}$  \\  
    PGExplainer &  0.832$_{\pm 0.032}$ & 0.793$_{\pm 0.054}$ & 0.660$_{\pm 0.036}$  & 0.702$_{\pm 0.018}$  & 0.734$_{\pm 0.117}$  & 0.796$_{\pm 0.010}$ \\  
    ReFine   & 0.612$_{\pm 0.004}$ & 0.606$_{\pm 0.002}$ & 0.768$_{\pm 0.001}$ & 0.571$_{\pm 0.000}$ & 0.698$_{\pm 0.001}$ & 0.629$_{\pm 0.005}$  \\ 
    MixupExplainer  & 0.863$_{\pm 0.103}$ & 0.611$_{\pm 0.032}$ & 0.811$_{\pm 0.006}$  & 0.706$_{\pm 0.013}$ & 0.906$_{\pm 0.059}$ & 0.859$_{\pm 0.019}$   \\ 
    {\ours}  & \textbf{0.977}$_{\pm 0.009}$ & \textbf{0.845}$_{\pm 0.036}$ & \textbf{0.934}$_{\pm 0.005}$  & \textbf{0.758}$_{\pm 0.068}$ &  \textbf{0.935}$_{\pm 0.008}$ & \textbf{0.960}$_{\pm 0.008}$  \\ 
    \bottomrule
    \end{tabular}
    \label{tab:aucresults_gcn}
\end{table*}

\begin{table}[h!]
    \centering
    \caption{Explanation accuracy in terms of AP on {\mutag} and {\bamo}.}
    \begin{tabular}{p{2.3cm} C{2.3cm} C{2.3cm}}  
    \toprule
    & {\mutag} & {\bamo}  \\ \midrule
    GradCAM   &$0.247_{\pm 0.000}$ & $0.664_{\pm 0.000}$   \\  
    GNNExplainer &$0.232_{\pm 0.001}$ & $0.608_{\pm 0.004}$  \\  
    PGExplainer &$0.611_{\pm 0.024}$ & $0.682_{\pm 0.117}$  \\  
    ReFine   &$0.227_{\pm 0.001}$ & $0.619_{\pm 0.002}$   \\ 
    MixupExplainer  &$0.647_{\pm 0.083}$ & $0.787_{\pm 0.073}$    \\ 
    {\ours}  & $\textbf{0.756}_{\pm 0.107}$ & $\textbf{0.839}_{\pm 0.036}$   \\ 
    \bottomrule
    \end{tabular}
    \label{tab:apresults_gcn}
\end{table}

\begin{table}[h!]
    \centering
    \fontsize{8}{9}\selectfont 
    \setlength\tabcolsep{3pt}
    \caption{Fidelity evaluation on {\mutag} and {\bamo}.}
    \setlength\tabcolsep{3.3pt}
    \scalebox{0.95}{
    \begin{tabular}{p{1.4cm} C{1.5cm} C{1.5cm} | C{1.5cm} C{1.5cm}}  
    \toprule
    & \multicolumn{2}{c|}{\mutag} & \multicolumn{2}{c}{\bamo}  \\ 
    & $Fid_{\alpha_1,{+}} \uparrow$ & $Fid_{\alpha_2,{-}} \downarrow$ & $Fid_{\alpha_1,{+}} \uparrow$ & $Fid_{\alpha_2,{-}} \downarrow$ \\
    \midrule
    GradCAM   & 0.004$_{\pm 0.000}$ & $0.162_{\pm 0.000}$ & $0.072_{\pm 0.000}$ & $0.107_{\pm 0.000}$  \\  
    GNNExp. & $0.031_{\pm 0.001}$ & $0.148_{\pm 0.001}$ & $0.057_{\pm 0.002}$ & $0.132_{\pm 0.001}$  \\  
    PGExp. & $0.034_{\pm 0.011}$ & $0.148_{\pm 0.005}$ & $0.065_{\pm 0.017}$ & $0.126_{\pm 0.009}$ \\  
    ReFine   & $0.003_{\pm 0.000}$ & $0.160_{\pm 0.001}$ & $0.060_{\pm 0.005}$ & $0.125_{\pm 0.001}$  \\ 
    MixupExp.  & $0.037_{\pm 0.006}$ & $0.146_{\pm 0.003}$ & $0.074_{\pm 0.005}$ & $0.112_{\pm 0.003}$ \\ 
    ProxyExp.  & $\textbf{0.040}_{\pm 0.002}$ & $\textbf{0.145}_{\pm 0.001}$ & $\textbf{0.086}_{\pm 0.003}$ & $\textbf{0.106}_{\pm 0.002}$ \\ 
    \bottomrule
    \end{tabular}
    \label{tab:fidresults_gcn}
    }
\end{table}

We present empirical results that illustrate the effectiveness of our proposed method. These experiments are mainly designed to explore the following research questions:

\begin{itemize}[leftmargin=*]
    \item \textbf{RQ1:} Can the proposed framework outperform other baselines in identifying explanations for GNNs?
    \item \textbf{RQ2:} Is the distribution shifting severe in explanation subgraphs? Can the proposed approach alleviate that?
    \item \textbf{RQ3:} How does each component of {\ours} impact the overall performance in generating explanations? 
\end{itemize}

\subsection{Experimental Settings}

To evaluate the performance of {\ours}, we use six benchmark datasets with ground-truth explanations. These include four real-world datasets: {\mutag} \cite{kazius2005derivation}, {\benz} \cite{sanchez2020evaluating},{\alk} \cite{sanchez2020evaluating},  and {\fluo} \cite{sanchez2020evaluating}, along with two synthetic datasets: {\bamo}~\cite{luo2020parameterized} and {\bathree} \cite{chen2023dexplainer}.  We take GradCAM~\cite{pope2019explainability}, GNNExplainer\cite{ying2019gnnexplainer}, PGExplainer\cite{luo2020parameterized}, ReFine~\cite{wang2021towards}, and MixupExplainer~\cite{zhang2023mixupexplainer} for comparison. We follow the experimental setting in previous works \cite{ying2019gnnexplainer, luo2020parameterized, sanchez2020evaluating} to train a Graph Convolutional Network (GCN) model~\cite{kipf2017semisupervised} with three layers. 
Experiments on another representative GNN, Graph Isomorphism Network (GIN)~\cite{xu2018powerful}, can be found in Appendix \ref{sec:ginexplain}. We use the Adam optimizer~\cite{kingma2014adam} with the inclusion of a weight decay $5e-4$. Detailed information regarding datasets and baselines is delineated in Appendix \ref{sec:datasetandbaselines}.

To evaluate the quality of explanations, we approach the explanation task as a binary classification of edges. Edges that are part of ground truth subgraphs are labeled as positive, while all others are deemed negative. The importance weights given by the explanation methods are interpreted as prediction scores. An effective explanation technique is one that assigns higher weights to edges within the ground truth subgraphs compared to those outside of them. We utilize the AUC-ROC for quantitative assessment~\cite{ying2019gnnexplainer,luo2020parameterized}.

\subsection{Quantitative Evaluation (RQ1)}
To answer RQ1, we compare the proposed method, {\ours}, to other baselines. Each experiment was conducted 10 times using random seeds, and the average AUC scores as well as standard deviations are presented in Table \ref{tab:aucresults_gcn}.

The results demonstrate that {\ours} provides the most accurate explanations across all datasets. Specifically, it improves the AUC scores by an average of $10.6\%$ on real-world datasets and $7.5\%$ on synthetic datasets over the leading baselines. Comparisons with baseline methods highlight the advantages of our proposed explanation framework. Besides, {\ours} captures underlying explanatory factors consistently across diverse datasets. For instance, MixupExplainer exhibits proficiency on the synthetic {\bamo} dataset but performs poorly on the real-world {\benz} dataset. The reason is that MixupExplainer relies on the independence assumption of explanation and non-explanation subgraphs, which may not hold in real-world datasets. In contrast, {\ours} consistently demonstrates high performance across different datasets, showcasing its robustness and adaptability.

Considering the importance of precision for the positive class in our context, we further adopt AP to evaluate the performance. As shown in Table \ref{tab:apresults_gcn}, with AP scores, {\ours} consistently outperforms the other baselines in two benchmark datasets {\mutag} and {\bamo}. 

Additionally, some existing works, such as SubgraphX~\cite{yuan2021explainability}, adopt faithfulness-based metrics for evaluation. However, these metrics are problematic due to the OOD problem~\cite{zheng2024towards,amara2023ginx}. So we use the robust fidelity metrics ($Fid_{\alpha_1,{+}}$, $Fid_{\alpha_2,{-}}$) as described in \cite{zheng2024towards} with default settings ($\alpha_1=0.1, \alpha_2=0.9$) to evaluate model faithfulness. Table \ref{tab:fidresults_gcn} demonstrates that {\ours} is consistently outperforms all baselines in both $Fid_{\alpha_1,{+}}$ and $Fid_{\alpha_2,{-}}$.

\subsection{Alleviating Distribution Shifts (RQ2)}
In this section, we assess {\ours}'s ability to generate in-distribution proxy graphs.  Due to the intractable of direct computation, we follow the previous work~\cite{chen2023dexplainer} to utilize Maximum Mean Discrepancy (MMD) between distributions of multiple graph statistics, including degree distributions, clustering coefficients, and spectrum distributions, between the generated proxy graphs and original graphs. Specifically, we utilize Gaussian Earth Mover’s Distance kernel when computing MMDs. Smaller MMD values indicate similar graph distributions. For comparison, we also include the Ground truth explanations and the ones generated by PGExplainer.

\begin{table}[!tp]
\centering
\fontsize{8}{9}\selectfont  
\setlength\tabcolsep{3pt}
\caption{MMD results between the ground truth explanations and original graphs (GT); PGExplainer explanations and original graphs (PGE); proxy graphs in our methods and original graphs (Proxy).}
\label{tab:mmd}
\setlength\tabcolsep{3.3pt}
\scalebox{0.92}{
\begin{tabular}{l|ccc|ccc|ccc}
\toprule
 & \multicolumn{3}{c|}{\mutag}  & \multicolumn{3}{c|}{\benz} & \multicolumn{3}{c}{\alk} \\
\midrule
Metric & GT & PGE & Proxy & GT & PGE & Proxy & GT & PGE & Proxy \\
\midrule
\textit{Deg.} & 0.614 & 0.468 & \textbf{0.123}  & 0.843 & 0.393 & \textbf{0.236}  & 0.872 & 0.665 & \textbf{0.177} \\
\textit{Clus.} & \textbf{0.003} & \textbf{0.003} & 0.009  & 0.009 & \textbf{0.002} & 0.004 & \textbf{0.011} & \textbf{0.011} & \textbf{0.011} \\
\textit{Spec.} & 0.414 & 0.341 & \textbf{0.186}  & 0.295 & 0.163 & \textbf{0.101}  & 0.596 & 0.447 & \textbf{0.049} \\
\textit{Sum.} & 1.032 & 0.813 & \textbf{0.317}  & 1.147 & 0.558 & \textbf{0.341}  & 1.479 & 1.123 & \textbf{0.237} \\
\midrule
 & \multicolumn{3}{c|}{\fluo} & \multicolumn{3}{c|}{\bamo} & \multicolumn{3}{c}{\bathree} \\
\midrule
 Metric & GT & PGE & Proxy & GT & PGE & Proxy & GT & PGE & Proxy \\
\midrule
\textit{Deg.} & 0.638 & 0.488 & \textbf{0.196} & 0.759 & 0.496 & \textbf{0.060} & 0.541 & 0.149 & \textbf{0.092} \\
\textit{Clus.}   & \textbf{0.012} & \textbf{0.012} & \textbf{0.012} &\textbf{0.447} & 0.463 & 0.584 & 0.262 & 0.382 & \textbf{0.245} \\
\textit{Spec.}  & 0.351 & 0.315 & \textbf{0.100} & 0.245 & 0.256 & \textbf{0.091} & 0.217 & 0.063 & \textbf{0.062} \\
\textit{Sum.}   & 1.000 & 0.815 & \textbf{0.308}  & 1.451 & 1.215 & \textbf{0.735} & 1.020 & 0.594 & \textbf{0.399} \\
\bottomrule
\end{tabular}
}
\end{table}

The results are shown in Table \ref{tab:mmd}. ``GT'' denotes the MMDs between the ground truth explanations and original graphs. ``PGE'' represents the MMDs between explanations generated by PGExplainer and original graphs. ``Proxy'' denotes the MMDs between proxy graphs in our method and original graphs. We have the following observations. First, the MMDs between ground truth and original graphs are usually large, verifying our motivation that a model trained on original graphs may not have correct predictions on the OOD explanation subgraphs. Second, the explanations generated by a representative work, PGExplainer, are often OOD from original graphs, indicating that the original GIB-based objective function may be sub-optimal. Third, in most cases, proxy graphs generated by our method are with smaller MMDs, demonstrating their in-distribution property.

\subsection{Ablation Studies (RQ3)}
\begin{figure}
    \centering
    \includegraphics[width=0.45\textwidth]{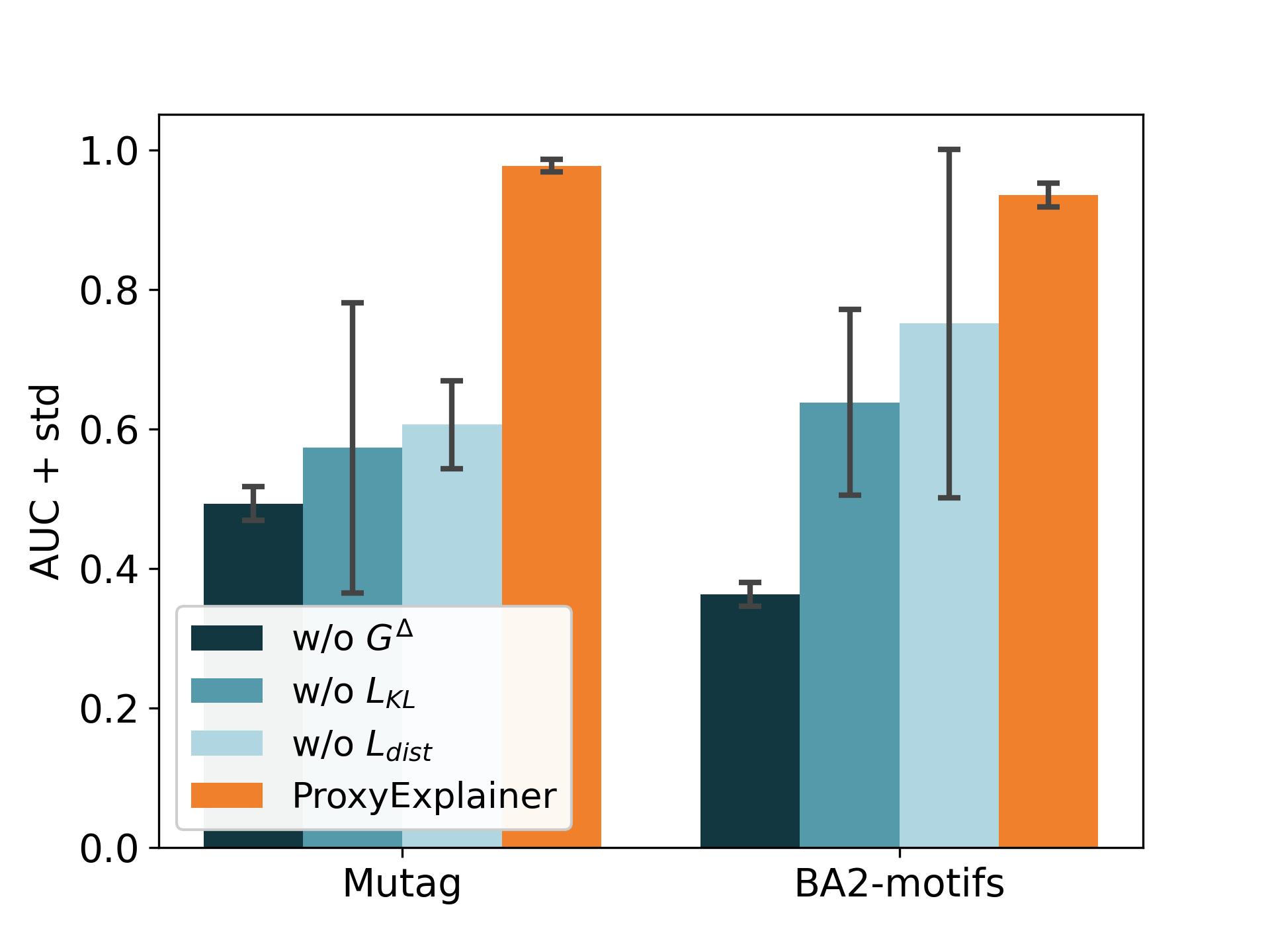}  
    \caption{Ablation Studies on {\mutag} and {\bamo}.}
    \label{fig:ablation_sample}
\end{figure}

In this section, we conduct ablation studies to investigate the roles of different components. Specifically, we consider the following variants of {\ours}: (1) w/o $G^\Delta$: in this variant, we remove the non-explanatory subgraph generator (VGAE), which is the bottom half as shown in Figure~\ref{fig:framework};
(2) w/o $\mathcal{L}_{\text{KL}}$: in this variant, we remove the KL divergence from the training loss in {\ours}; (3) w/o $\mathcal{L}_{\text{dist}}$: in this variant, we remove the distribution loss from {\ours}. The results of the ablation study on {\mutag} and {\bamo} are reported in Figure~\ref{fig:ablation_sample}. 

Figure~\ref{fig:ablation_sample} illustrates a notable performance drop for all variants, indicating that each component contributes positively to the effectiveness of \ours. 
Especially, in the real-life dataset {\mutag}, without the in-distribution constraint, w/o $\mathcal{L}_{\text{dist}}$ is much worse than {\ours}, indicating the vital role of in-distributed proxy graphs in our framework. Extensive ablation studies on other datasets can be found in the Appendix \ref{sec:extensiveablation}.

\begin{figure*}[h]
    \centering
    \begin{subfigure}[b]{0.19\textwidth}
        \centering
        \includegraphics[width=\textwidth]{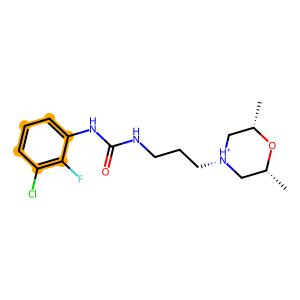}
        \caption{Ground Truth}
    \end{subfigure}
    \begin{subfigure}[b]{0.19\textwidth}
        \centering
        \includegraphics[width=\textwidth]{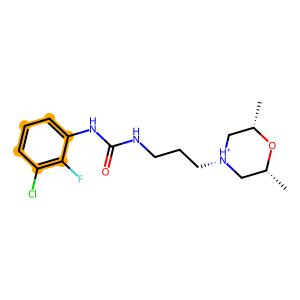}
        \caption{{\ours}}
    \end{subfigure}
    \begin{subfigure}[b]{0.19\textwidth}
        \centering
        \includegraphics[width=\textwidth]{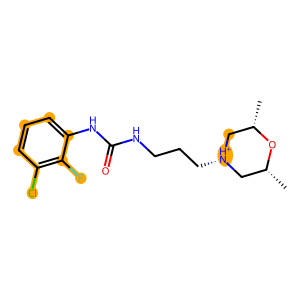}
        \caption{MixupExplainer}
    \end{subfigure}
    \begin{subfigure}[b]{0.19\textwidth}
        \centering
        \includegraphics[width=\textwidth]{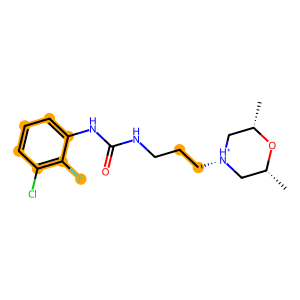}
        \caption{PGExplainer}
    \end{subfigure}
    \begin{subfigure}[b]{0.19\textwidth}
        \centering
        \includegraphics[width=\textwidth]{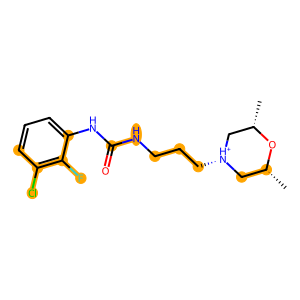}
        \caption{GNNExplainer}
    \end{subfigure}
    \caption{Visualization of explanation results from different explanation models on {\benz}. The generated explanations are highlighted with bold orange edges.}
    \label{fig:case study on ben}
\end{figure*}

\begin{figure*}[h]
    \centering
    \begin{subfigure}[b]{0.19\textwidth}
        \centering
        \includegraphics[width=\textwidth]{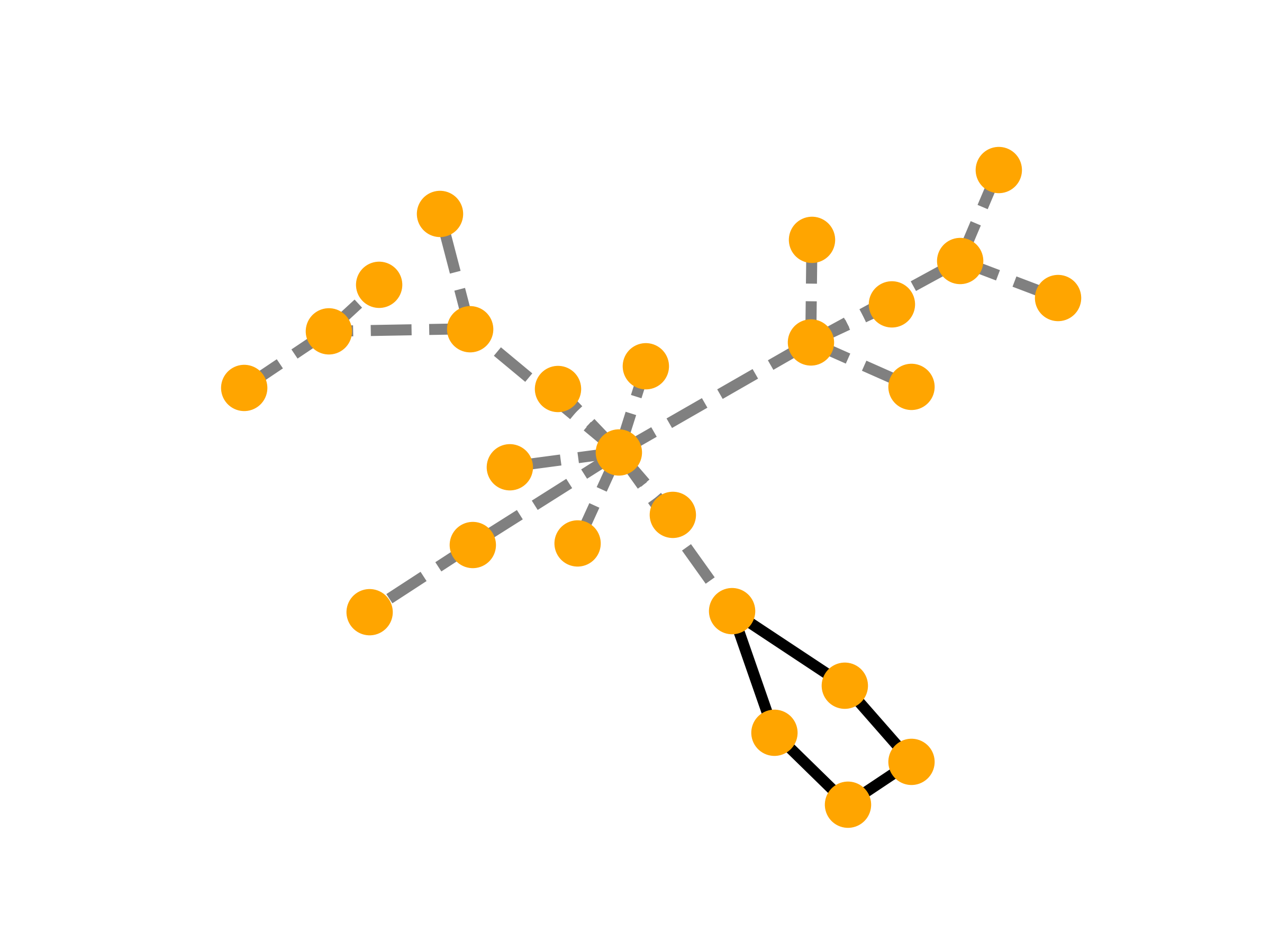}
        \caption{Ground Truth}
    \end{subfigure}
    \begin{subfigure}[b]{0.19\textwidth}
        \centering
        \includegraphics[width=\textwidth]{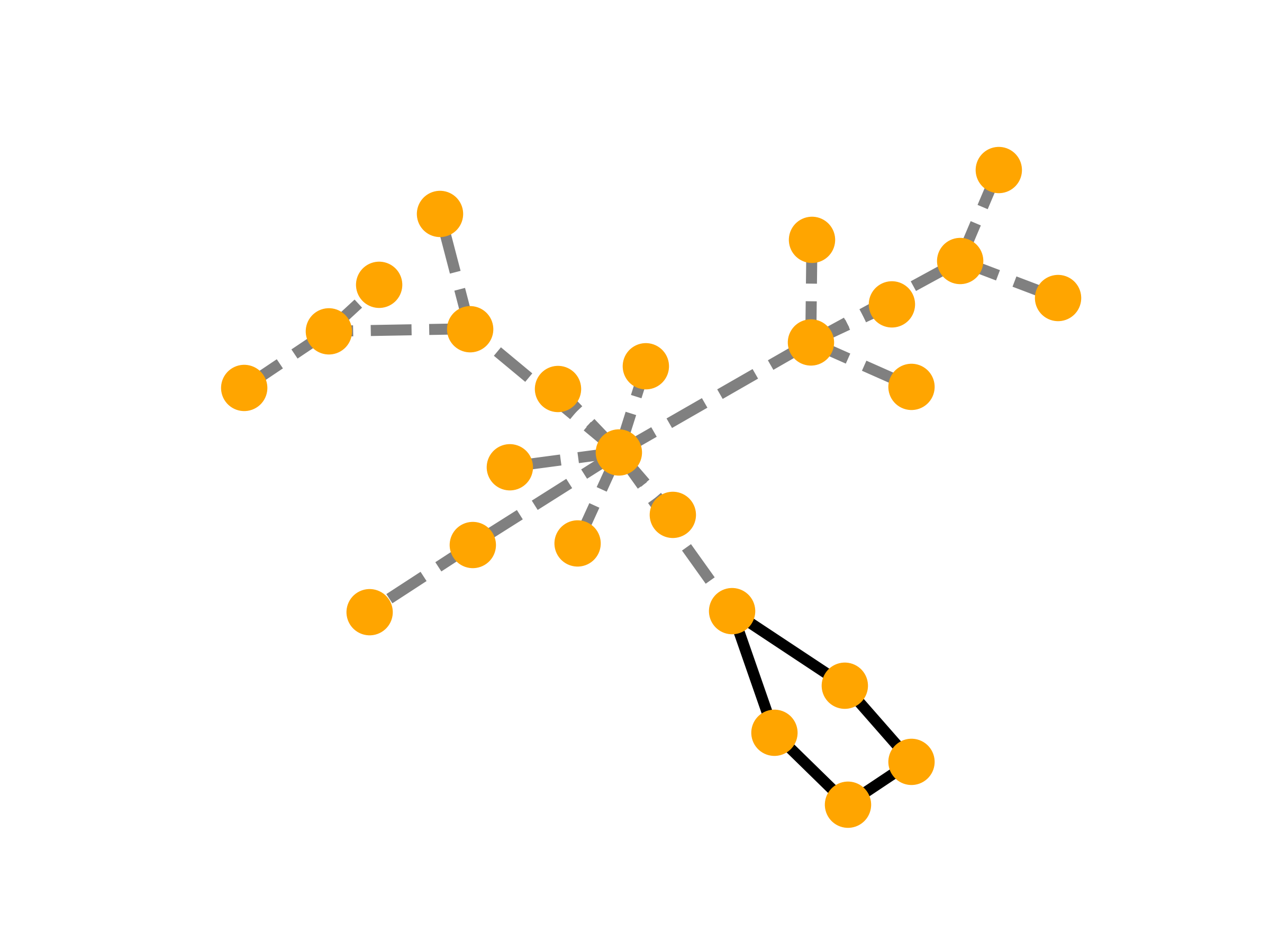}
        \caption{{\ours}}
    \end{subfigure}
    \begin{subfigure}[b]{0.19\textwidth}
        \centering
        \includegraphics[width=\textwidth]{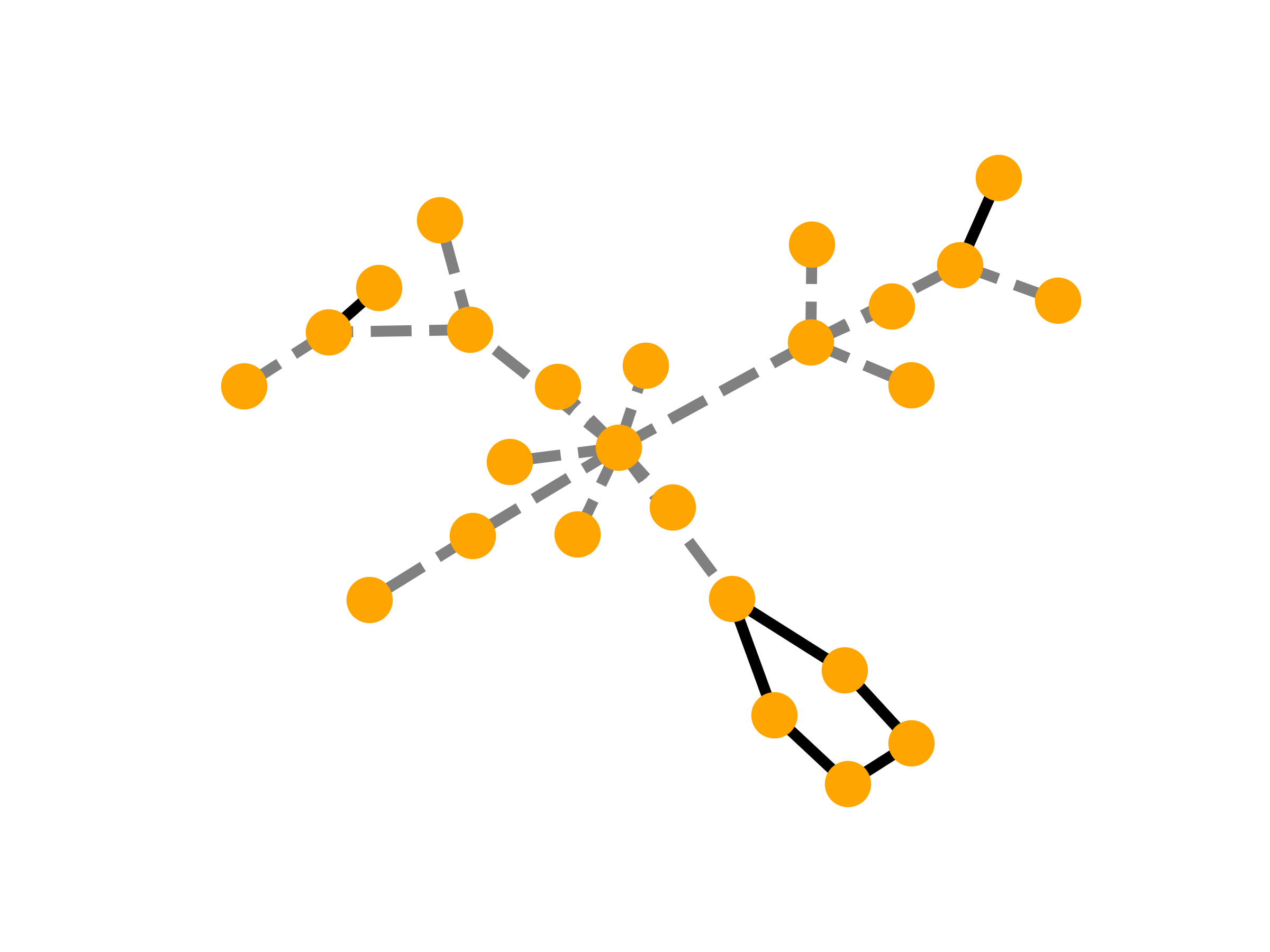}
        \caption{MixupExplainer}
    \end{subfigure}
    \begin{subfigure}[b]{0.19\textwidth}
        \centering
        \includegraphics[width=\textwidth]{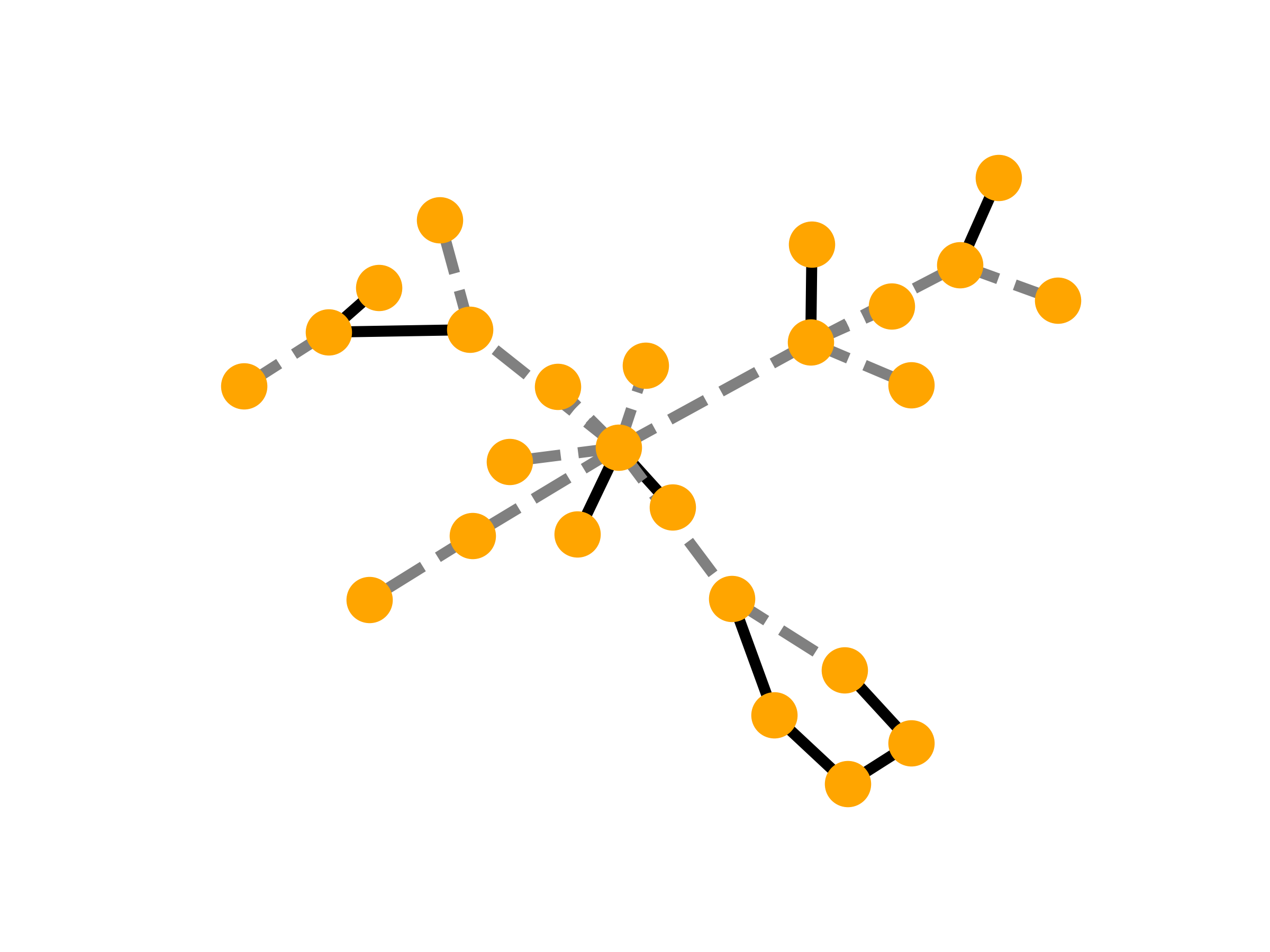}
        \caption{PGExplainer}
    \end{subfigure}
    \begin{subfigure}[b]{0.19\textwidth}
        \centering
        \includegraphics[width=\textwidth]{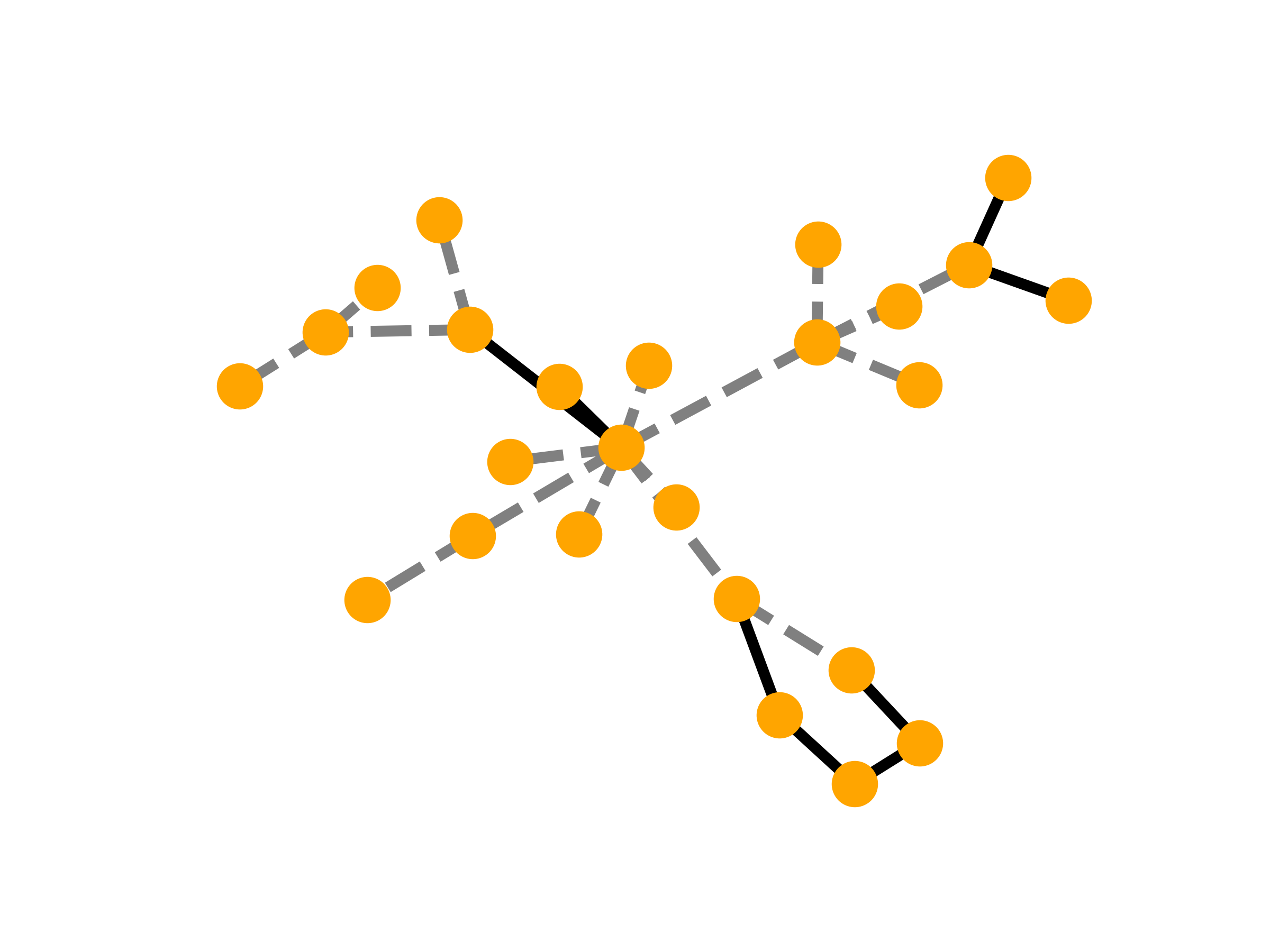}
        \caption{GNNExplainer}
    \end{subfigure}
    \caption{Visualization of explanation results from different explanation models on {\bamo}. The generated explanations are highlighted with bold black edges.}
    \label{fig:case study on {\bamo}}
\end{figure*}

\subsection{Case Studies}
In this part, we conduct case studies to qualitatively compare the performances of {\ours} against others. We adopt examples from {\benz} and {\bamo} in this part. We show visualization results in Figure \ref{fig:case study on ben} and Figure \ref{fig:case study on {\bamo}}. Explanations are highlighted with bold black and bold orange edges, respectively. From the results, our {\ours} stands out by generating more compelling explanations compared to baselines. Specifically, {\ours} maintains clarity without introducing irrelevant edges and exhibits more concise results compared to alternative methods. The visualized performance underscores {\ours}'s ability to provide meaningful and focused subgraph explanations. 
More visualizations on these two datasets can be found in the Appendix \ref{sec:extensivecase}.
\vspace{-1mm}

\section{Conclusion}

In this paper, we systematically investigate the OOD problem in the de facto framework, GIB, for learning explanations in GNNs, which is highly overlooked in the literature. To address this issue, we extend the GIB by innovatively including in-distributed proxy graphs. On top of that, we derive a tractable objective function for practical implementations.  We further present a new explanation framework that utilizes two graph auto-encoders to generate proxy graphs. We conduct comprehensive empirical studies on both benchmark synthetic datasets and real-life datasets to demonstrate the effectiveness of our method in alleviating the OOD problem and achieving high-quality explanations. There are several topics we will investigate in the future. First, the OOD problem also exists in obtaining model-level explanations and counterfactual explanations. We will extend our method to these research problems. Second, we will also analyze explainable learning methods with proxies in other data structures, such as image, language, and time series. 

\section*{Impact Statement}
This paper presents work whose goal is to advance the field of Machine Learning. There are many potential societal consequences of our work, none of which we feel must be specifically highlighted here.

\section*{Acknowledgments}
This project was partially supported by NSF grant IIS-2331908. The views and conclusions contained in this paper are those of the authors and should not be interpreted as representing any funding agencies.

\bibliography{reference}
\bibliographystyle{icml2024}

\appendix
\onecolumn

\section{Notations}
In Table \ref{tab:notation}, we summarized the main notations we used and their descriptions in this paper.
\begin{table*}[h]
    \centering
    \caption{Symbols and their descriptions.}
    \begin{tabular}{C{3cm} p{12cm}}
    \hline
    Symbols & Descriptions \\
    \hline
    $\mathcal{G}$ & A set of graphs\\
    $G, \gV,\gE$ & Graph instance, node set, edge set \\
    $v_i$ & The $i$-th node \\
    $\mX$ & Node feature matrix \\
    $\mA$ & Adjacency matrix \\
    $\mZ$ & Node representation matrix \\
    $Y$ & Label of graph $G$ \\
    $\mathcal{Y}$ & A set of labels \\
    $G^*$ & Optimal explanatory subgraph \\
    $\gG^*$ & A set of $G^*$ \\
    $G'$ & Candidate explanatory subgraph \\
    $G^{\Delta}$ & Non-explanatory graph\\
    $\tilde{G}$ & Proxy graph of $G'$ with a fixed distribution \\
    $\textit{\textbf{h}}$, $\textit{\textbf{h}}'$, $\tilde{\textit{\textbf{h}}}$ & Graph embeddings\\
    $d$ & Dimension of node feature \\
    $f(\cdot)$ & To-be-explained GNN model \\
    $\Psi_\psi(\cdot)$ & Explanation function \\
    $\psi$ & Parameter of the explanation function \\
    $P_\gG$ & Distribution of original training graphs \\
    $P_{\gG'}$ & Distribution of explanation subgraphs \\ 
    $Q_{\boldsymbol{\phi}}$ & Parameterized function of $P(\tilde{G}|G')$ \\
    $ \boldsymbol{\phi}$ & Model parameters of $Q_{\boldsymbol{\phi}}$\\
    $\boldsymbol{\phi}^*$ &  Optimal $\boldsymbol{\phi}$ \\
    $\alpha$ & Balance parameter between $I(G, G')$ and $I(Y, G')$ \\
    $\bar{\gE}$ & The set of node pairs that are unconnected in $G$\\ $\tilde{p}_{uv}$ & Probability of node pair $(u,v)$ in $\tilde{G}$\\
    $\beta$ & A hyper-parameter to get a trade-off between connected and unconnected node pairs\\
    $f_\text{enc}(\cdot)$ & The front part of GNNs that learns node representations \\
    $f_\text{cls}(\cdot)$ & The back part of GNNs that predicts graph labels based on node embeddings \\
    $\sigma(\cdot)$ & Sigmoid function\\ 
    $\tau$ & Temperature hyper-parameter for approximation\\
    $\lambda$ & A hyper-parameter in Proxy loss function\\
    $\mathcal{L}_\text{dist}$ & Distribution loss between $\tilde{G}$ and $G$ \\
    $\mathcal{L}_\text{KL}$ & KL divergence between distribution of $\mZ^{\Delta}$ and its prior\\
    $\mathcal{L}_\text{proxy}$ & Proxy loss \\
    $\mathcal{L}_\text{exp}$ & Explainer loss\\
    \hline
    \end{tabular}
    \label{tab:notation}
\end{table*}

\section{Algorithm} \label{sec:algorithm}
We take as input a set of graphs $\mathcal{G} =  \{ G_i\}_{i=0}^{N}$. 
For each graph $G_i$, we use an explainer model to identify a subgraph $G'_i$ as the explanation and then compute the non-explanatory graph $G^{\Delta}_i$, which is obtained by removing edges in $G_i$ that exist in $G'_i$. We use GAE to reconstruct a subgraph, denoted by $\tilde{G}' _i$ from the subgraph $G’ _i$ and another VGAE to generate a new subgraph, denoted by $\tilde{G}^{\Delta}_i$, from the non-explainable subgraph $G^{\Delta}_i$. The proxy graph, $\tilde{G}_i$, is obtained by combining them, whose adjacency matrix can be denoted by $\tilde{\mA} _i = \tilde{\mA}' _i +\tilde{\mA}^{\Delta}_i$. Here $\tilde{\mA}' _i$  and $\tilde{\mA}^{\Delta}_i$ are adjacency matrices of $\tilde{G}' _i$ and $\tilde{G}^{\Delta}_i$, respectively. We alternatively train the explainer model and proxy graph generator as shown in Algorithm~\ref{alg:model}.

\begin{algorithm}[tb]
    \caption{Algorithm of {\ours}}
    \label{alg:model}
\begin{algorithmic}
   \STATE {\bfseries Input:} A set of graphs $\mathcal{G} = \{G_i\}_{i=0}^N$, with each $G_i=(\gV_i,\gE_i,\mX_i), $a pretrain to-be-explained model $f(\cdot)$, hyper parameters $\alpha, \lambda, M$, epochs $E$.
       \STATE Initialize an explainer function $\Psi_\psi (\cdot)$
       \STATE $ \text{epoch} \gets 0 $ 
       \WHILE{epoch $< E$}
        \FOR{$G_i \in \mathcal{\mG}$}
            \STATE $G' _i \leftarrow \Psi_\psi(G_i)$ 
            \STATE $G^{\Delta } _i \leftarrow $ $G_i - G' _i$ 
            \STATE Compute $\tilde{\mA}'_i$ with~\eqref{eq:autoencoder1}
            \STATE Compute $\mZ^\Delta$ and $\tilde{\mA}^\Delta_i$ with~\eqref{eq:decoder2}
            \STATE Compute proxy loss $\mathcal{L}_\text{proxy}$ with~\eqref{eq:proxyloss}
            \STATE Update parameters in proxy graph generator with backpropagation
            \IF{ epoch$\%M == 0$}
                \STATE Compute explainer loss $\mathcal{L}_\text{exp}$
                \STATE Update parameters in the explainer with backpropagation
            \ENDIF
        \ENDFOR
        \STATE $ \text{epoch} \gets \text{epoch} + 1 $ 
        \ENDWHILE      
\end{algorithmic}
\end{algorithm}
\section{Full Experimental Setups}\label{sec:datasetandbaselines}

We conduct all experiments on a Linux machine with 8 Nvidia A100-PCIE GPUs, each with 40GB of memory. The CUDA version is 12.4 and the driver version is 550.54.15. We use Python 3.9 and Torch 2.0.1 in our project.  The code is available at \href{https://github.com/realMoana/ProxyExplainer}{https://github.com/realMoana/ProxyExplainer}.

\subsection{Datasets}

\textbf{{\mutag} \cite{kazius2005derivation}.}
{\mutag} dataset includes 4,337 molecular graphs, each of them is classified into two groups based on its mutagenic effect on the Gram-negative bacterium S. Typhimurium. This classification comes from several specific virulence gland groups with mutagenicity in molecular mapping by Kazius et al.\cite{kazius2005derivation}.\\
\textbf{{\benz} \cite{sanchez2020evaluating}.}
{\benz} consists 12,000 molecular graphs from the ZINC15\cite{teague2015zinc} database, which can be classified into two classes. The main goal is to determine if a {\benz} ring is exited in each molecule. In cases where multiple {\benz} rings are present, each {\benz} ring is treated as a distinct explanation. \\
\textbf{{\alk} \cite{sanchez2020evaluating}.}
The {\alk} dataset consists of a total of 4,326 molecule graphs that are categorized into two different classes. Positive samples refer to molecules that have both alkane and carbonyl functional groups. The ground-truth explanation includes alkane and carbonyl functional groups in a given molecule. \\
\textbf{{\fluo} \cite{sanchez2020evaluating}.}
The {\fluo} dataset has 8,671 molecular graphs. The ground-truth explanation is based on the particular combination of fluoride atoms and carbonyl functional groups present in each molecule.\\
\textbf{{\bamo} \cite{luo2020parameterized}.}
The {\bamo} dataset consists 1,000 synthetic graphs, each derived from a basic Barabasi-Albert (BA) model. The dataset is divided into two categories: one part of the graphs add patterns that mimic the structure of a house, and the remaining integrate five-node cyclic patterns. The classification of these graphs depends on the specific patterns. \\
\textbf{{\bathree} \cite{chen2023dexplainer}.}
{\bathree} is an extended dataset inspired by the {\bamo} and contains 3,000 synthetic graphs. Each base graph is accompanied by one of three different patterns: house, cycle, or grid.\\

\begin{table*}[h]
    \centering
    \caption{Statistics of molecule datasets for graph classification with ground-truth explanations.}
    \begin{tabular}{>{\centering\arraybackslash}p{2.5cm} C{2cm} C{2cm} C{2cm} C{2cm} C{2cm} C{2cm}}  
    \toprule
    & {\mutag} & {\benz} & {\alk} & {\fluo} & {\bamo}  & {\bathree}  \\ \midrule
    Graphs & 4,337 & 12,000 & 4,326  & 8,671 & 1,000  & 3,000  \\ 
    Average nodes & 29.15 & 20.58 & 21.13  & 21.36 & 25.00 & 21.92 \\ 
    Average edges & 60.83 & 43.65 & 44.95  & 45.37 & 25.48 & 29.51 \\ 
    Node features & 14 & 14 & 14 & 14 & 10 & 4  \\  
    Original graph & \includegraphics[align=c,width=1.6cm, height=2.0cm]{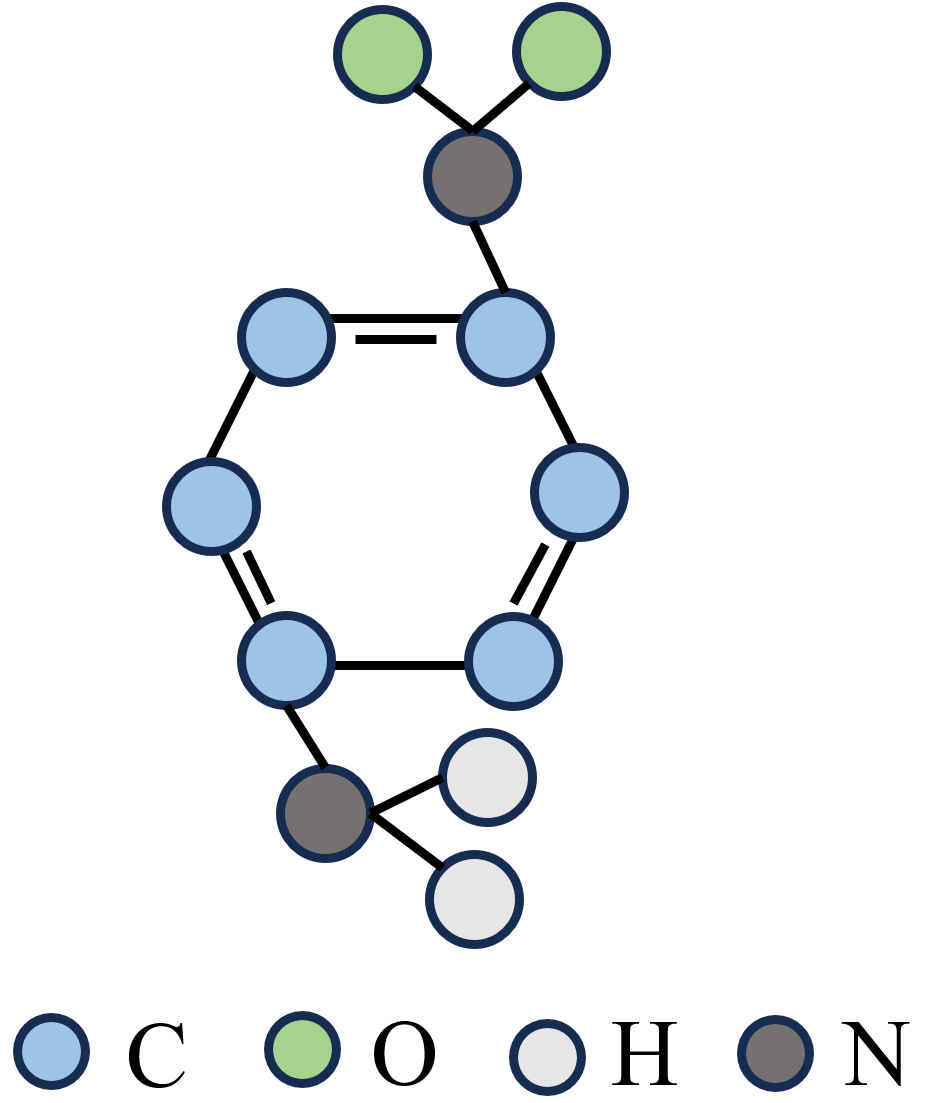} & \includegraphics[align=c,width=1.1cm, height=1.6cm]{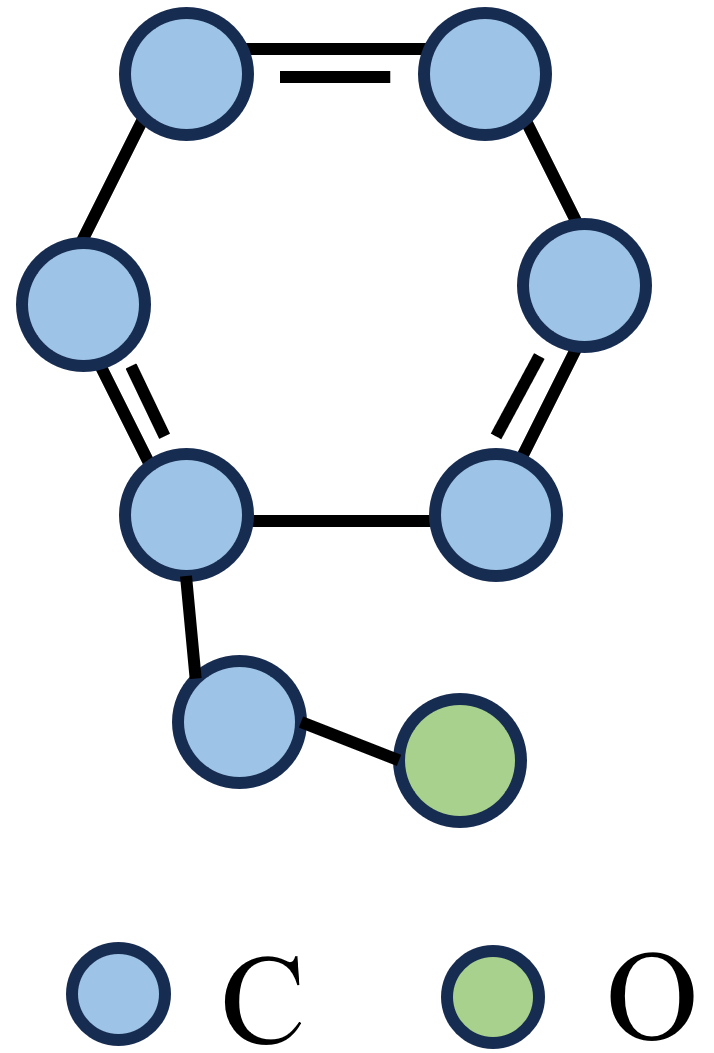} & \includegraphics[align=c,width=1.4cm, height=1.7cm]{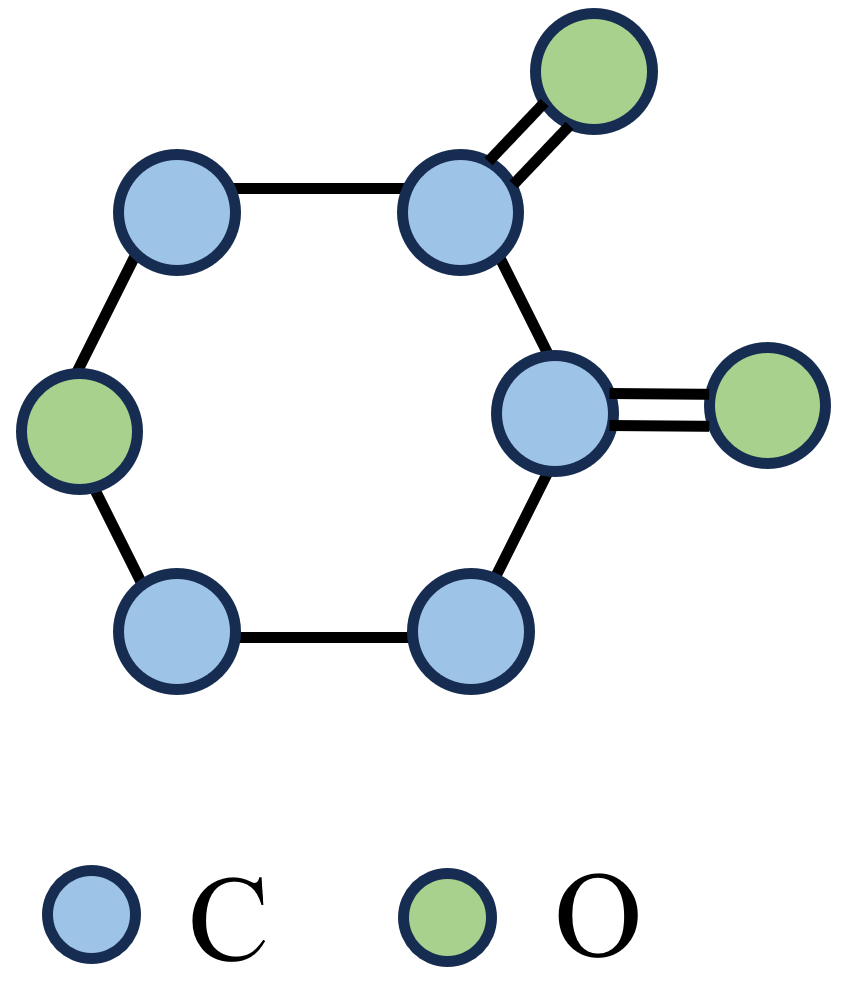}  & \includegraphics[align=c,width=1.8cm, height=2cm]{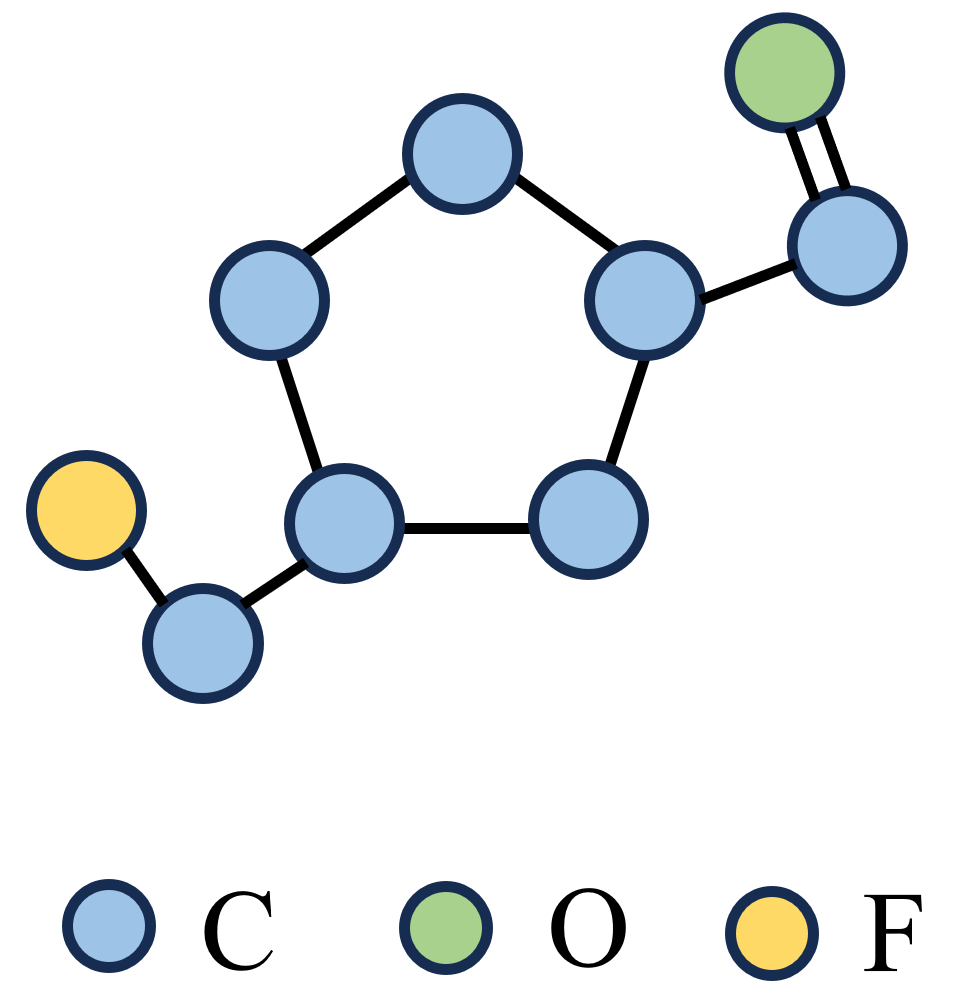} & \multicolumn{2}{c}{\includegraphics[align=c,width=3.3cm, height=1.2cm]{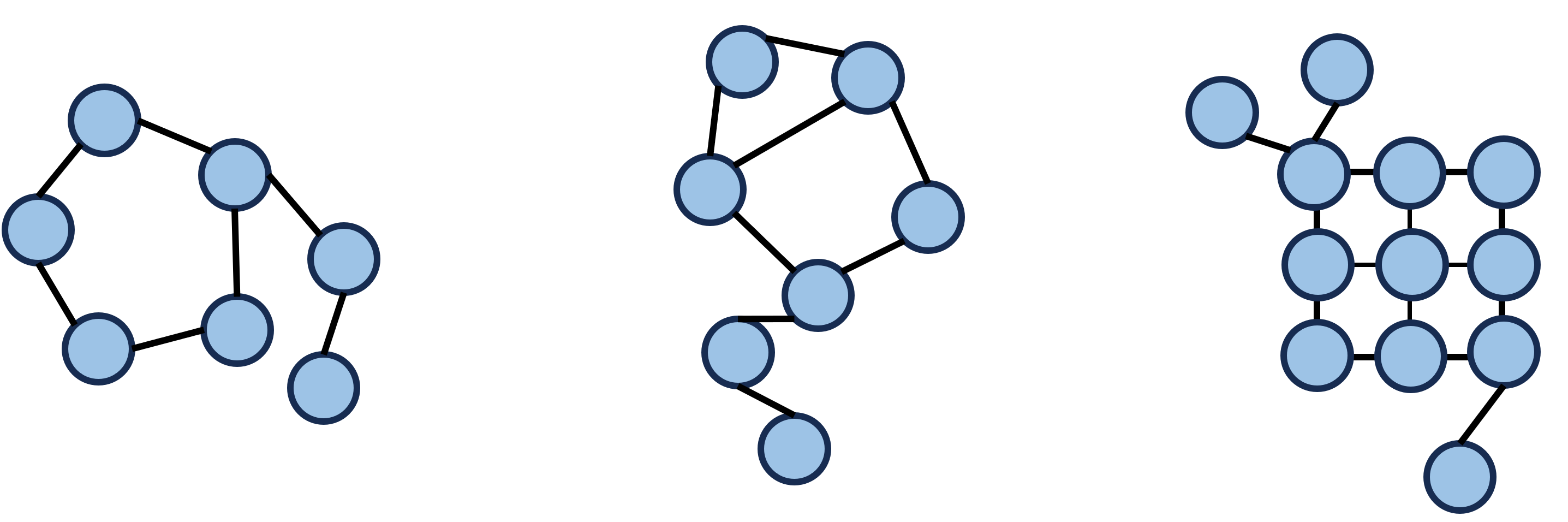}}  \\ 
    Ground truth explanation & \includegraphics[width=0.9cm, height=1.8cm]{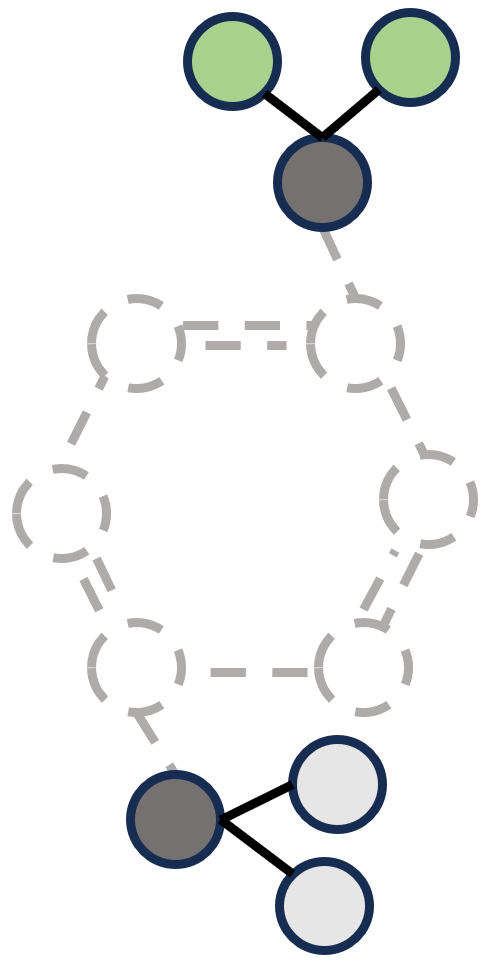} & \includegraphics[width=1.1cm, height=1.4cm]{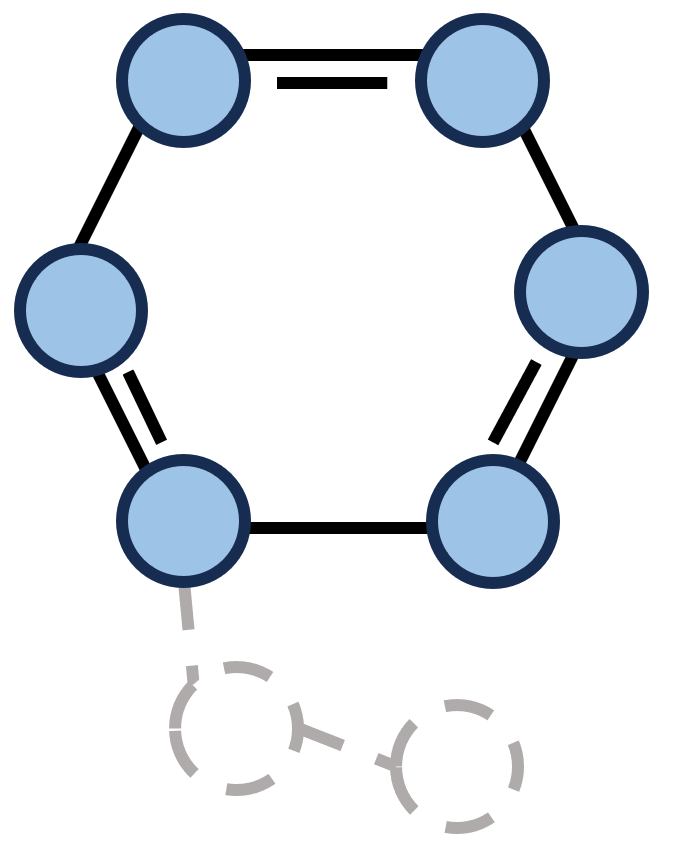} & \includegraphics[width=1.4cm, height=1.4cm]{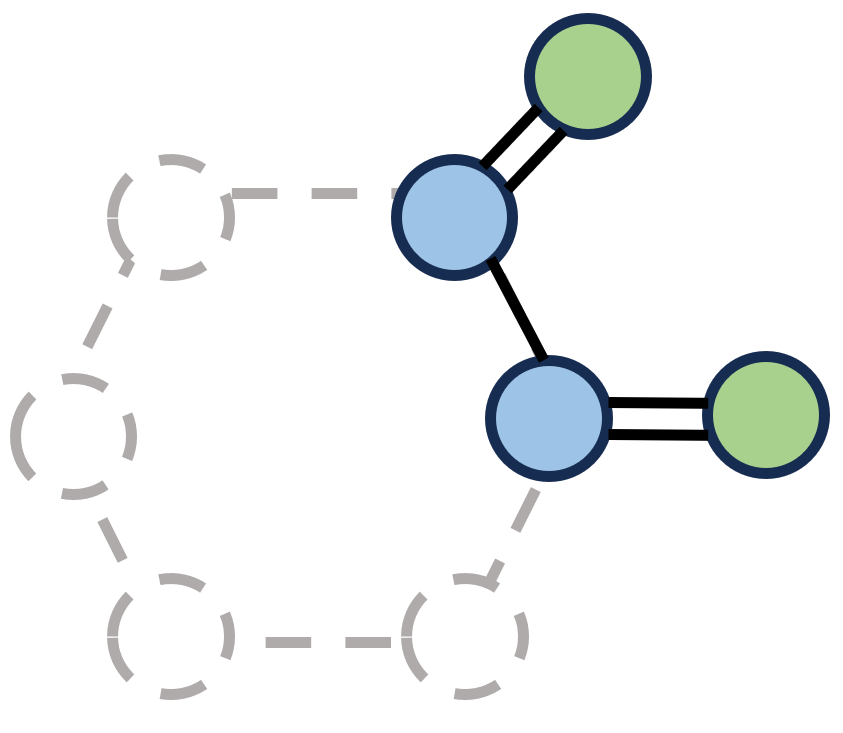}  & \includegraphics[width=1.8cm, height=1.5cm]{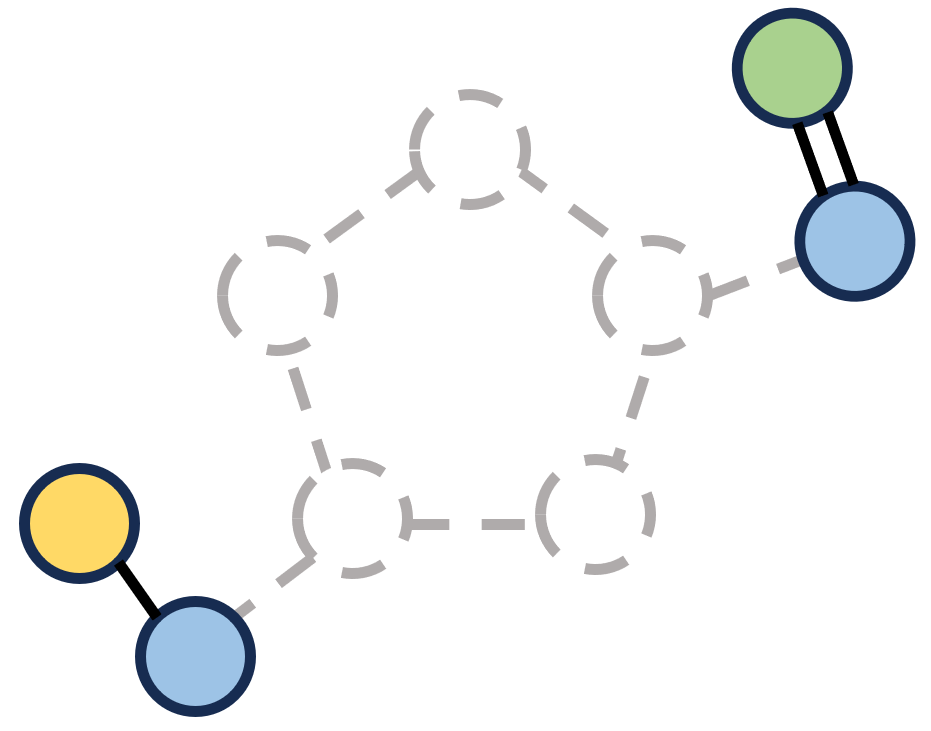} & \multicolumn{2}{c}{\includegraphics[width=3.3cm, height=1.2cm]{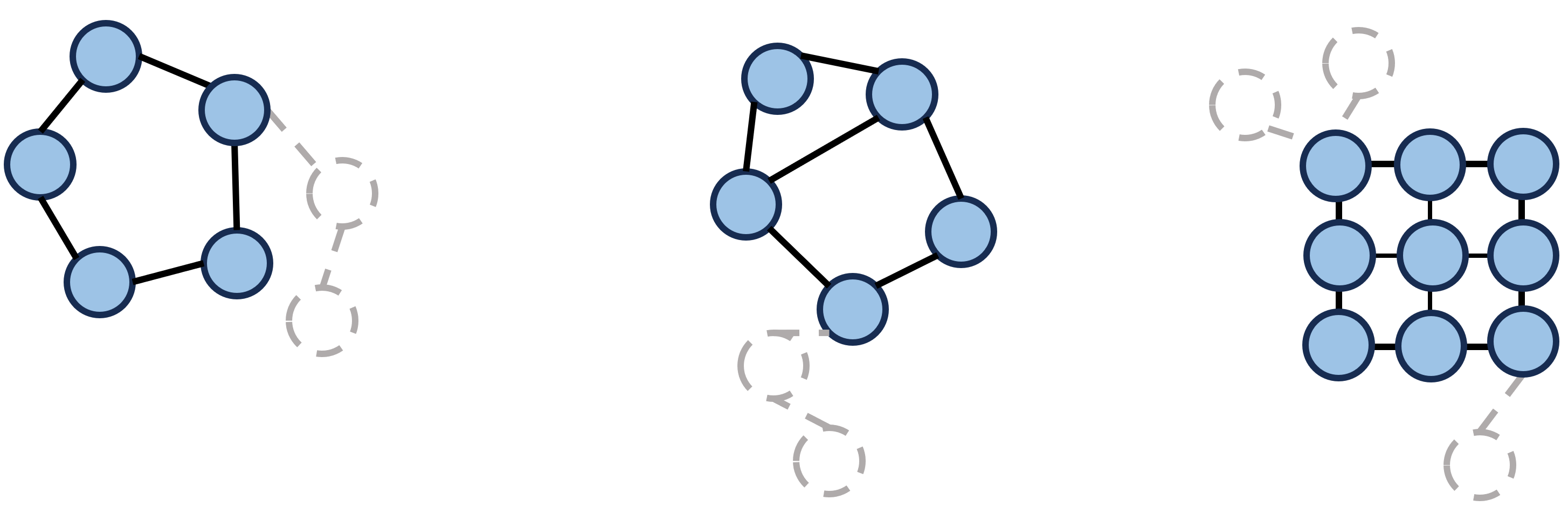}}   \\ 
    \textbf{} & NH$_{2}$, NO$_2$ & {\benz} Ring & Alkane,C=O  & F$^{-}$,C=O & House, cycle & House,cycle, grid  \\ 
    \bottomrule
    \end{tabular}
    \label{tab:dataset}
\end{table*}

\begin{table*}[h]
    \centering
    \caption{The graph-classification task performances of the GCN model}
    \begin{tabular}{p{2cm} C{2cm} C{2cm} C{2cm}C{2cm}C{2cm}C{2cm}} 
    \toprule
    & {\mutag} & {\benz} & {\alk}  & {\fluo}  & {\bamo}  & {\bathree} \\ \midrule
    Train Acc & 0.850 & 0.930 & 0.979  & 0.951 & 0.999   & 0.997   \\  
    Val Acc & 0.834 & 0.927 & 0.986  & 0.956 &  1.0  & 0.997 \\ 
    Test Acc & 0.804 & 0.915 & 0.975   & 0.951 &  1.0  & 0.977 \\ 
    \bottomrule
    \end{tabular}
    \label{tab:pretrainresults}
\end{table*}

\subsection{Baselines}
To evaluate our model, we well-train the GCN model in to ensure that it takes good performance in graph classification tasks. The results are displayed in Table~\ref{tab:pretrainresults}.  For a comprehensive comparison, we incorporate various post-hoc explanation methods, including GradCAM, GNNExplainer, PGExplainer, ReFine, and MixupExplainer.
\begin{itemize} [leftmargin=*]
\item \textbf{GradCAM \cite{pope2019explainability}.}
 This method utilizes gradients as a weighting mechanism to merge various feature maps. It operates on heuristic assumptions and cannot elucidate node classification models.
 \item \textbf{GNNExplainer \cite{ying2019gnnexplainer}.}
It learns soft masks for edges and node features, and aims to elucidate predictions through mask optimization. GNNExplainer integrates these masks with the original graph through element-wise multiplications. The masks are optimized by enhancing the mutual information between the original graph and the modified graph prediction results.
\item \textbf{PGExplainer \cite{luo2020parameterized}.} 
This method extends the idea of GNNExplainer by assuming that the graph is a random Gilbert graph. PGExplainer generates each edge embedding by combining the embeddings of its constituent nodes, then uses these edge embeddings to determine a Bernoulli distribution to indicate whether to mask an edge or not, and utilizes a Gumbel-Softmax approach to model the Bernoulli distribution for end-to-end training.
\item \textbf{ReFine \cite{wang2021towards}.}
ReFine identifies the edge probabilities for the entire category by maximizing the mutual information and contrastive loss between categories. In fine-tuning, it uses the edge probabilities from the previous stage to sample edges, and find explanations that maximize mutual information for specific instances.
\item \textbf{MixupExplainer \cite{zhang2023mixupexplainer}.} This method combines original explanatory subgraphs with randomly sampled, label-independent base graphs in a non-parametric way to mitigate the common OOD issue which found in previous methods.

\end{itemize}

\section{Extensive Experiments}

\subsection{Extensive Distribution Analysis} \label{sec:extensivedistr}
As shown in previous work~\cite{zhang2023mixupexplainer}, another way to approximate the distribution differences of graphs is to compare their vector embeddings. Here, we adopt the same setting and use Cosine similarity and Euclidean distance to intuitively approximate the similarities between graph distributions. Table \ref{tab:distribution} presents the computed Cosine similarity and Euclidean distance between the distribution embeddings of the original graph $\textit{\textbf{h}}$, the ground truth explanation subgraph $\textit{\textbf{h}}'$, and the generated proxy graph $\tilde{\textit{\textbf{h}}}$. Notably, our proxy graph embedding $\tilde{\textit{\textbf{h}}}$ exhibits higher Cosine similarity scores and lower Euclidean distance with the original graph embedding $\textit{\textbf{h}}$ compared to the ground truth explanation embedding $\textit{\textbf{h}}'$. We observe an average improvement of $19.5\%$ in Cosine similarity and $35.6\%$ in Euclidean distance. Particularly in the {\bamo} dataset, there is a significant improvement of $60.4\%$ in Cosine similarity and $51.8\%$ in Euclidean distance. These findings underscore the effectiveness of our {\ours} method in mitigating distribution shifts caused by induction bias in the to-be-explained GNN model $f(\cdot)$, thereby enhancing explanation performance.

\begin{table*}[h!]
    \centering
    \caption{The Cosine similarity score and Euclidean distance between the distribution embeddings of the original graph $\textit{\textbf{h}}$, explanation subgraph $\textit{\textbf{h}}'$, and our proxy graph $\tilde{\textit{\textbf{h}}}$ on different datasets.}
    \begin{tabular}{p{4cm}C{1.6cm}C{1.6cm} C{1.6cm}C{1.6cm}C{1.6cm}C{1.6cm}}  
    \toprule
    & {\mutag} & {\benz} & {\alk}  & {\fluo}  & {\bamo} & {\bathree} \\ \midrule
    Avg. Cosine($\textit{\textbf{h}}$, $\textit{\textbf{h}}'$) $\uparrow$  & 0.883 & 0.835 & 0.889  & 0.904 & 0.571 & 0.686 \\ 
    Avg. Cosine($\textit{\textbf{h}}$, $\tilde{\textit{\textbf{h}}}$) $\uparrow$  & \textbf{0.985} & \textbf{0.905} & \textbf{0.938}  & \textbf{0.908} & \textbf{0.916} & \textbf{0.918}\\  
    \hline 
    Avg. Euclidean($\textit{\textbf{h}}$, $\textit{\textbf{h}}'$) $\downarrow$  & 0.975 & 1.010 & 0.940  & 0.806  & 1.210 & 1.199  \\ 
    Avg. Euclidean($\textit{\textbf{h}}$, $\tilde{\textit{\textbf{h}}}$) $\downarrow$ & \textbf{0.368} & \textbf{0.767} & \textbf{0.719}   & \textbf{0.779} & \textbf{0.583} & \textbf{0.613}  \\ 
    \bottomrule
    \end{tabular}
    \label{tab:distribution}
\end{table*}

\subsection{Extensive Ablation Study} \label{sec:extensiveablation}
\begin{figure*}[h]
    \centering
    \begin{subfigure}[b]{0.32\textwidth}
        \centering
        \includegraphics[width=\textwidth]{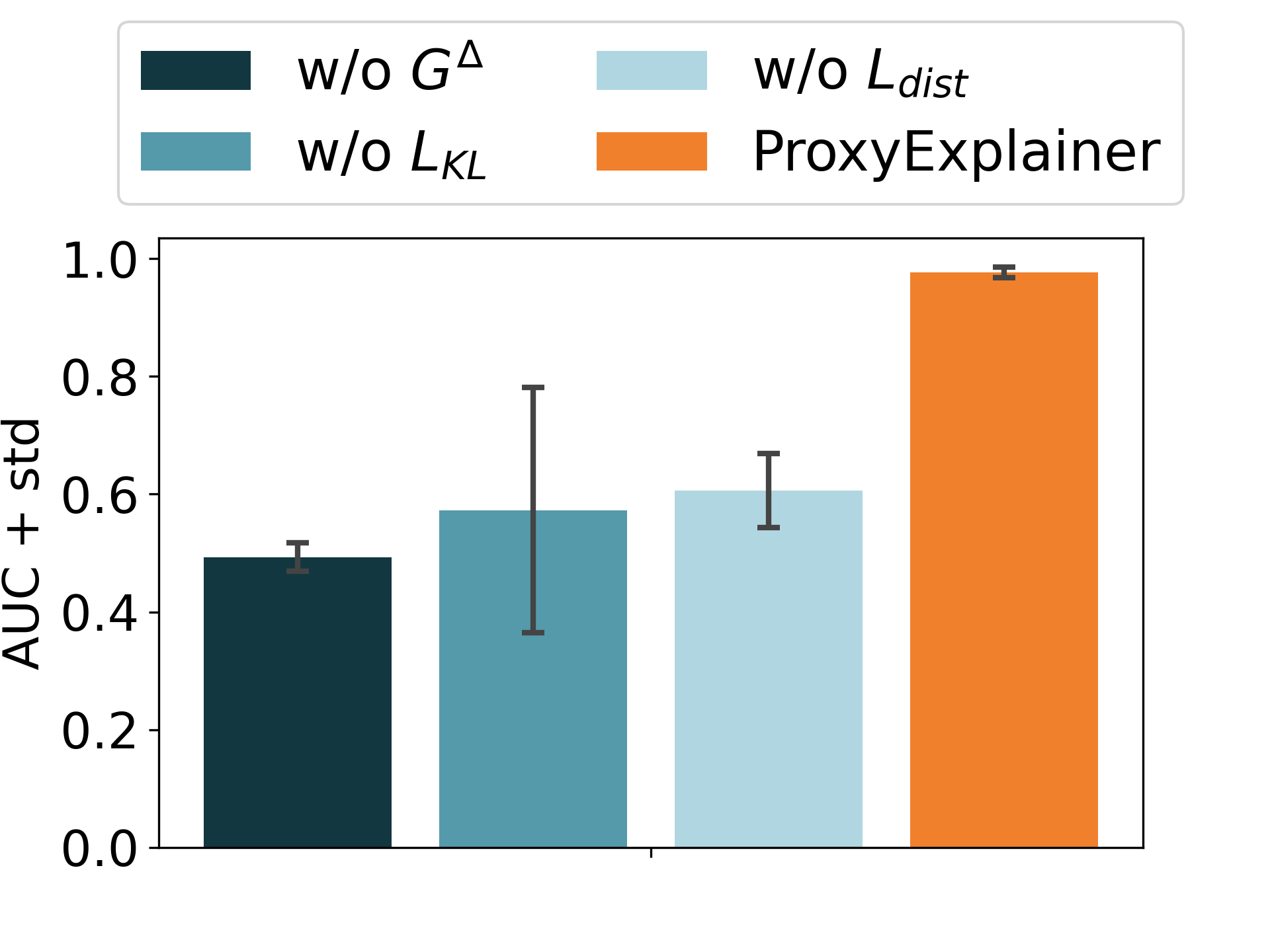}
        \caption{{\mutag}}
        \label{fig:mutag}
    \end{subfigure}
    \begin{subfigure}[b]{0.32\textwidth}
        \centering
        \includegraphics[width=\textwidth]{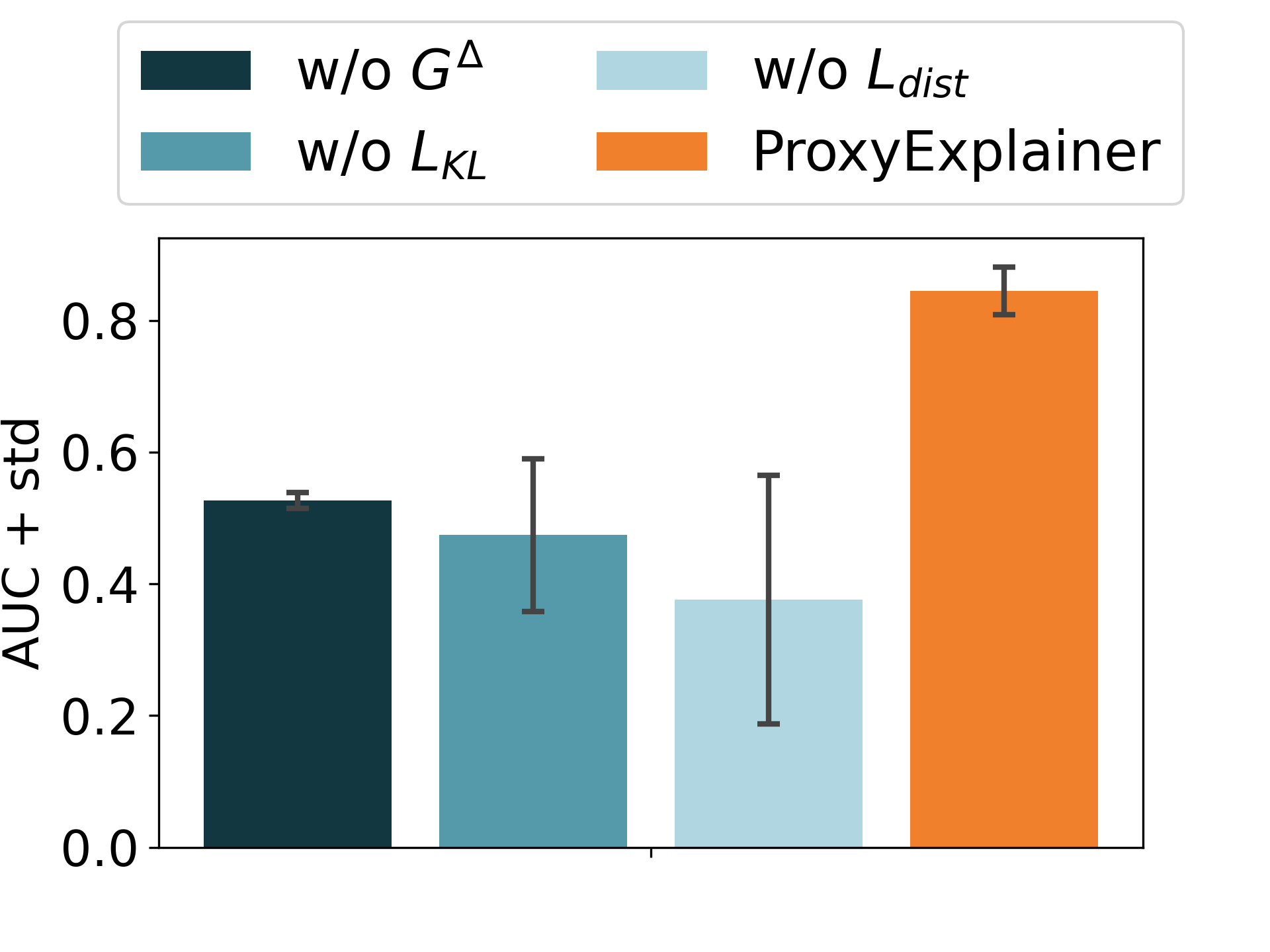}
        \caption{{\benz}}
        \label{fig:ben}
    \end{subfigure}
    \begin{subfigure}[b]{0.32\textwidth}
        \centering
        \includegraphics[width=\textwidth]{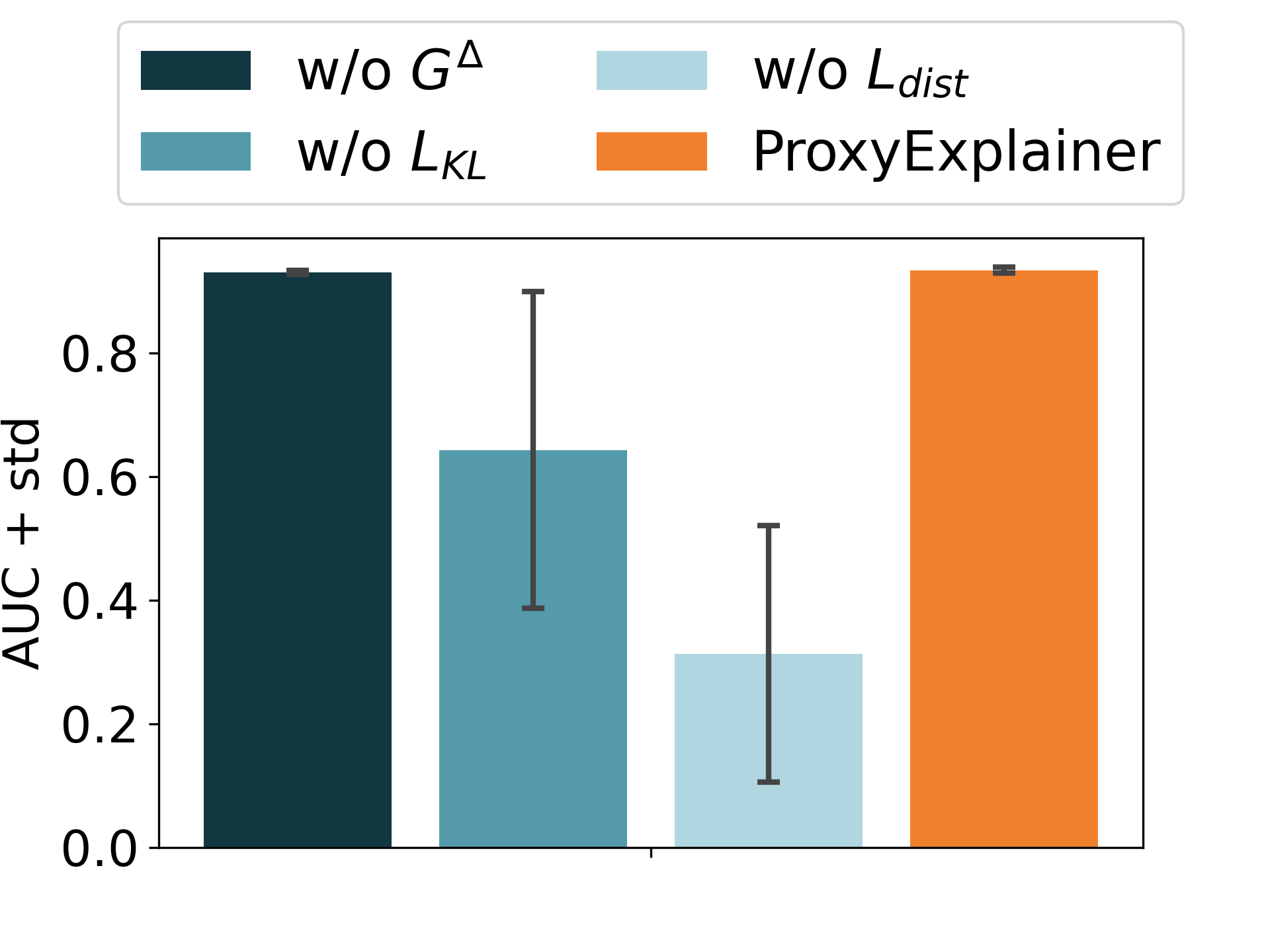}
        \caption{{\alk}}
        \label{fig:ac}
    \end{subfigure}
    \begin{subfigure}[b]{0.32\textwidth}
        \centering
        \includegraphics[width=\textwidth]{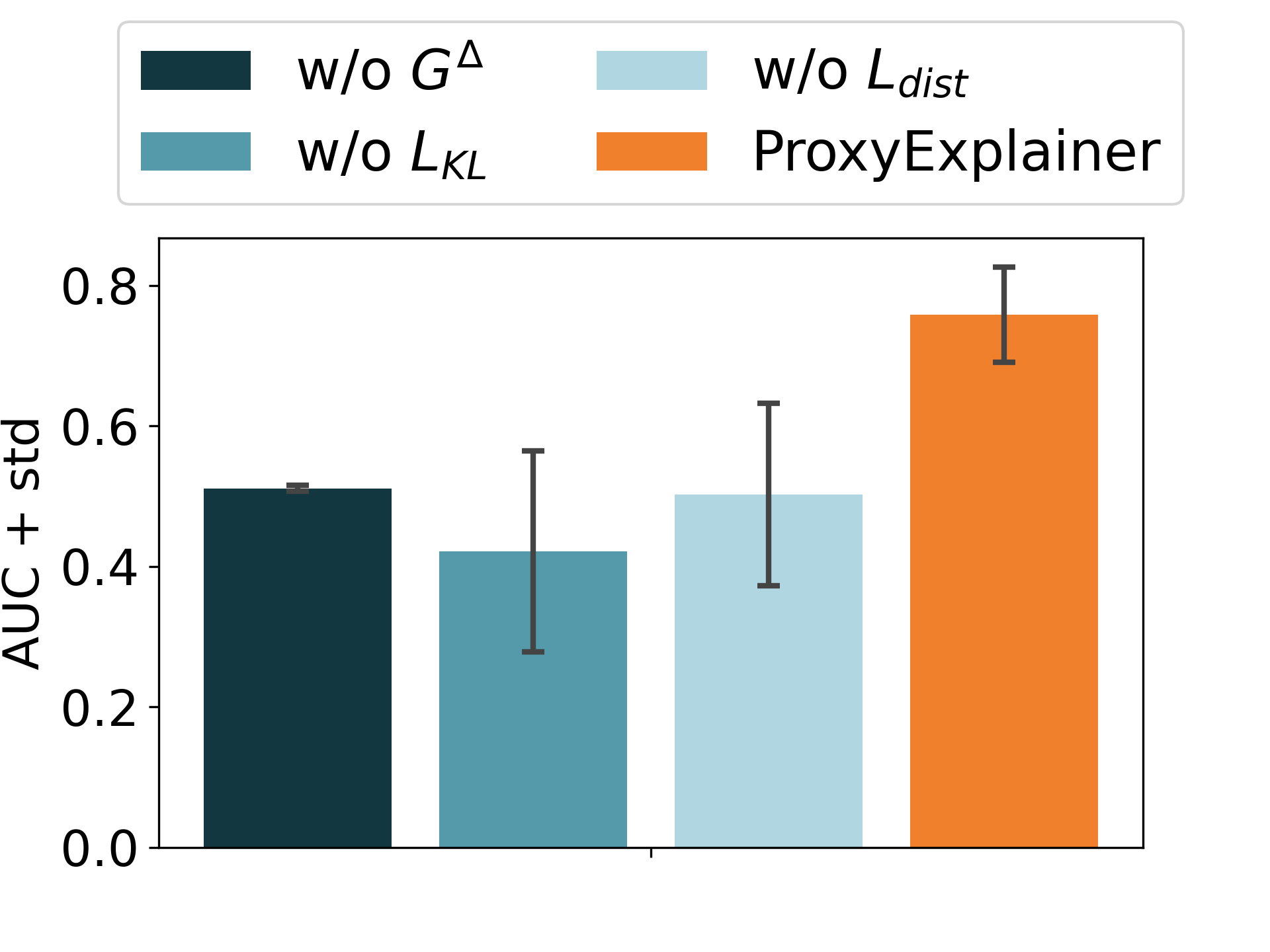}
        \caption{{\fluo}}
        \label{fig:fc}
    \end{subfigure}
    \begin{subfigure}[b]{0.32\textwidth}
        \centering
        \includegraphics[width=\textwidth]{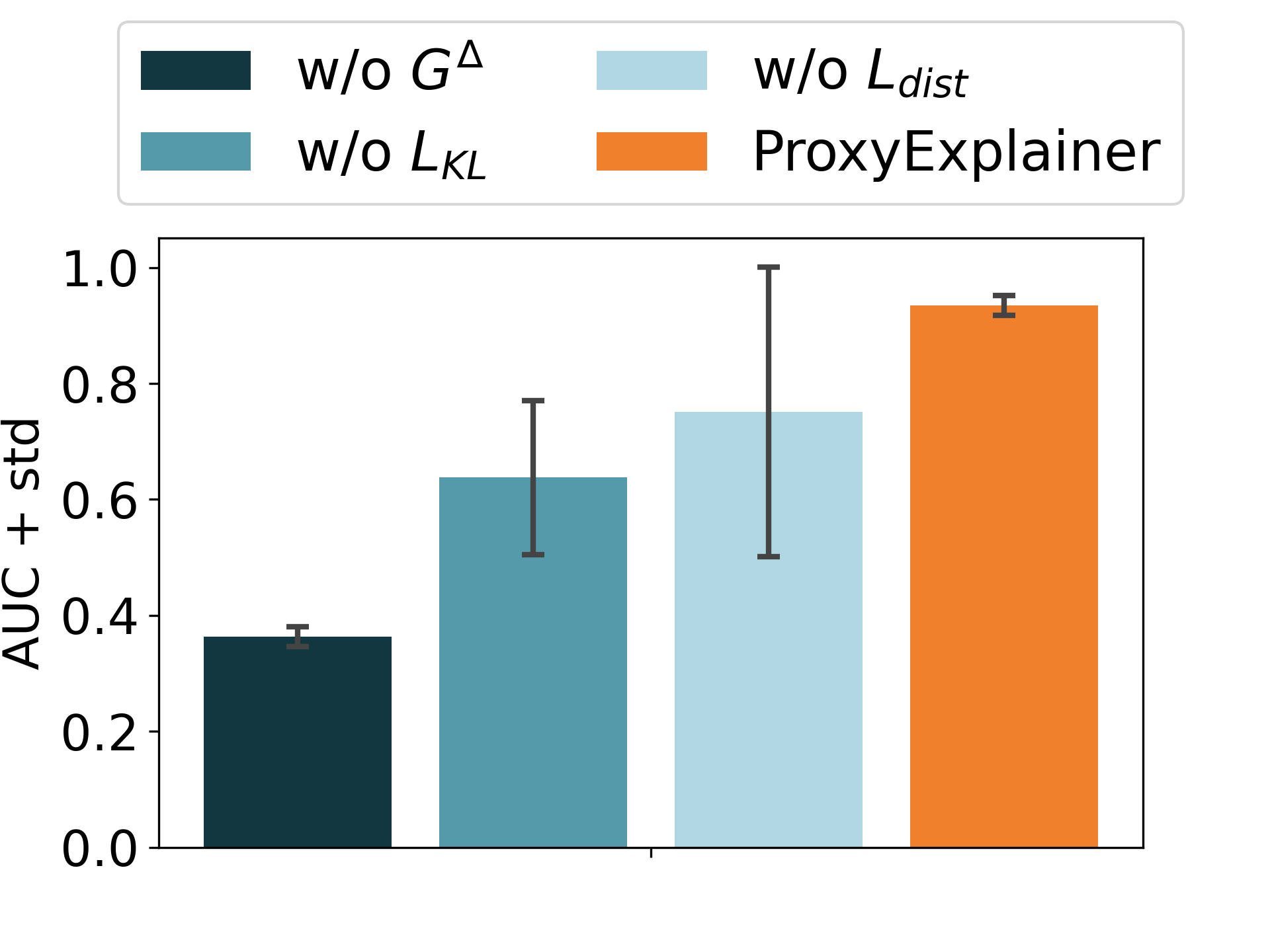}
        \caption{{\bamo}}
        \label{fig:ba2}
    \end{subfigure}
    \begin{subfigure}[b]{0.32\textwidth}
        \centering
        \includegraphics[width=\textwidth]{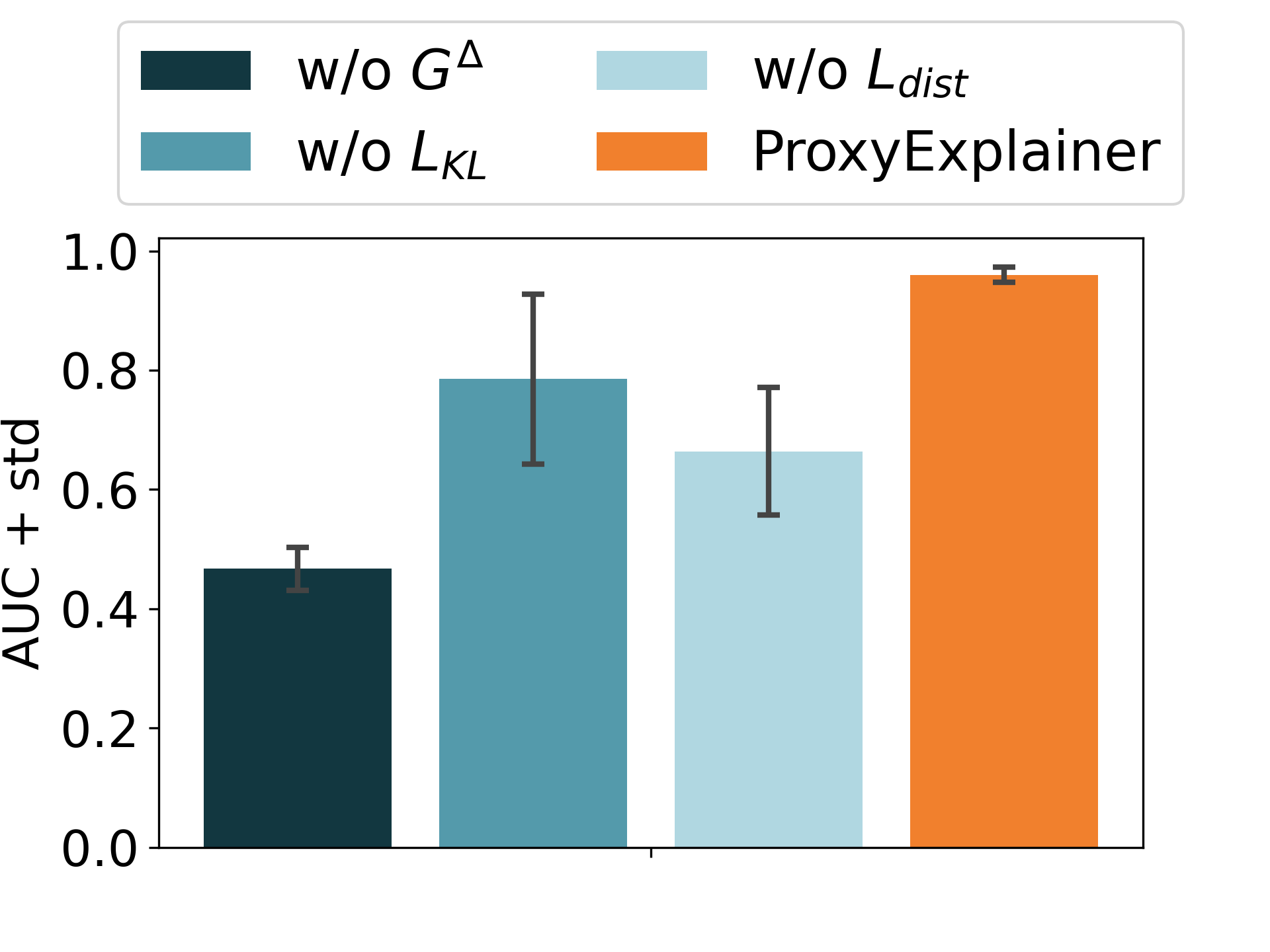}
        \caption{{\bathree}}
        \label{fig:ba3}
    \end{subfigure}
    \caption{Ablation study on all datasets.}
    \label{fig:Ablationstudyonalldatasets}
\end{figure*}

In Figure \ref{fig:Ablationstudyonalldatasets}, we conduct a comprehensive ablation study across all datasets to examine the impact of different components. The findings illustrate the effectiveness of these components and their positive contribution to our {\ours}.

\subsection{Extensive Experiment with GIN} \label{sec:ginexplain}

\begin{figure*}[h]
    \centering
    \includegraphics[width=0.6\textwidth]{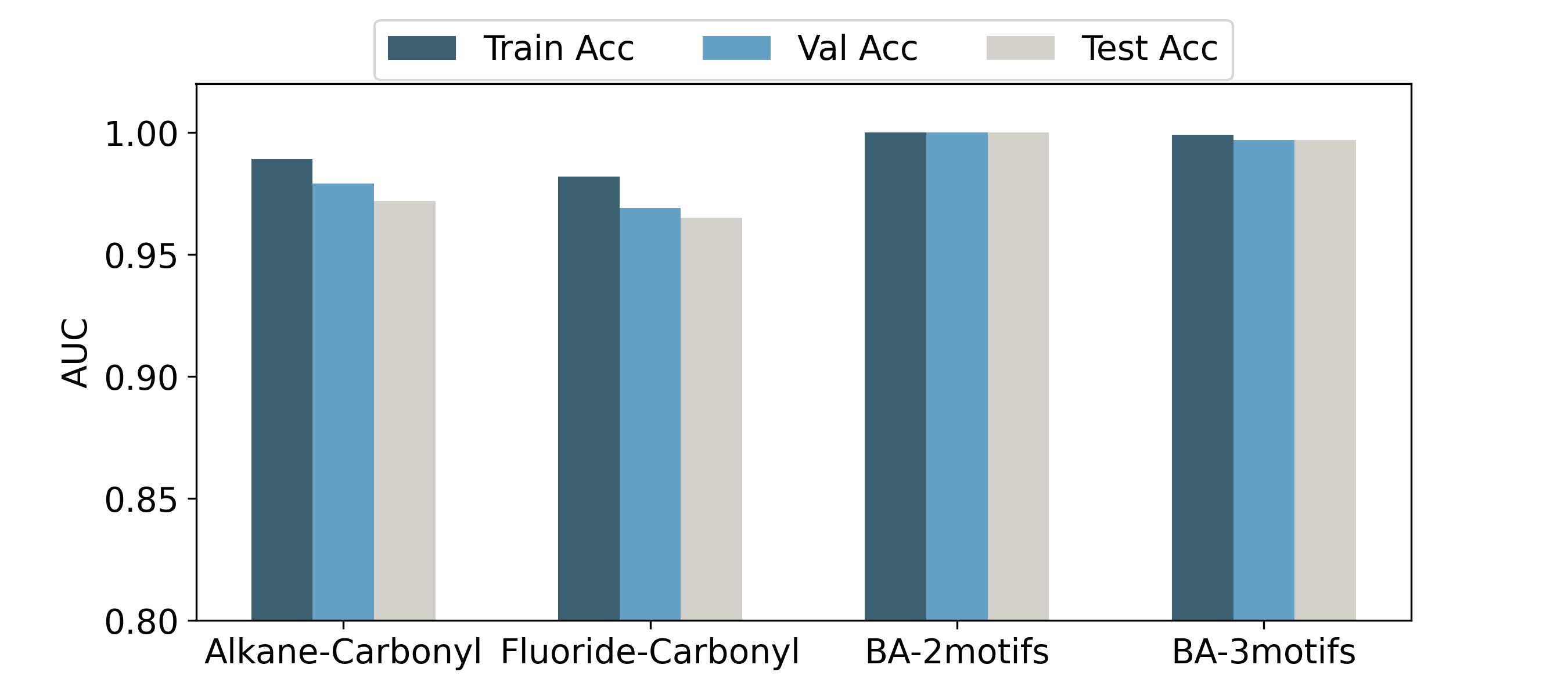} 
    \caption{The graph-classification task performances of the GIN model.}
    \label{fig:auc_gin}
\end{figure*}

\begin{figure*}[h]
    \centering
    \begin{subfigure}[b]{0.38\textwidth}
        \centering
        \includegraphics[width=1.0\textwidth]{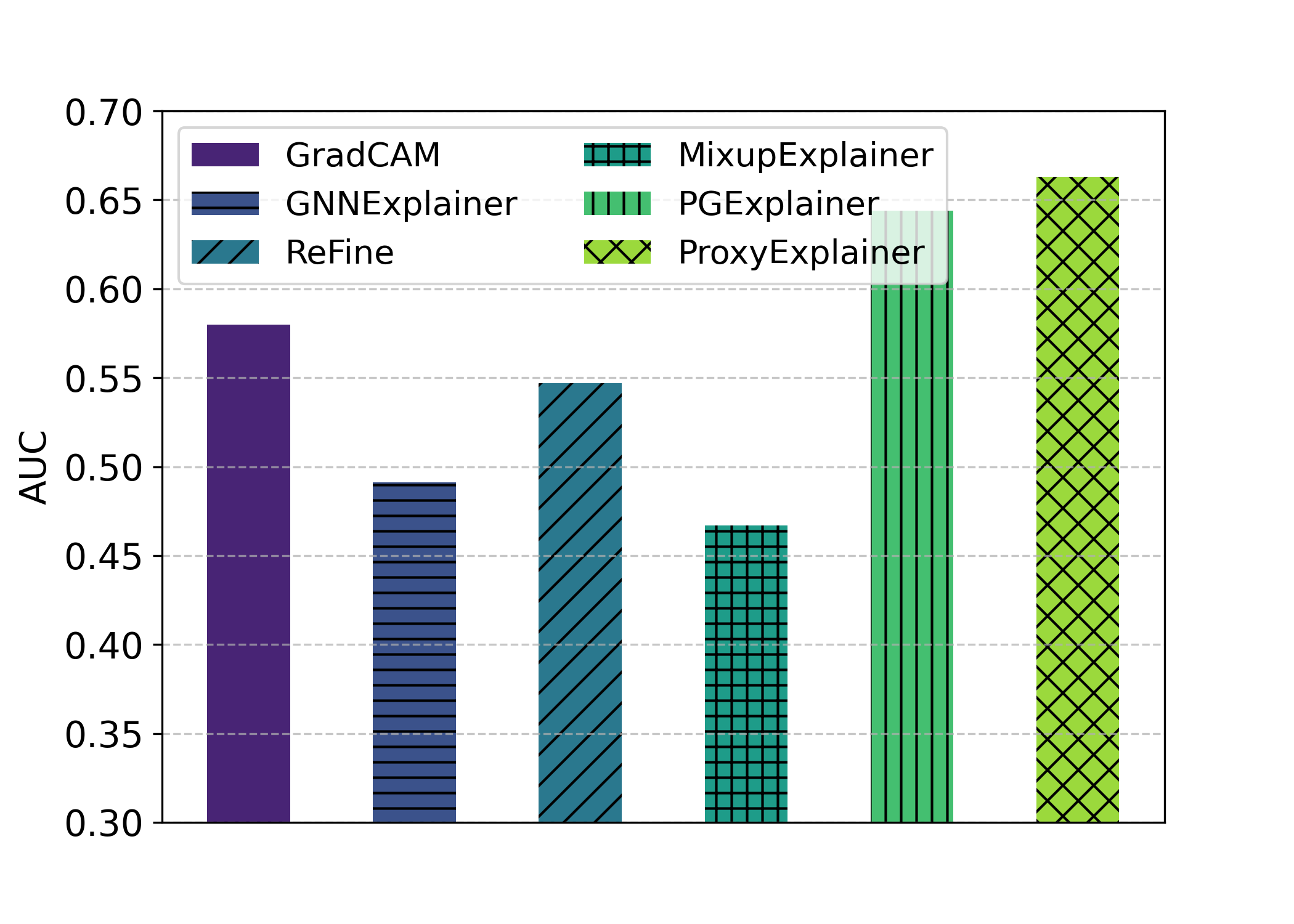}
        \captionsetup{skip=1pt}
        \caption{{\alk}}
        \label{fig:gin_ac}
    \end{subfigure}
    \begin{subfigure}[b]{0.38\textwidth}
        \centering
        \includegraphics[width=1.0\textwidth]{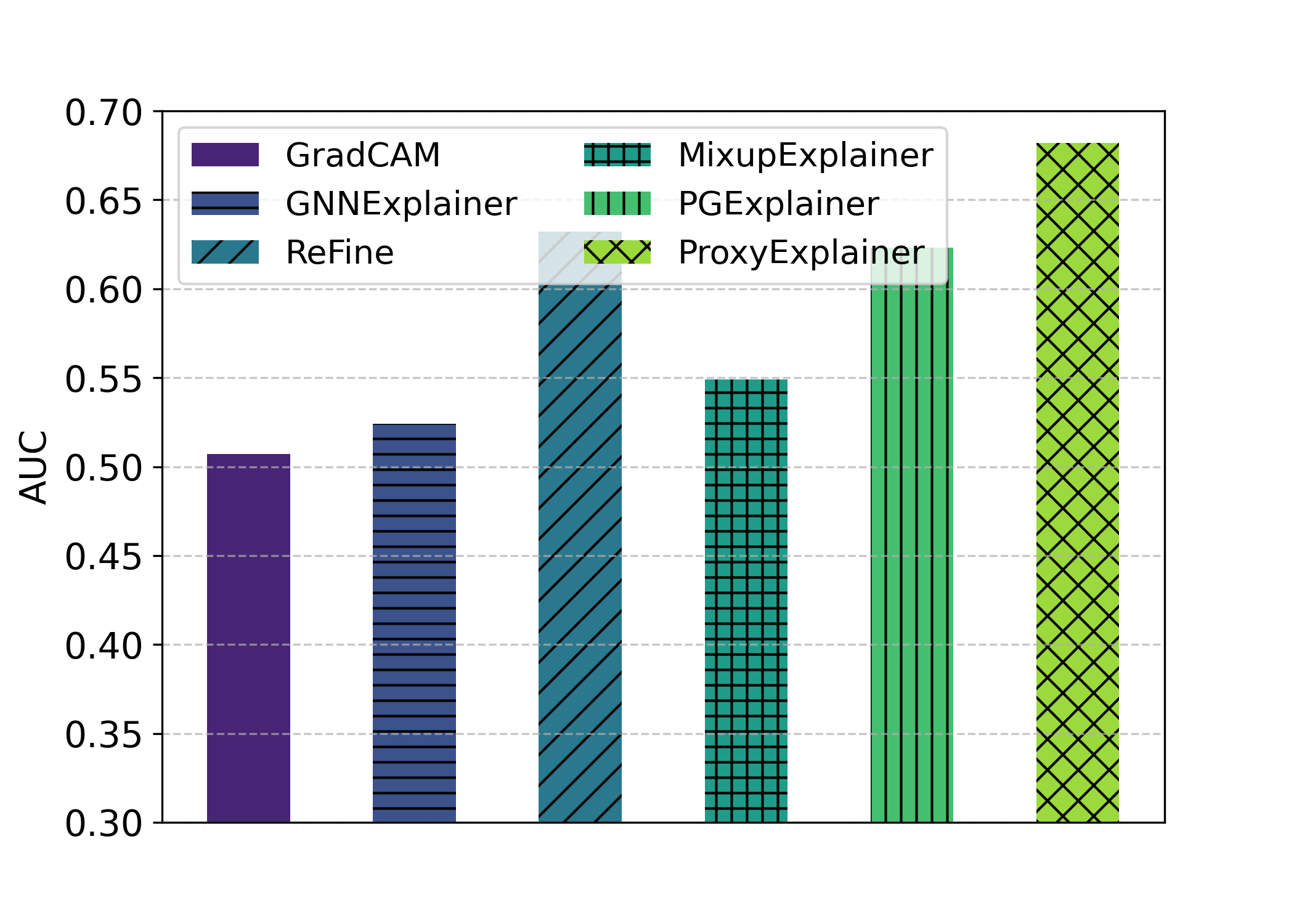}
        \captionsetup{skip=1pt}
        \caption{{\fluo}}
        \label{fig:gin_fc}
    \end{subfigure}
    \begin{subfigure}[b]{0.38\textwidth}
        \centering
        \includegraphics[width=1\textwidth]{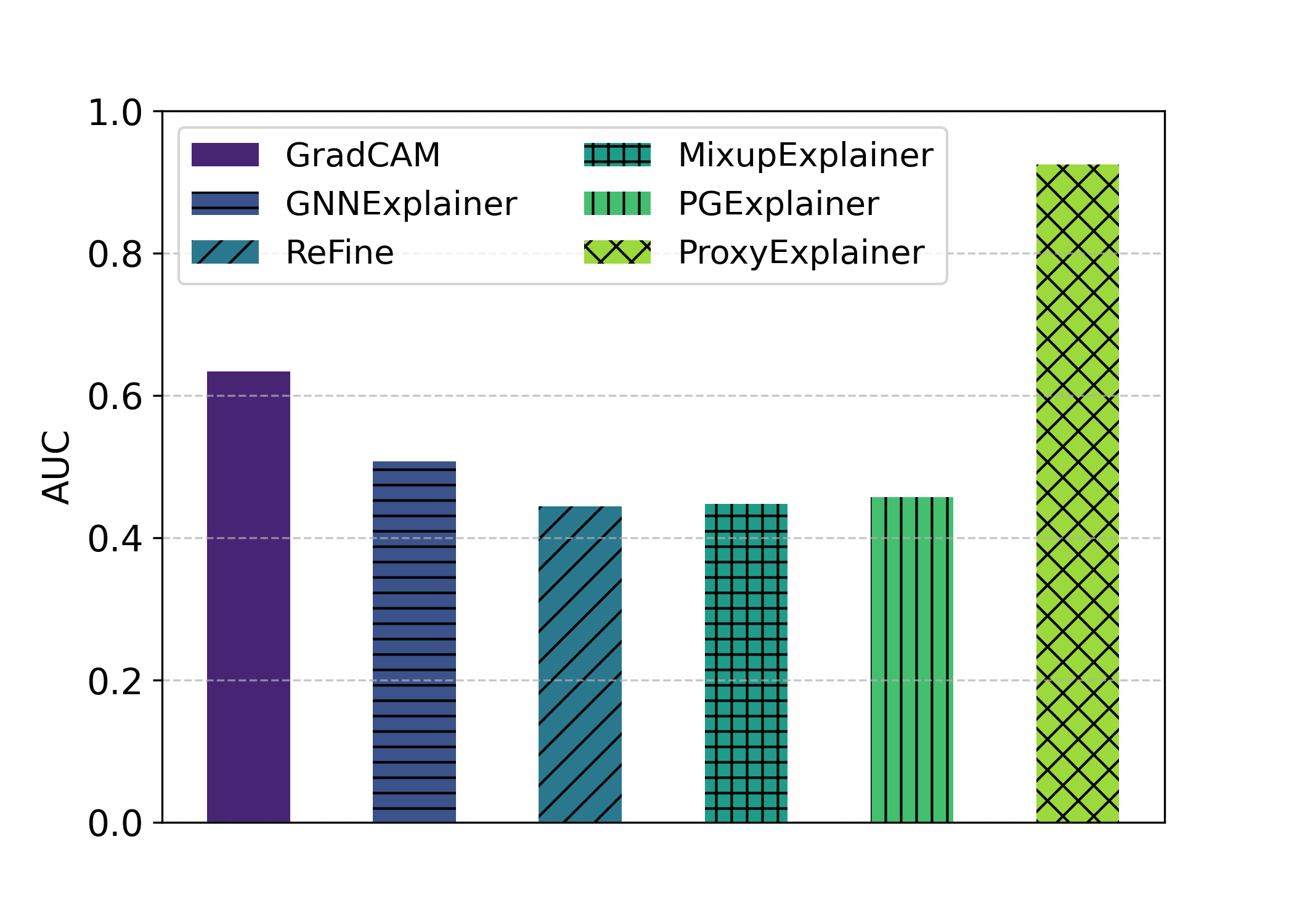}
        \captionsetup{skip=1pt}
        \caption{{\bamo}}
        \label{fig:gin_ba2}
    \end{subfigure}
    \begin{subfigure}[b]{0.38\textwidth}
        \centering
        \includegraphics[width=1.0\textwidth]{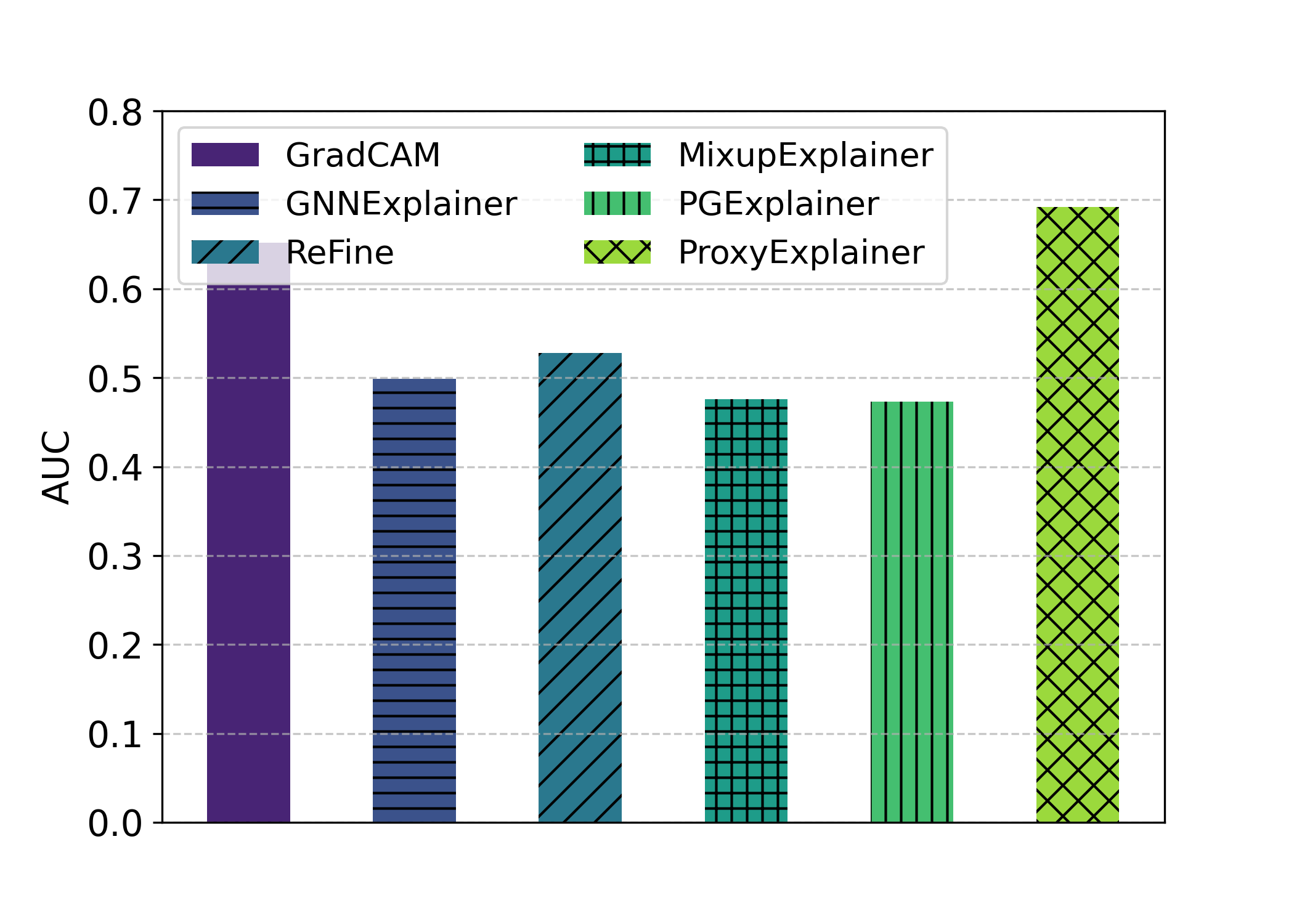}
        \captionsetup{skip=1pt}
        \caption{{\bathree}}
        \label{fig:gin_ba3}
    \end{subfigure}
    \caption{Explanation accuracy in terms of AUC-ROC on edges based on GIN.}
    \label{fig:genera_gin}
\end{figure*}

In order to show the robustness of our {\ours} in explaining different GNN models, we first pre-train GIN on two real-world datasets and two synthetic datasets to ensure that GIN has the ability to classify graphs accurately. The results are displayed in Figure~\ref{fig:auc_gin}. Then we use both baseline methods in the previous experiment and our method {\ours} to provide explanations for the pre-trained GIN model. As seen in Figure~\ref{fig:genera_gin}, it is noticeable that {\ours} achieves the best performance on these datasets. Specifically, it improves the AUC scores of $2.9\%$ on {\alk}, $7.9\%$ on {\fluo}, $45.9\%$ on {\bamo}, and $6.1\%$ on {\bathree}. From the analysis, we can see that our {\ours} can identify accurate explanations across different datasets.

\subsection{Parameter Sensitive Analysis}
In this section, we analyze the influence of parameters including $\lambda$, which controls the KL divergence during the proxy graph generation process, and the dimension $D$  of node latent embedding. We vary $\lambda$ from 0.01 to 1.0. For $D$, we vary it among \{32, 64, 128, 256, 512, 1024\}. We only change the value of a specific parameter while fixing the remaining parameters to their respective optimal values.

\begin{figure*}[h!]
    \centering
    \begin{subfigure}[b]{0.49\textwidth}
        \centering
        \includegraphics[width=\textwidth]{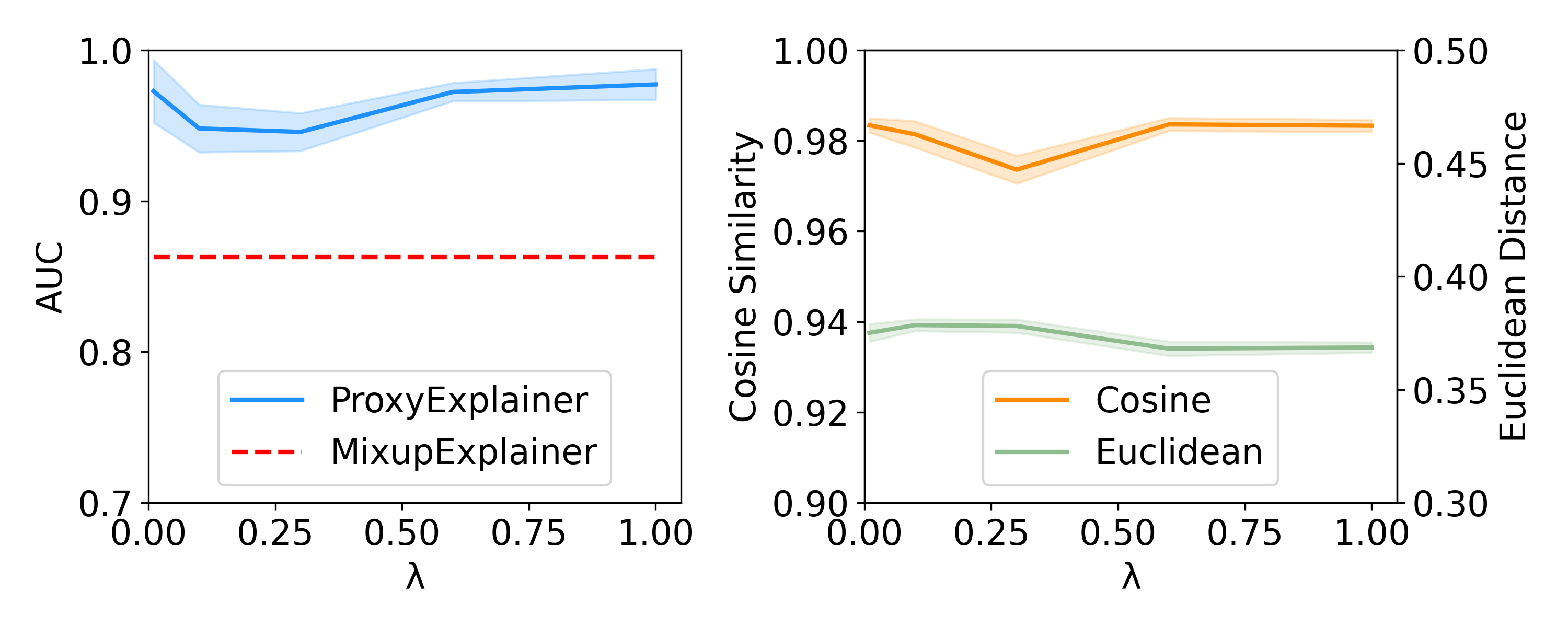}
        \caption{{\mutag}}
        \label{fig:lambda_all:sixth}
    \end{subfigure}
    \hfill 
    \begin{subfigure}[b]{0.49\textwidth}
        \centering
        \includegraphics[width=\textwidth]{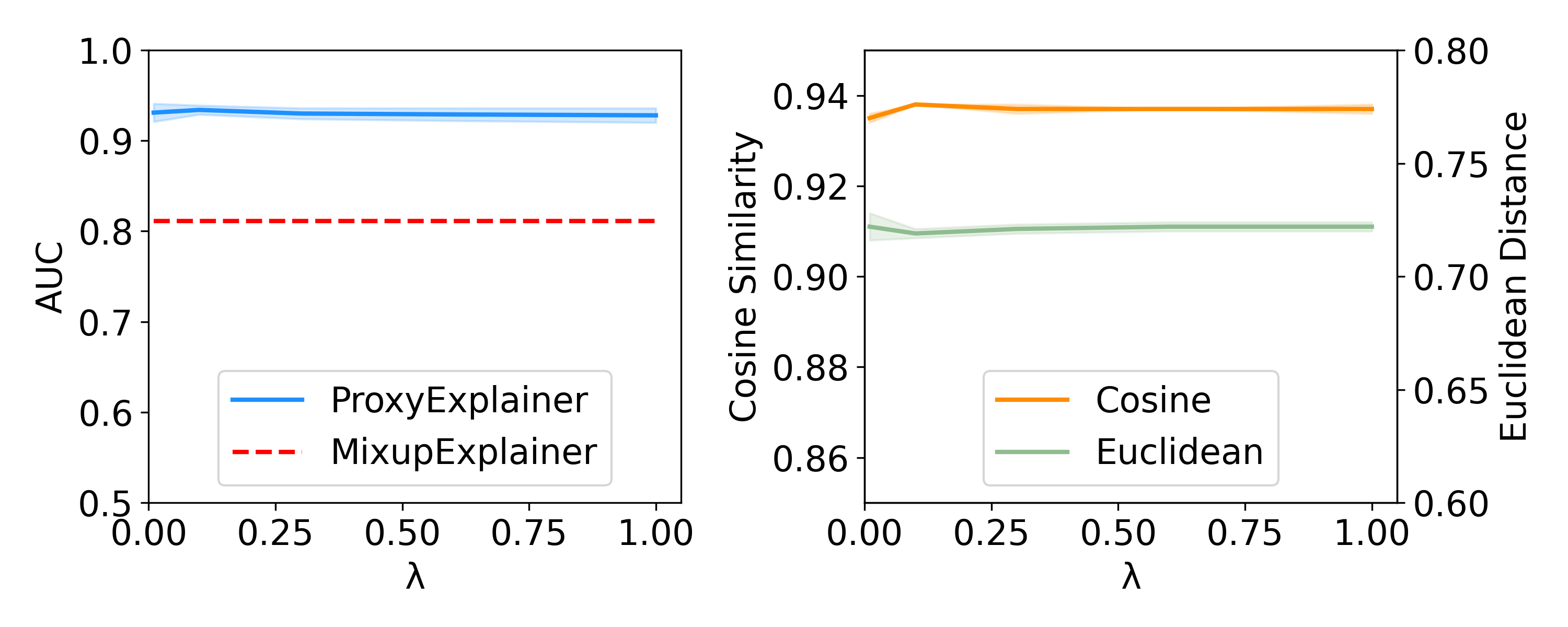}
        \caption{{\alk}}
        \label{fig:lambda_all:third}
    \end{subfigure}

    \begin{subfigure}[b]{0.49\textwidth}
        \centering
        \includegraphics[width=\textwidth]{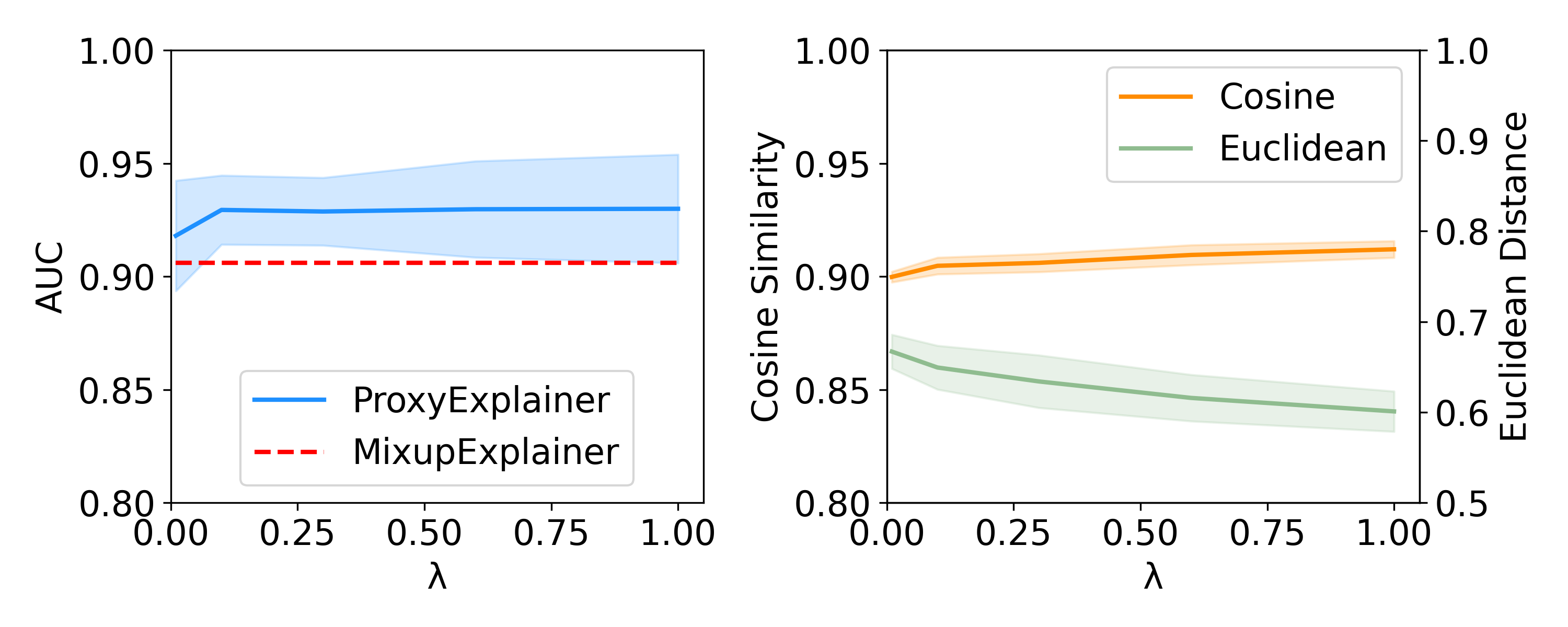}
        \caption{{\bamo}}
        \label{fig:lambda_all:first}
    \end{subfigure}
    \hfill 
    \begin{subfigure}[b]{0.49\textwidth}
        \centering
        \includegraphics[width=\textwidth]{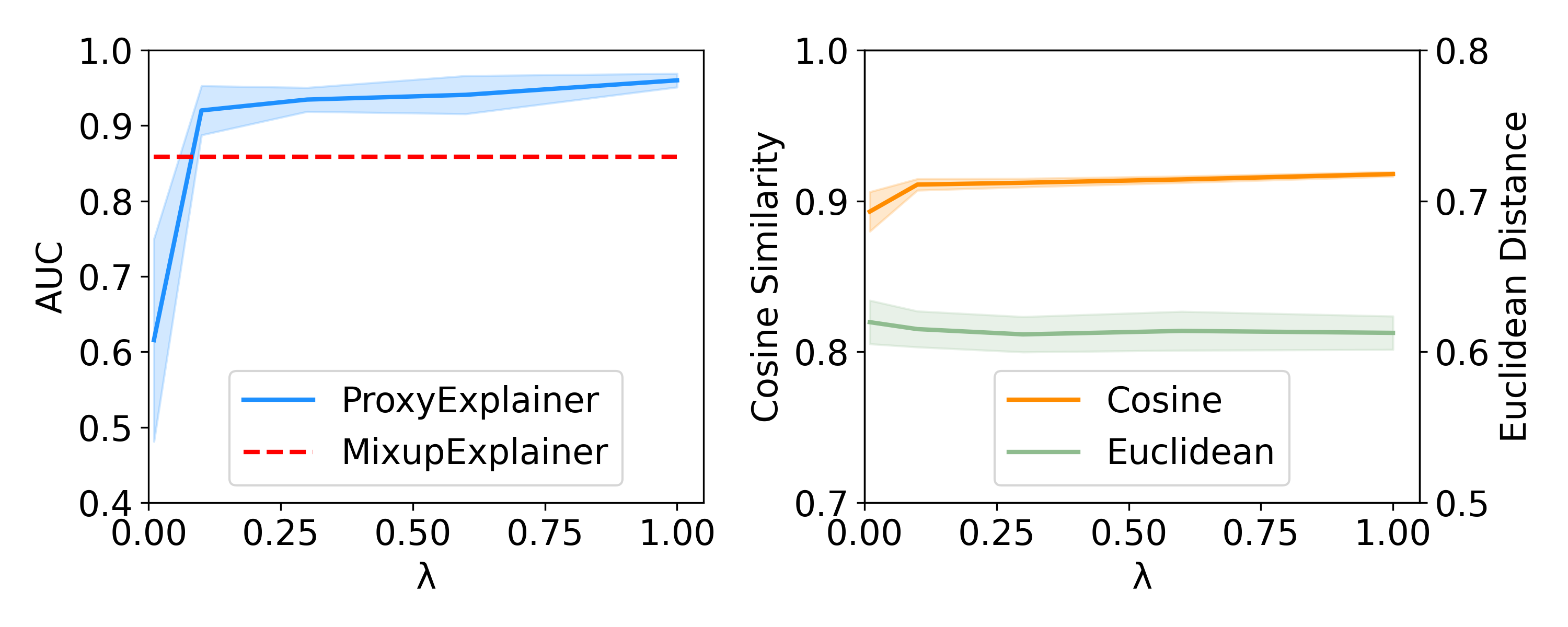}
        \caption{{\bathree}}
        \label{fig:lambda_all:second}
    \end{subfigure}
    \caption{Parameter analysis of $\lambda$ on four datasets. The left side of each graph shows the explanation performance. The right side shows the Distance Analysis between $\textit{\textbf{h}}$ and $\tilde{\textit{\textbf{h}}}$.}
    \label{fig:lambda_all}
\end{figure*}

\begin{figure*}[h!]
    \centering
    \begin{subfigure}[b]{0.49\textwidth}
        \centering
        \includegraphics[width=\textwidth]{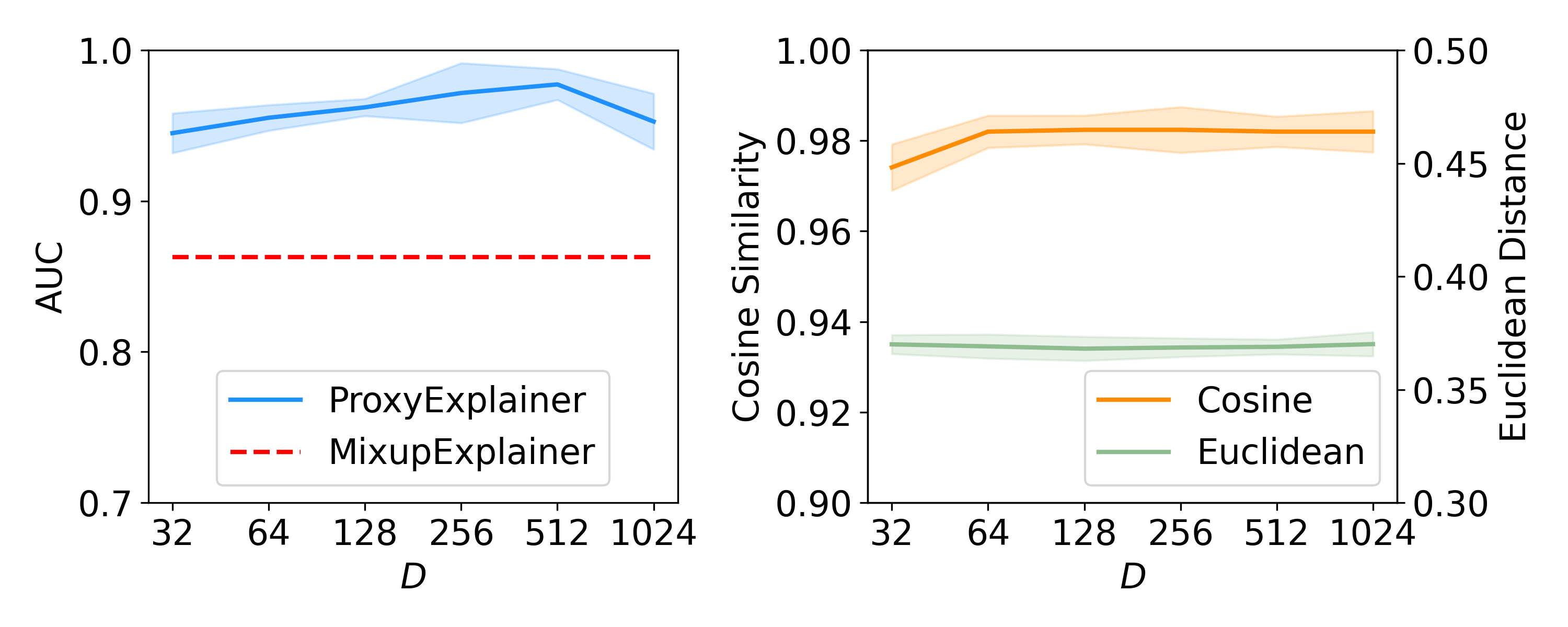}
        \caption{{\mutag}}
        \label{fig:sixth}
    \end{subfigure}
    \hfill 
    \begin{subfigure}[b]{0.49\textwidth}
        \centering
        \includegraphics[width=\textwidth]{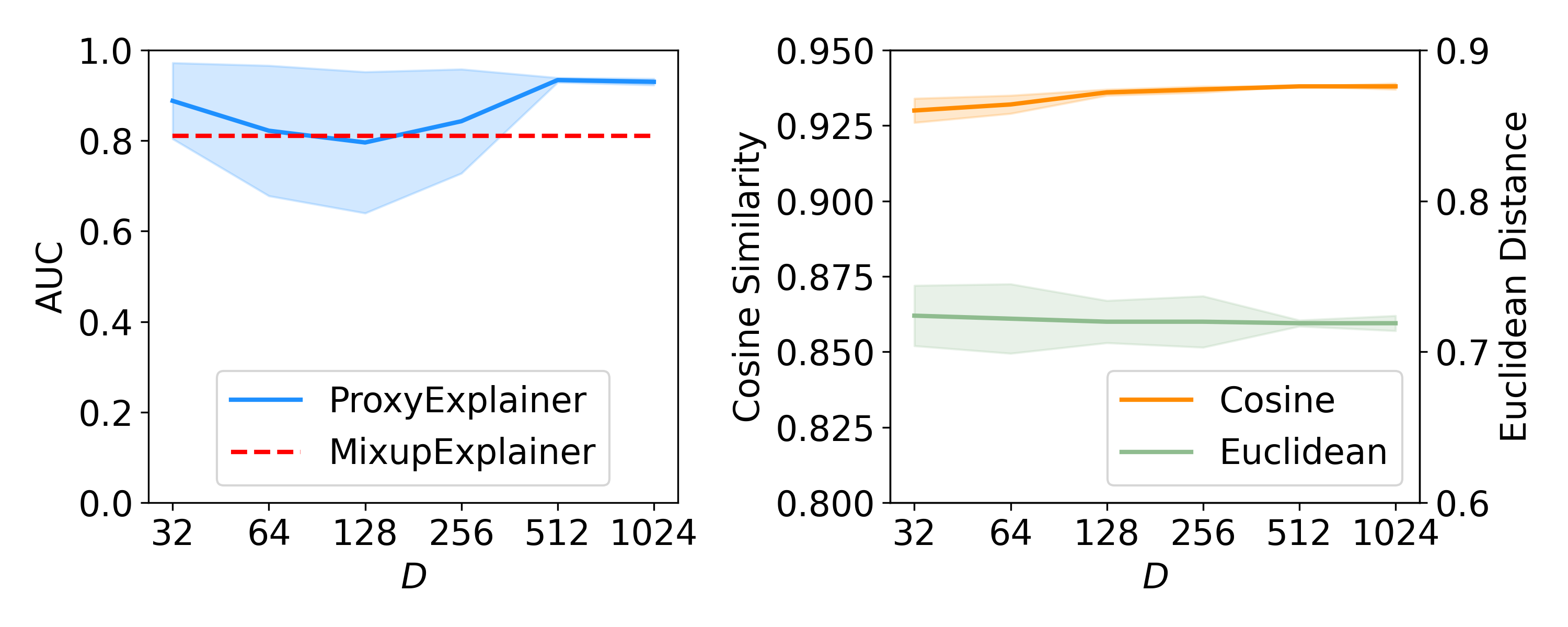}
        \caption{{\alk}}
        \label{fig:third}
    \end{subfigure}

    \begin{subfigure}[b]{0.49\textwidth}
        \centering
        \includegraphics[width=\textwidth]{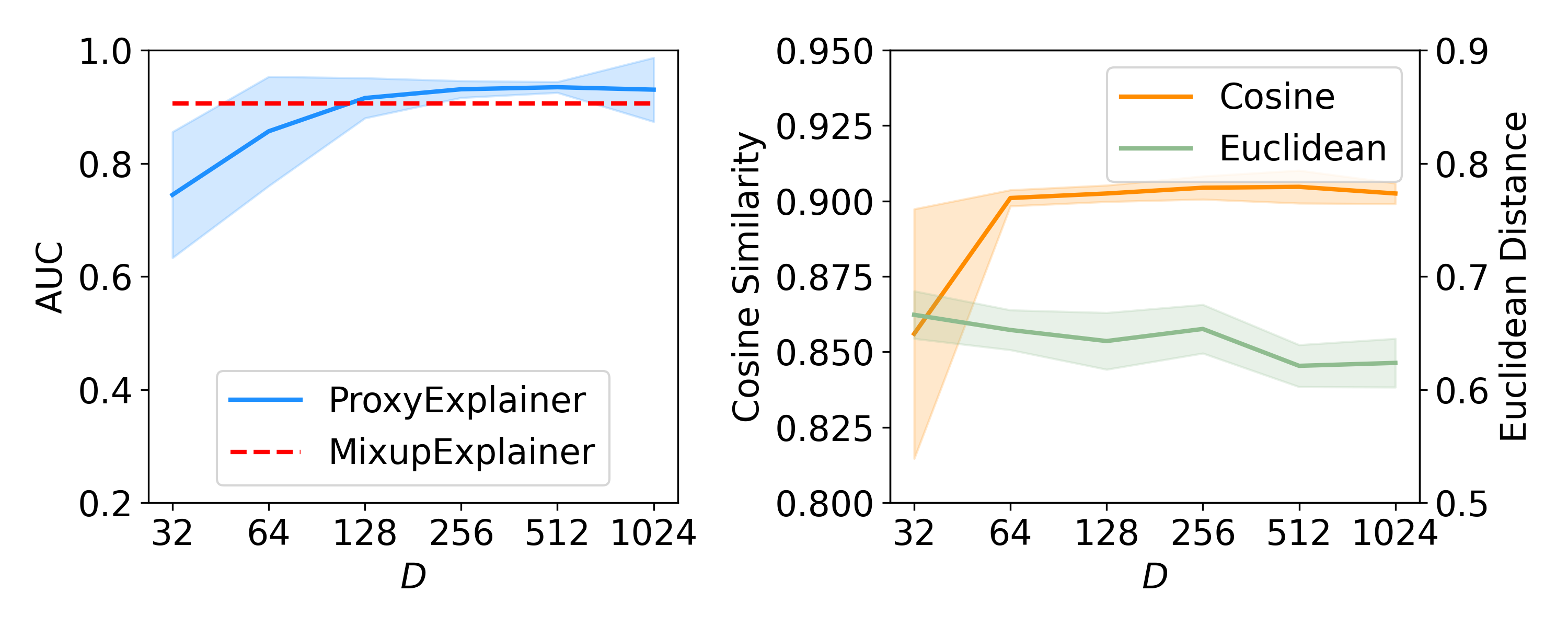}
        \caption{{\bamo}}
        \label{fig:first}
    \end{subfigure}
    \hfill 
    \begin{subfigure}[b]{0.49\textwidth}
        \centering
        \includegraphics[width=\textwidth]{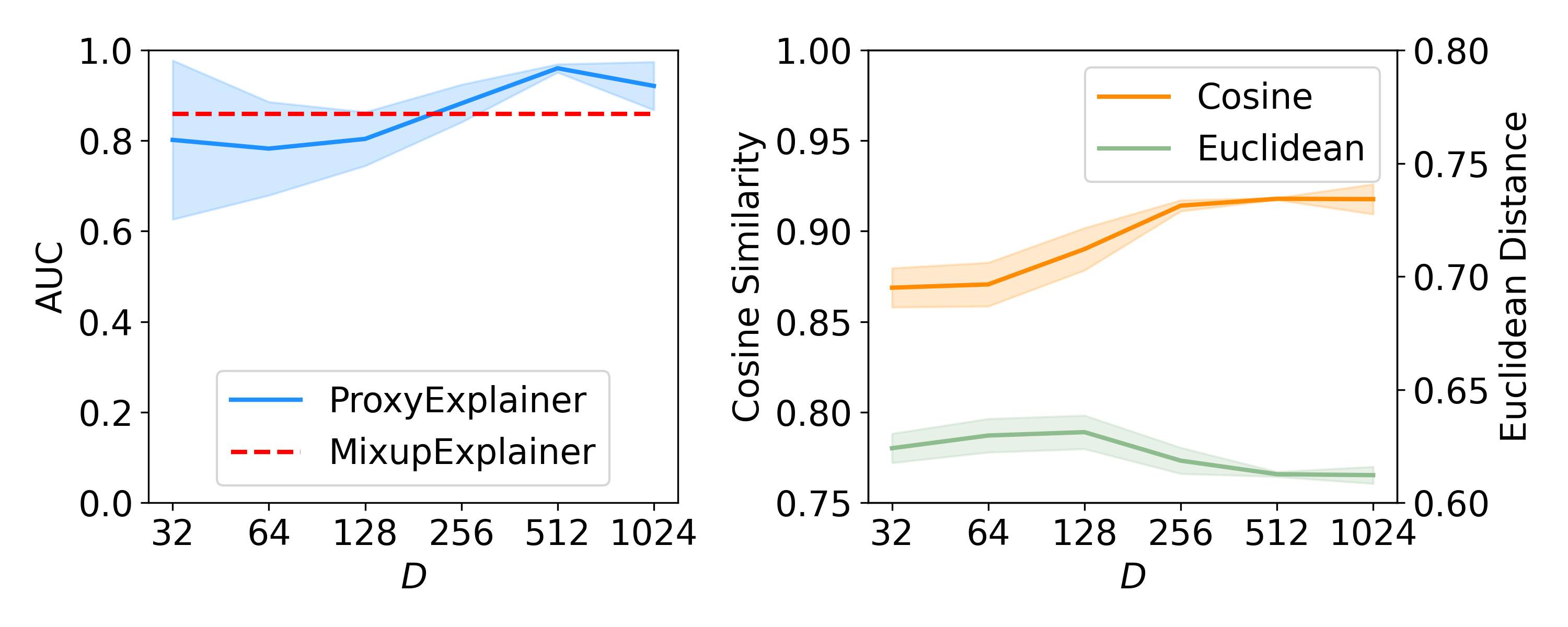}
        \caption{{\bathree}}
        \label{fig:second}
    \end{subfigure}

    \caption{Parameter analysis of the dimension of node latent embedding. The left side of each graph shows the explanation performance and the right side displays the Cosine score and Euclidean distance between $\textit{\textbf{h}}$ and $\tilde{\textit{\textbf{h}}}$.}
    \label{fig:d_all}
\end{figure*}

Figure \ref{fig:lambda_all} shows the performance of our {\ours} with respect to $\lambda$ on two real-world datasets ({\mutag} and {\alk}) and two synthetic datasets ({\bamo} and {\bathree}). From Figure \ref{fig:lambda_all}, we can see that {\ours} consistently outperforms the best baseline, MixupExplainer, for $\lambda \in [0.25, 1.0]$. This indicates that {\ours} is stable. Figure \ref{fig:d_all} presents how the dimension of node latent embedding affects performance in {\ours}, the results show that {\ours} can reach the best performance at $D = 512$.

\newpage
\subsection{Extensive Case Study} \label{sec:extensivecase}

\begin{figure*}[h]
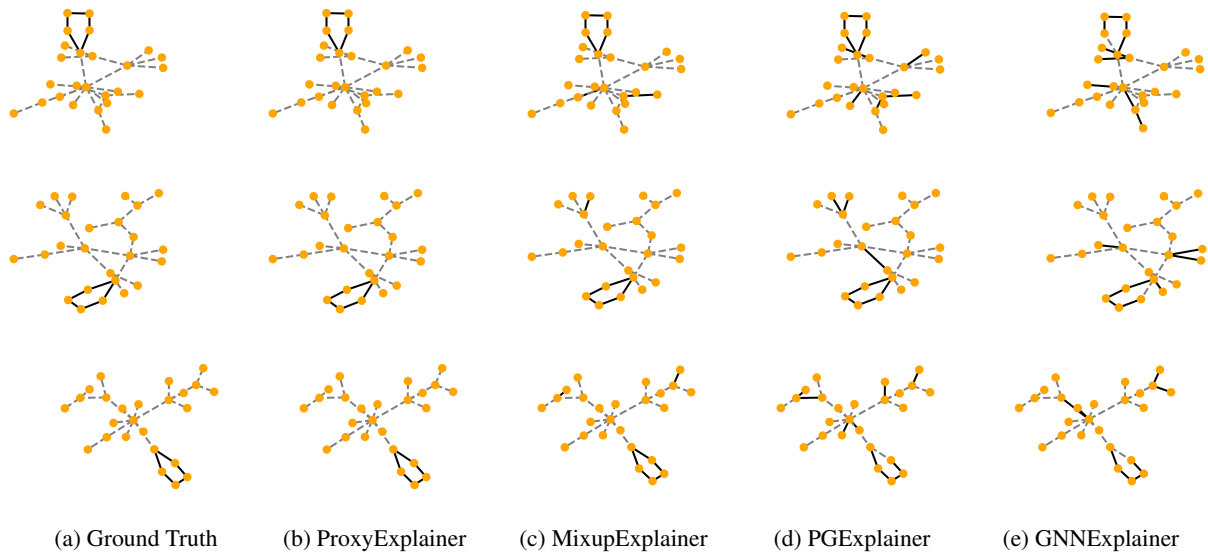

    \centering
    \foreach \i in {6,7,8,9,10} {
        \begin{subfigure}[b]{0.18\textwidth}
            \centering
            \includegraphics[width=\textwidth]{figures/ba2_graph\i.png}
        \end{subfigure}
        \hfill
    }

    \foreach \i in {11,12,13,14,15} {
        \begin{subfigure}[b]{0.18\textwidth}
            \centering
            \includegraphics[width=\textwidth]{figures/ba2_graph\i.png}
        \end{subfigure}
        \hfill
    }

    \begin{subfigure}[b]{0.18\textwidth}
        \centering
        \includegraphics[width=\textwidth]{figures/ba2_graph0.png}
        \caption{Ground Truth}
    \end{subfigure}
    \begin{subfigure}[b]{0.18\textwidth}
        \centering
        \includegraphics[width=\textwidth]{figures/ba2_graph1.png}
        \caption{{\ours}}
    \end{subfigure}
    \begin{subfigure}[b]{0.18\textwidth}
        \centering
        \includegraphics[width=\textwidth]{figures/ba2_graph2.png}
        \caption{MixupExplainer}
    \end{subfigure}
    \begin{subfigure}[b]{0.18\textwidth}
        \centering
        \includegraphics[width=\textwidth]{figures/ba2_graph3.png}
        \caption{PGExplainer}
    \end{subfigure}
    \begin{subfigure}[b]{0.18\textwidth}
        \centering
        \includegraphics[width=\textwidth]{figures/ba2_graph4.png}
        \caption{GNNExplainer}
    \end{subfigure}
    \caption{Visualization of explanation on {\bamo}.}
    \vspace{-0.6em}
    \label{fig:casestudy_ba2}
\end{figure*}

\begin{figure*}[h]
    \centering
    \foreach \i in {6,7,8,9,10} {
        \begin{subfigure}[b]{0.18\textwidth}
            \centering
            \includegraphics[width=\textwidth]{figures/ba3_graph\i.png}
        \end{subfigure}
        \hfill
    }

    \foreach \i in {11,12,13,14,15} {
        \begin{subfigure}[b]{0.18\textwidth}
            \centering
            \includegraphics[width=\textwidth]{figures/ba3_graph\i.png}
        \end{subfigure}
        \hfill
    }

    \begin{subfigure}[b]{0.18\textwidth}
        \centering
        \includegraphics[width=\textwidth]{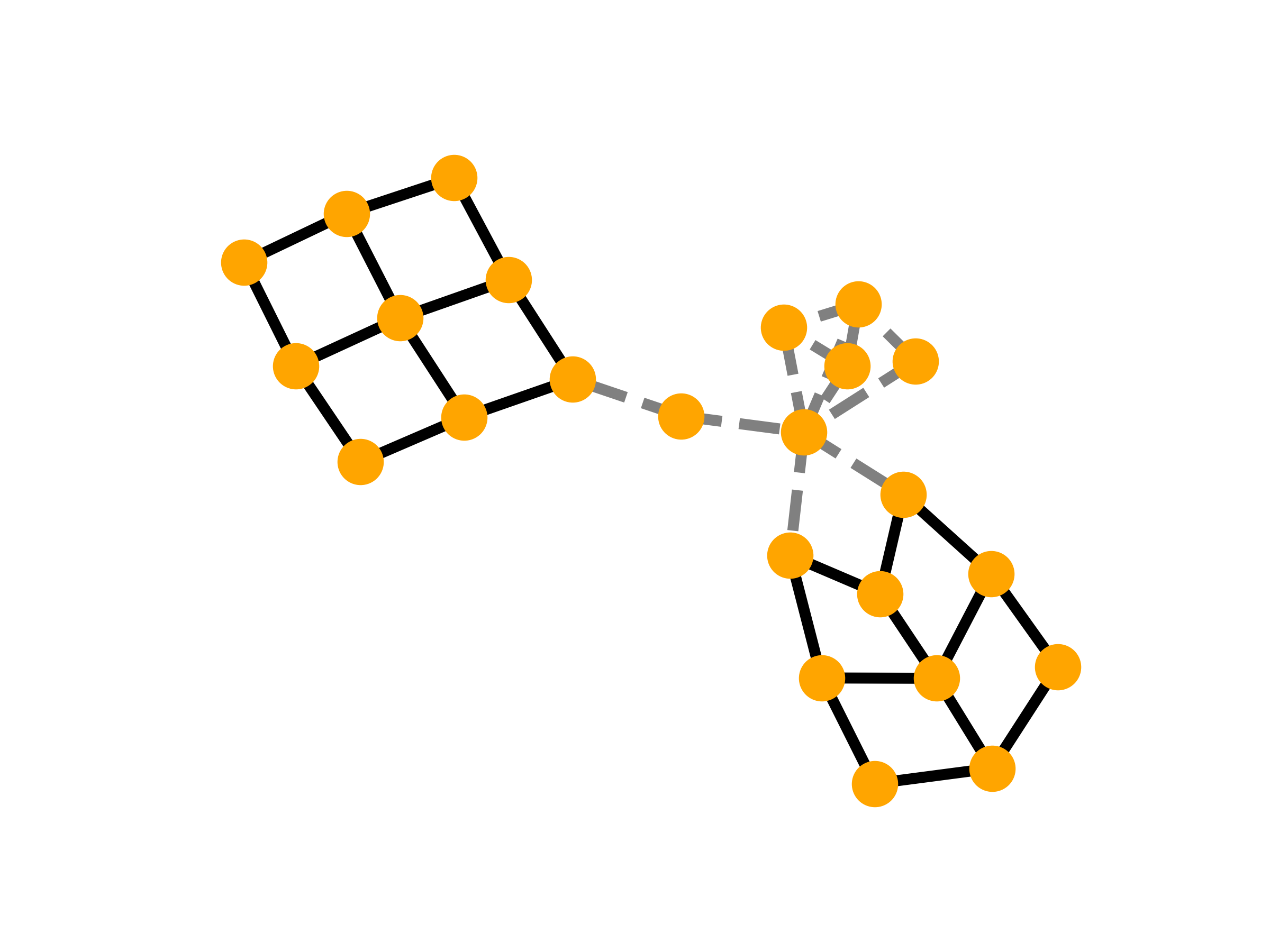}
        \caption{Ground Truth}
    \end{subfigure}
    \begin{subfigure}[b]{0.18\textwidth}
        \centering
        \includegraphics[width=\textwidth]{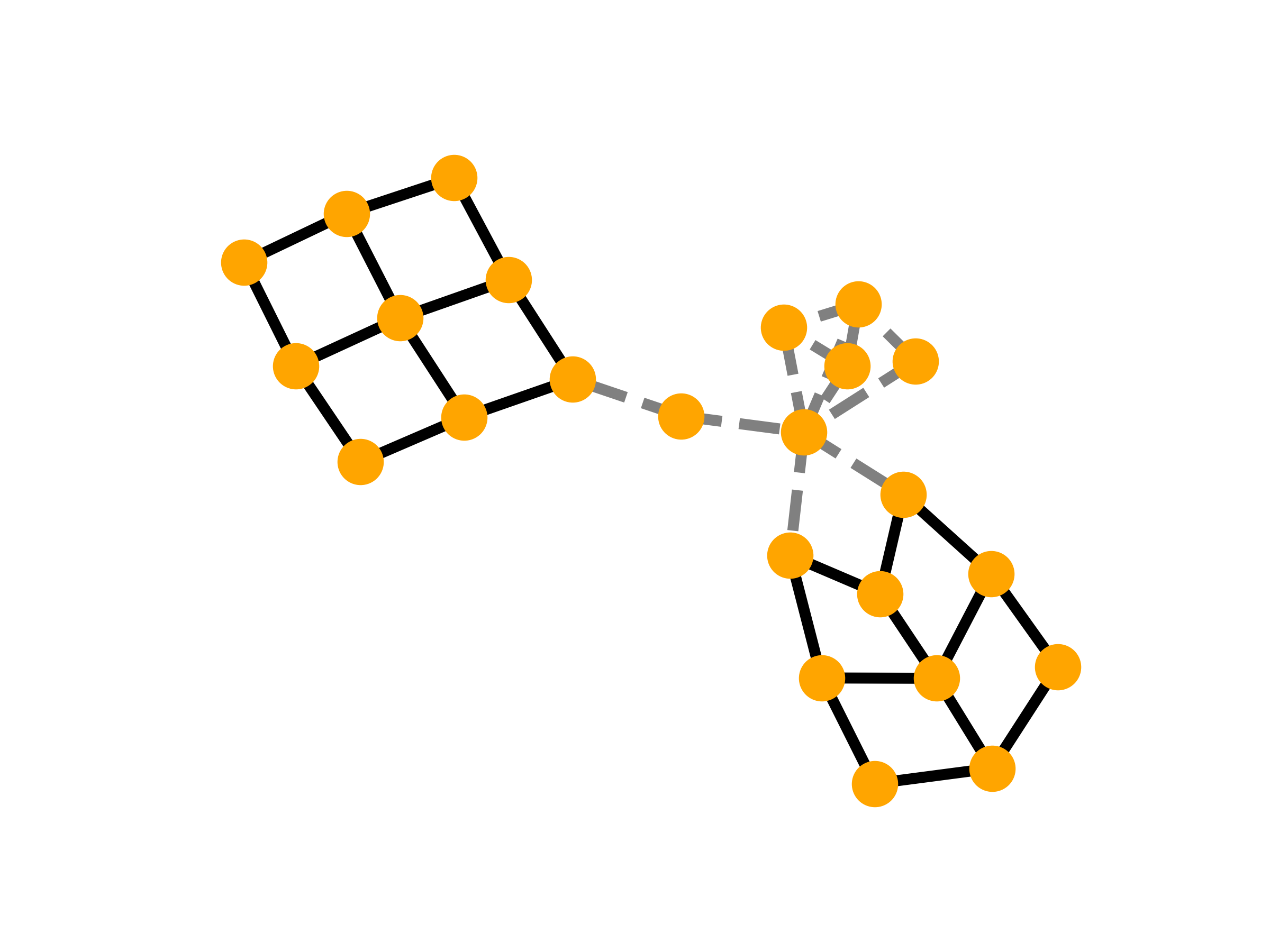}
        \caption{{\ours}}
    \end{subfigure}
    \begin{subfigure}[b]{0.18\textwidth}
        \centering
        \includegraphics[width=\textwidth]{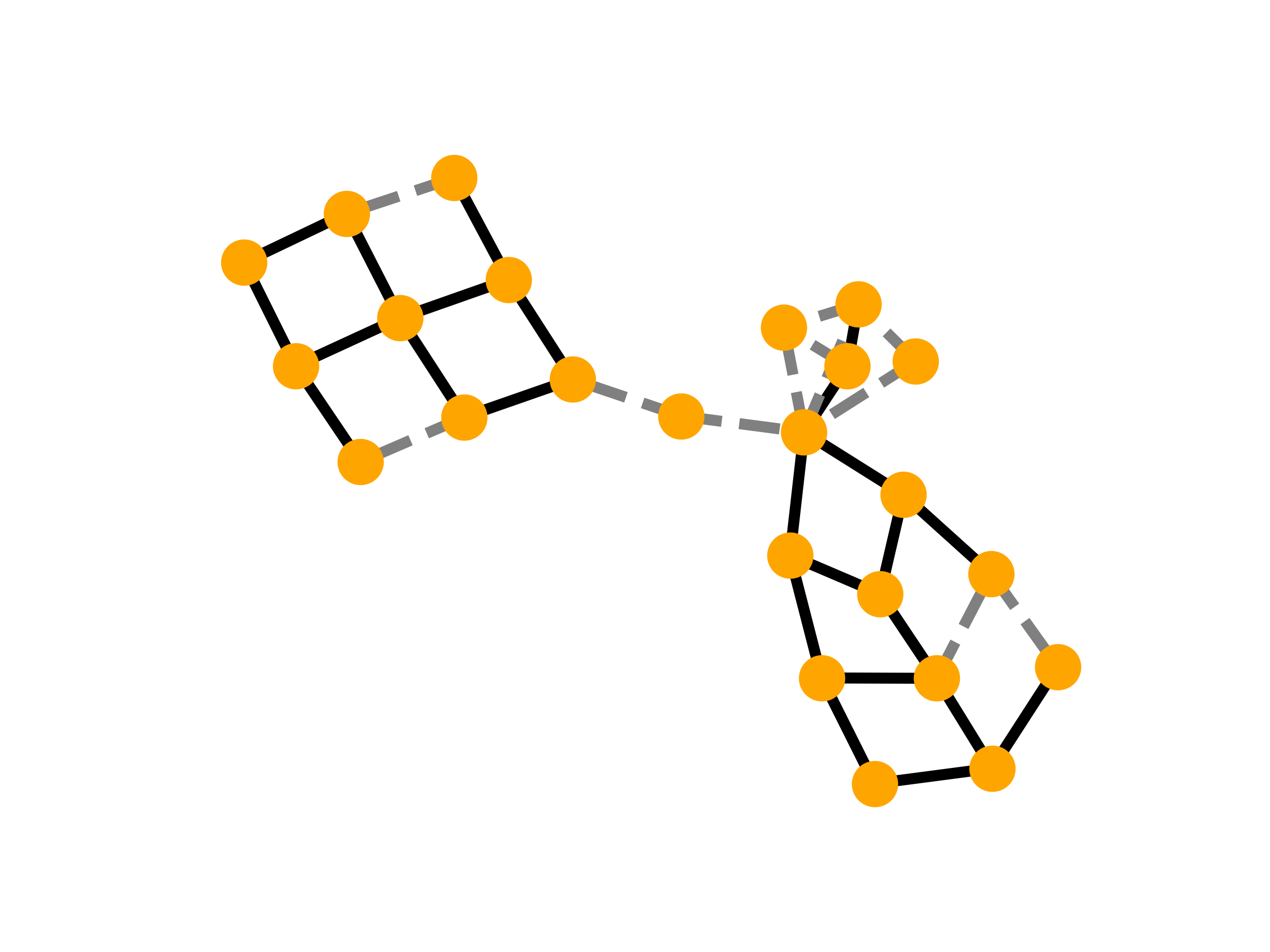}
        \caption{MixupExplainer}
    \end{subfigure}
    \begin{subfigure}[b]{0.18\textwidth}
        \centering
        \includegraphics[width=\textwidth]{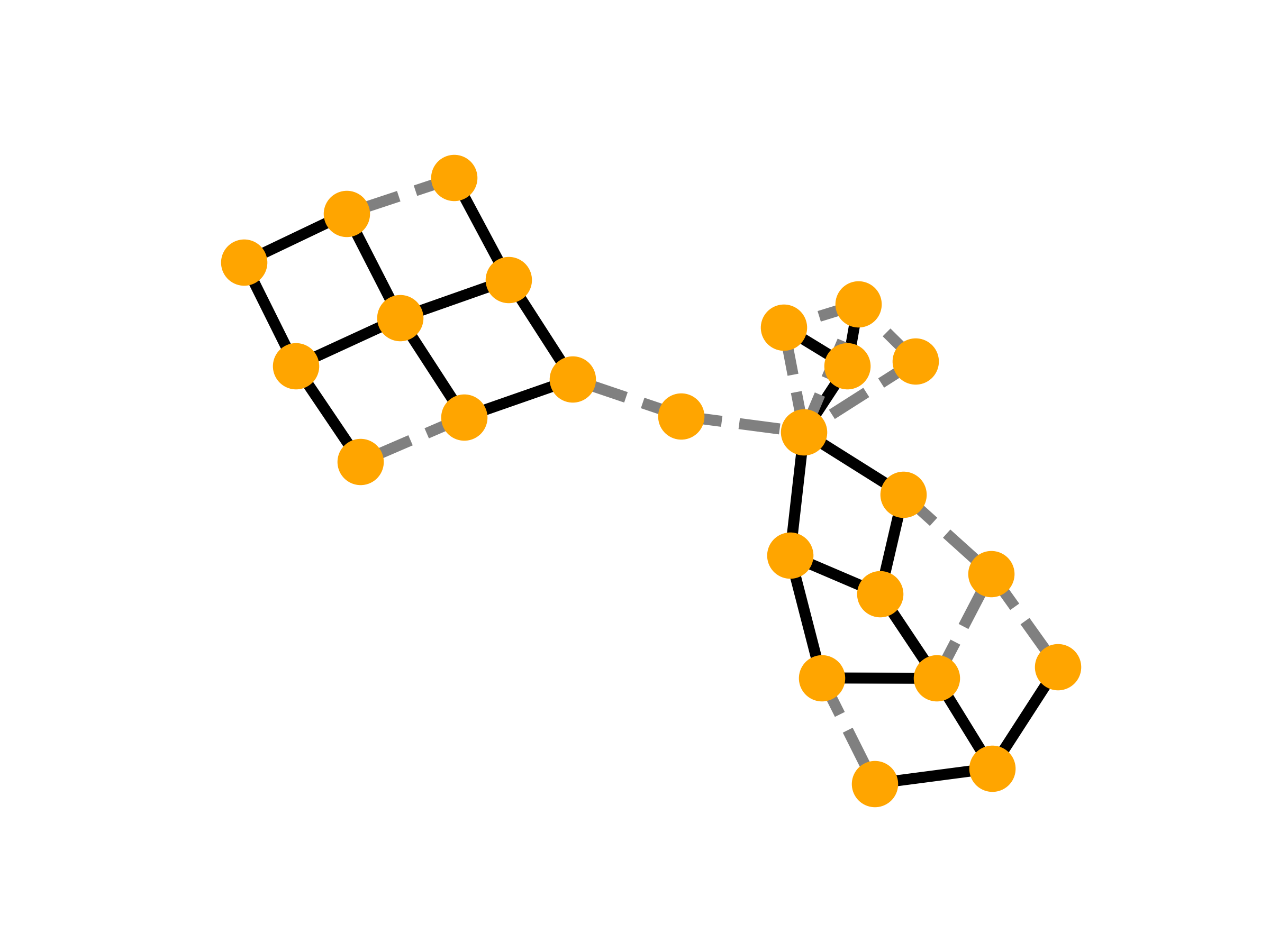}
        \caption{PGExplainer}
    \end{subfigure}
    \begin{subfigure}[b]{0.18\textwidth}
        \centering
        \includegraphics[width=\textwidth]{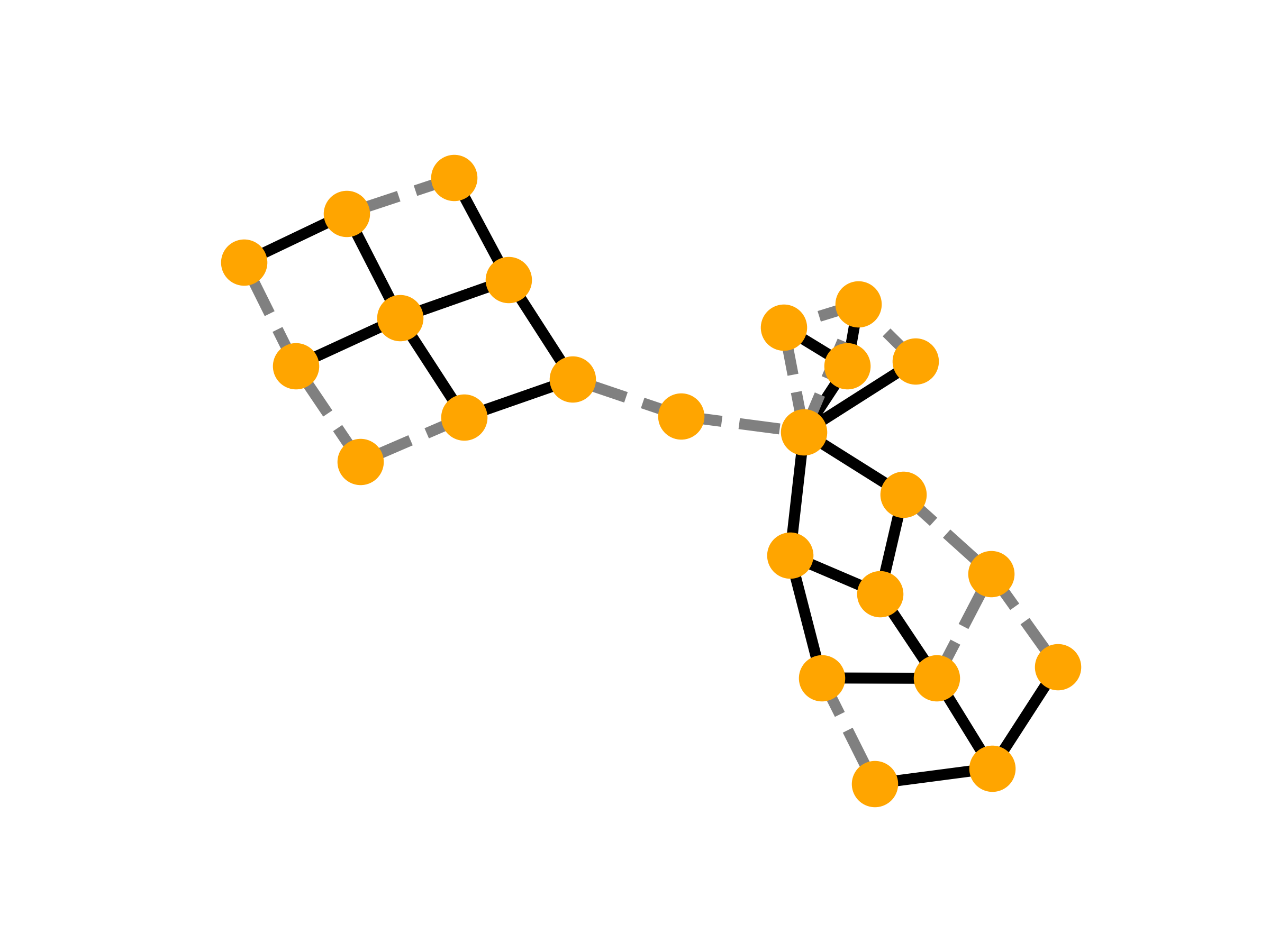}
        \caption{GNNExplainer}
    \end{subfigure}
    \caption{Visualization of explanation on {\bathree}.}
        \vspace{-0.6em}
    \label{fig:casestudy_ba3}
\end{figure*}

\begin{figure*}[h]
    \centering
    \foreach \i in {1,2,3,4,5} {
        \begin{subfigure}[b]{0.19\textwidth}
            \centering
            \includegraphics[width=\textwidth]{figures/figure\i.png}
        \end{subfigure}
        \hfill
    }

    \foreach \i in {6,7,8,9,10} {
        \begin{subfigure}[b]{0.19\textwidth}
            \centering
            \includegraphics[width=\textwidth]{figures/figure\i.png}
        \end{subfigure}
        \hfill
    }
    \foreach \i in {16,17,18,19,20} {
        \begin{subfigure}[b]{0.19\textwidth}
            \centering
            \includegraphics[width=\textwidth]{figures/figure\i.png}
        \end{subfigure}
        \hfill
    } 
    \foreach \i in {11,12,13,14,15} {
        \begin{subfigure}[b]{0.19\textwidth}
            \centering
            \includegraphics[width=\textwidth]{figures/figure\i.png}
            \label{fig:\i}
        \end{subfigure}
        \hfill
    }

    \begin{subfigure}[b]{0.19\textwidth}
        \centering
        \includegraphics[width=\textwidth]{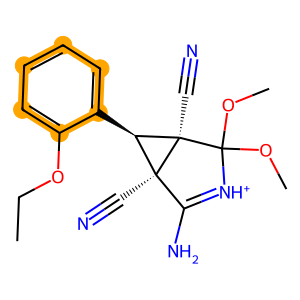}
        \caption{Ground Truth}
    \end{subfigure}
    \begin{subfigure}[b]{0.19\textwidth}
        \centering
        \includegraphics[width=\textwidth]{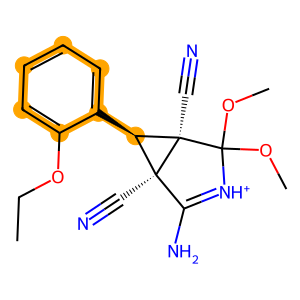}
        \caption{{\ours}}
    \end{subfigure}
    \begin{subfigure}[b]{0.19\textwidth}
        \centering
        \includegraphics[width=\textwidth]{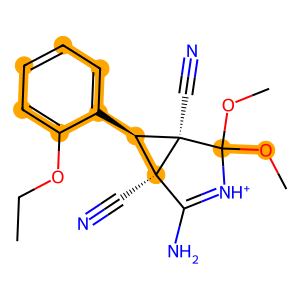}
        \caption{MixupExplainer}
    \end{subfigure}
    \begin{subfigure}[b]{0.19\textwidth}
        \centering
        \includegraphics[width=\textwidth]{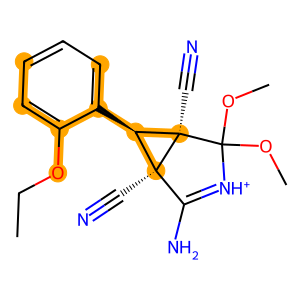}
        \caption{PGExplainer}
    \end{subfigure}
    \begin{subfigure}[b]{0.19\textwidth}
        \centering
        \includegraphics[width=\textwidth]{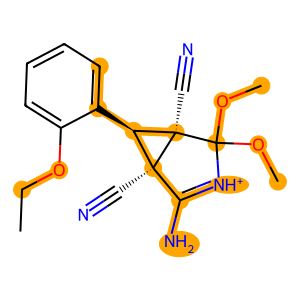}
        \caption{GNNExplainer}
    \end{subfigure}
    \caption{Visualization of explanation on  {\benz}.}
    \label{fig:casestudy_ben}
\end{figure*}

We show more visualization examples of explanation graphs on {\bamo}, {\bathree}, and {\benz} in Figure \ref{fig:casestudy_ba2}, Figure \ref{fig:casestudy_ba3}, and Figure \ref{fig:casestudy_ben}, respectively.

\end{document}